\documentclass{article}
\usepackage[T1]{fontenc}
\usepackage{iclr2026_conference,times}

\usepackage{amsmath,amsfonts,bm}

\def\eqref#1{equation~\ref{#1}}

\def\1{\bm{1}}

\DeclareMathAlphabet{\mathsfit}{\encodingdefault}{\sfdefault}{m}{sl}
\SetMathAlphabet{\mathsfit}{bold}{\encodingdefault}{\sfdefault}{bx}{n}

\usepackage{hyperref}
\usepackage{url}
\usepackage{graphicx}
\usepackage{wrapfig}
\usepackage{booktabs}
\usepackage{subcaption}
\usepackage{tabularx}
\usepackage{array}
\usepackage{multirow}
\usepackage[most]{tcolorbox}
\usepackage{listings}
\usepackage[normalem]{ulem}
\usepackage[table]{xcolor}

\hypersetup{
    colorlinks=true,
    allcolors=blue!50!black
}

\renewcommand{\arraystretch}{1.2}
\newcommand{\se}[1]{\textcolor{red}{#1}}

\newcommand{\sx}[1]{\sout{#1}}
\definecolor{promptAccent}{RGB}{15,56,112}
\definecolor{promptBack}{RGB}{245,248,255}

\lstdefinestyle{mystyle}{
  basicstyle=\fontsize{6}{4.5}\ttfamily,
  breaklines=true,
  frame=single,
  backgroundcolor=\color{gray!5},
  columns=fullflexible
}

\newtcolorbox{PromptBox}[2][]{%
  enhanced, breakable,
  colback=promptBack, colframe=promptAccent,
  coltitle=white, colbacktitle=promptAccent,
  fonttitle=\bfseries\small\ttfamily,
  fontupper=\ttfamily\scriptsize,
  title={#2},
  boxed title style={boxrule=0pt, colframe=promptAccent, colback=promptAccent,
    top=2pt, bottom=2pt, left=6pt, right=6pt},
  attach boxed title to top left={yshift=-1.5mm, xshift=4mm},
  boxrule=0.75pt, arc=2pt, boxsep=5pt,
  before skip=6pt, after skip=6pt,
  borderline west={2pt}{0pt}{promptAccent},
  drop shadow southeast,
  sharp corners=south,
  before upper={\vspace{2pt}},
  #1
}

\newcommand{\headercell}[1]{\normalsize\textbf{#1}}

\newcommand{\desc}[1]{%
  \begin{minipage}[t]{\descw}\vspace{0pt}%
    \fontsize{4.5}{4.5}\selectfont
    \setlength{\parskip}{0pt}\setlength{\parindent}{0pt}%
    \detokenize{#1}%
  \end{minipage}%
}

\newcommand{\imgcell}[1]{%
  \begin{minipage}[t]{\imgw}\vspace{0pt}\centering
    \includegraphics[width=\linewidth,height=\imgh,keepaspectratio]{#1}%
 \end{minipage}%
}
\newcommand{\imgcellbad}[2][red]{%
  \begin{minipage}[t]{\imgw}\vspace{0pt}\centering
    \tcbox[
      enhanced,
      arc=2mm,
      colframe=#1,
      colback=#1!4,
      boxrule=0.5pt,
      left=0mm,right=0mm,top=0mm,bottom=0mm,
      nobeforeafter,
      width=\linewidth,
    ]{%
      \includegraphics[width=\linewidth,height=\imgh,keepaspectratio]{#2}%
    }%
  \end{minipage}%
}
\newcommand{\imgcellokay}[2][orange]{%
  \begin{minipage}[t]{\imgw}\vspace{0pt}\centering
    \tcbox[
      enhanced,
      arc=2mm,
      colframe=#1,
      colback=#1!4,
      boxrule=0.5pt,
      left=0mm,right=0mm,top=0mm,bottom=0mm,
      nobeforeafter,
      width=\linewidth,
    ]{%
      \includegraphics[width=\linewidth,height=\imgh,keepaspectratio]{#2}%
    }%
  \end{minipage}%
}
\newcommand{\imgcellgood}[2][yellow]{%
  \begin{minipage}[t]{\imgw}\vspace{0pt}\centering
    \tcbox[
      enhanced,
      arc=2mm,
      colframe=#1,
      colback=#1!4,
      boxrule=0.5pt,
      left=0mm,right=0mm,top=0mm,bottom=0mm,
      nobeforeafter,
      width=\linewidth,
    ]{%
      \includegraphics[width=\linewidth,height=\imgh,keepaspectratio]{#2}%
    }%
  \end{minipage}%
}
\newcommand{\imgcellverygood}[2][green]{%
  \begin{minipage}[t]{\imgw}\vspace{0pt}\centering
    \tcbox[
      enhanced,
      arc=2mm,
      colframe=#1,
      colback=#1!4,
      boxrule=0.5pt,
      left=0mm,right=0mm,top=0mm,bottom=0mm,
      nobeforeafter,
      width=\linewidth,
    ]{%
      \includegraphics[width=\linewidth,height=\imgh,keepaspectratio]{#2}%
    }%
  \end{minipage}%
}

\newcommand{\legendbox}[1]{%
  \tcbox[
    enhanced,
    arc=1mm,
    colframe=#1,
    colback=#1!15,
    boxrule=0.5pt,
    left=0mm,right=0mm,top=0mm,bottom=0mm,
    nobeforeafter,
  ]{\rule{1mm}{1mm}}%
}

\title{%
  \raisebox{-1.2ex}{\includegraphics[height=1cm]{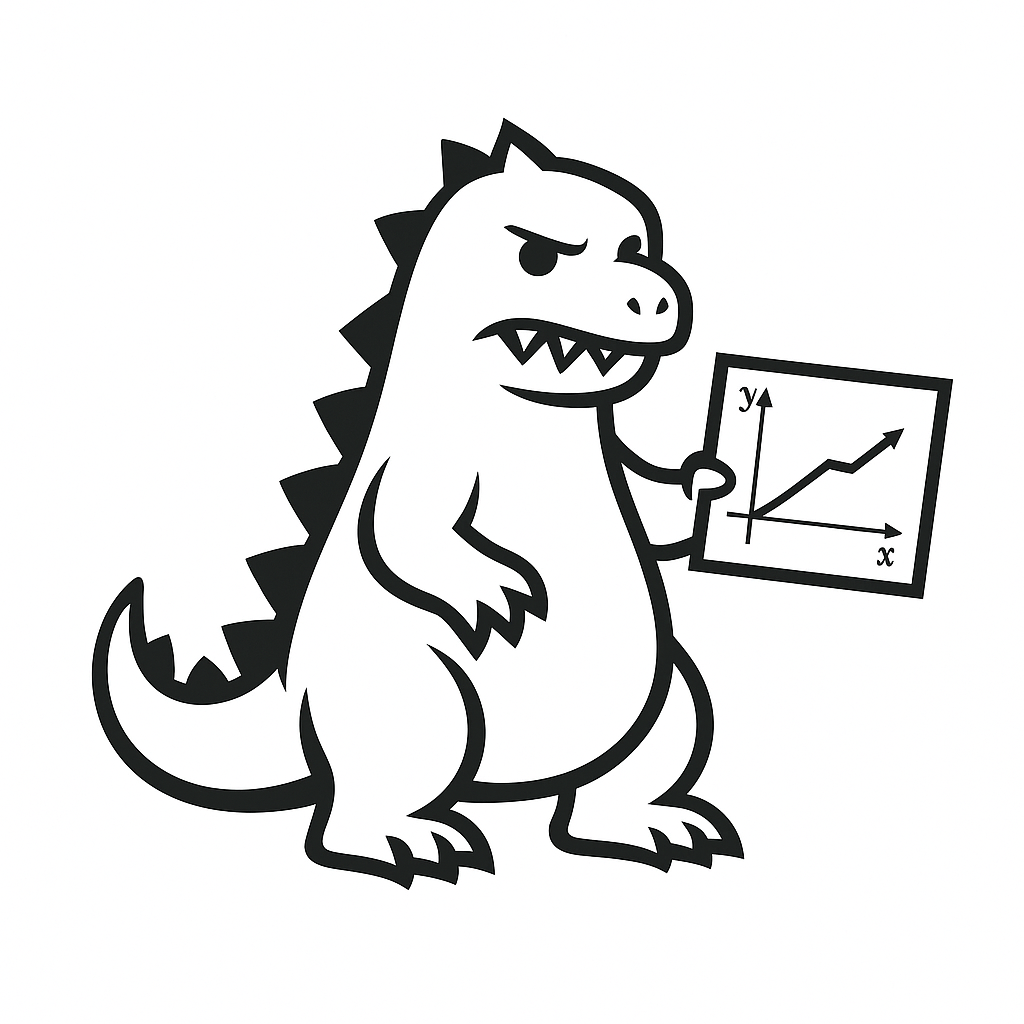}}%
  TikZilla: Scaling Text-to-TikZ with High-Quality Data and Reinforcement Learning
}

\author{Christian Greisinger \& Steffen Eger\\
\small University of Technology Nuremberg \texttt{\{christian.greisinger,steffen.eger\}@utn.de}}

\iclrfinalcopy
\begin{document}

\maketitle

\lhead{Published as a conference paper at ICLR 2026}

\begin{abstract}
Large language models (LLMs) are increasingly used to assist scientists across diverse workflows. A key challenge is generating high-quality figures from textual descriptions, often represented as TikZ programs that can be rendered as scientific images. Prior research has proposed a variety of datasets and modeling approaches for this task. However, existing datasets for Text-to-TikZ are too small and noisy to capture the complexity of TikZ, causing mismatches between text and rendered figures. Moreover, prior approaches rely solely on supervised fine-tuning (SFT), which does not expose the model to the rendered semantics of the figure, often resulting in errors such as looping, irrelevant content, and incorrect spatial relations. To address these issues, we construct DaTikZ-V4, a dataset more than four times larger and substantially higher in quality than DaTikZ-V3, enriched with LLM-generated figure descriptions. Using this dataset, we train TikZilla, a family of small open-source Qwen models (3B and 8B) with a two-stage pipeline of SFT followed by reinforcement learning (RL). For RL, we leverage an image encoder trained via inverse graphics to provide semantically faithful reward signals. Extensive human evaluations with over 1,000 judgments show that TikZilla improves by 1.5-2 points over its base models on a 5-point scale, surpasses GPT-4o by 0.5 points, and matches GPT-5 in the image-based evaluation, while operating at much smaller model sizes. Models and datasets are available on
\href{https://huggingface.co/collections/nllg/tikzilla}{%
  \raisebox{-0.18\height}{\includegraphics[height=1.1em]{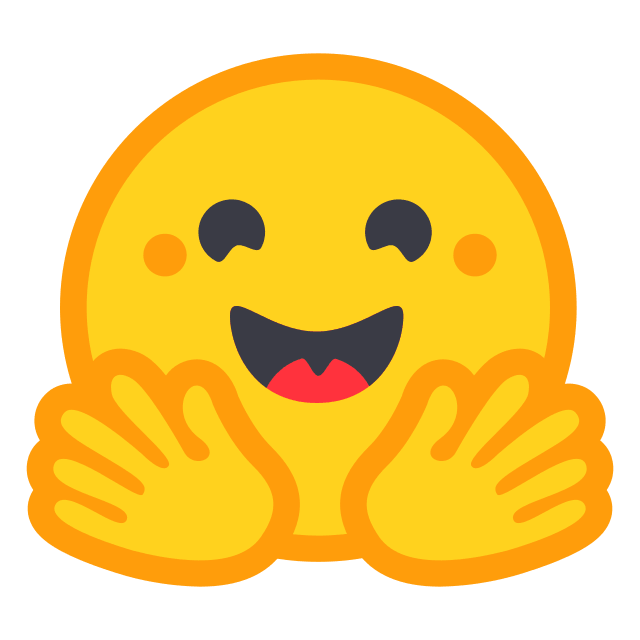}}%
  \,Hugging Face}.
Code is available on
\href{https://github.com/NL2G/TikZilla}{%
  \raisebox{-0.18\height}{\includegraphics[height=1.1em]{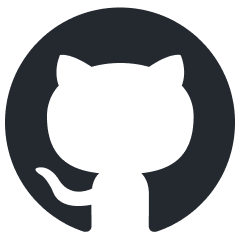}}%
  \,GitHub}.
\end{abstract}

\section{Introduction}
\label{sec:Introduction}

Large language models (LLMs) have become an increasingly valuable tool for scientists across domains~\citep{bi-etal-2024-ai, eger2025transformingsciencelargelanguage}, driven by scaling model size, hardware, and data~\citep{minaee2025largelanguagemodelssurvey}, as well as by research expanding multimodal capabilities~\citep{wu2023multimodallargelanguagemodels} and enabling advanced reasoning~\citep{lu2024aiscientistfullyautomated}. As a result, an increasing number of tools have been developed to support scientists throughout the research process, which range from idea generation~\citep{gottweis2025aicoscientist} to the full automation of scientific outputs ~\citep{lu2024aiscientistfullyautomated}. However, these fully autonomous tools are still far from meeting the high scientific standards required for practical use. Achieving such standards involves overcoming complex subtasks, such as generating accurate scientific images based on textual descriptions~\citep{rodriguez2023figgen,rodriguez2024starvectorgeneratingscalablevector, zou2024vgbench}.

Graphics programming languages such as TikZ are the de facto standard in academia due to their precision, interpretability and seamless integration in the LaTeX ecosystem. However, their steep learning curve and highly varied syntax make them difficult for both humans and LLMs to master~\citep{belouadi2024automatikz}. Prior works have attempted to bridge this gap by finetuning LLMs on caption-TikZ pairs~\citep{belouadi2024automatikz, belouadi2025tikzerozeroshottextguidedgraphics}. Due to the sparsely available data,~\cite{belouadi2025tikzerozeroshottextguidedgraphics} leverage captioned images without the underlying graphics program available, therefore having access to a much richer dataset. However, these efforts remain limited by noisy captions, a lack of executable and standardized TikZ code, as well as a lack of direct visual feedback, leaving models prone to low compilation rates, hallucinations, overly long responses, and low-quality outputs.

\setlength{\aboverulesep}{0pt}
\setlength{\belowrulesep}{0pt}
\setlength{\tabcolsep}{4pt}
\renewcommand{\arraystretch}{1.03}
\setlength{\abovetopsep}{0pt}

\newlength{\descw}\setlength{\descw}{0.333\textwidth}
\newlength{\imgw}\setlength{\imgw}{0.14\textwidth}
\newlength{\imgh}\setlength{\imgh}{3.2cm}

\newcolumntype{Y}{>{\raggedright\arraybackslash}p{\descw}}
\newcolumntype{C}{>{\centering\arraybackslash}p{\imgw}}

\newcommand{\exrowone}[5]{%
  \desc{#1} & \imgcell{#2} & \imgcellbad{#3} & \imgcellokay{#4} & \imgcellverygood{#5} \\
}
\newcommand{\exrowtwo}[5]{%
  \desc{#1} & \imgcell{#2} & \imgcellbad{#3} & \imgcellokay{#4} & \imgcellgood{#5} \\
}
\newcommand{\exrowthree}[5]{%
  \desc{#1} & \imgcell{#2} & \imgcellokay{#3} & \imgcellgood{#4} & \imgcellverygood{#5} \\
}
\newcommand{\exrowfour}[5]{%
  \desc{#1} & \imgcell{#2} & \imgcellbad{#3} & \imgcellbad{#4} & \imgcellgood{#5} \\
}

\begin{table}[t]
\centering
\caption{Exemplary scientific TikZ figures produced by one baseline LLM (GPT-4o) and two of our finetuned LLMs (TikZilla-3B and TikZilla-3B-RL) using the prompts from the first column which have been VLM augmented based on the Ground Truth figures in the second column. \legendbox{green}-boxed figures have been rated as very good, \legendbox{yellow} as good, \legendbox{orange} as bad, and \legendbox{red} as very bad by human annotators. Additional examples are provided in the Appendix (Table~\ref{tab:examples_1},~\ref{tab:examples_2},~\ref{tab:examples_3}, and~\ref{tab:examples_4})}
\label{tab:examples}
\begin{tabularx}{\textwidth}{YCCCC}
    \toprule
    \headercell{Prompt} & \headercell{Ground Truth} & \headercell{GPT-4o} & \headercell{TikZilla-3B} & \headercell{TikZilla-3B-RL} \\
    \midrule

    \exrowone{A lattice diagram consists of nodes connected by thin black lines. At the top center, node "XYAB" is labeled with $h=4$ in green below it. Directly below, five nodes "AX", "AY", "XY", "XB", and "YB" are horizontally aligned, each labeled with $h=3$ in green below. Below these, nodes "A", "X", "Y", and "B" are horizontally aligned, each labeled with $h=2$ in green below. At the bottom center, node "$\emptyset$" is labeled with $h=0$ in green above. Lines connect "XYAB" to each of the nodes in the second row. "AX" connects to "A" and "X", "AY" connects to "A" and "Y", "XY" connects to "X" and "Y", "XB" connects to "X" and "B", and "YB" connects to "Y" and "B". Each node is connected to the node "$\emptyset$" at the bottom.}
        {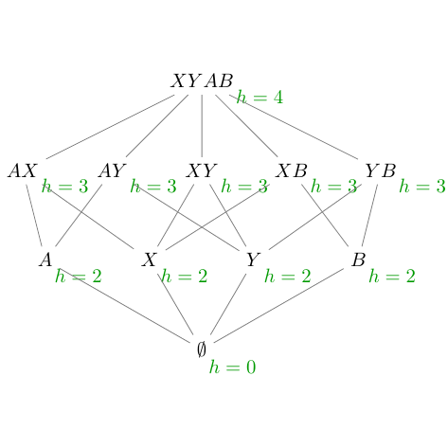}
        {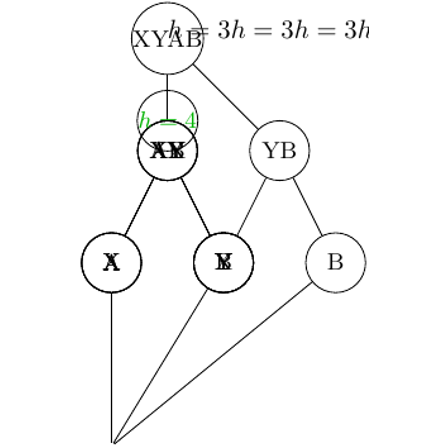}
        {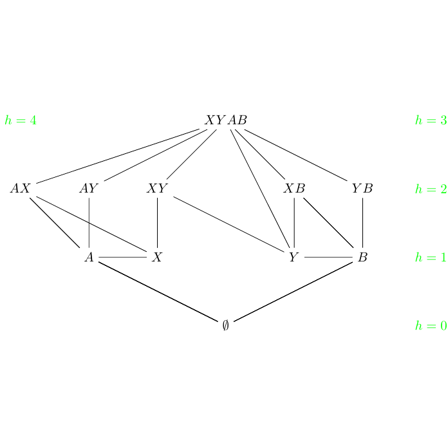}
        {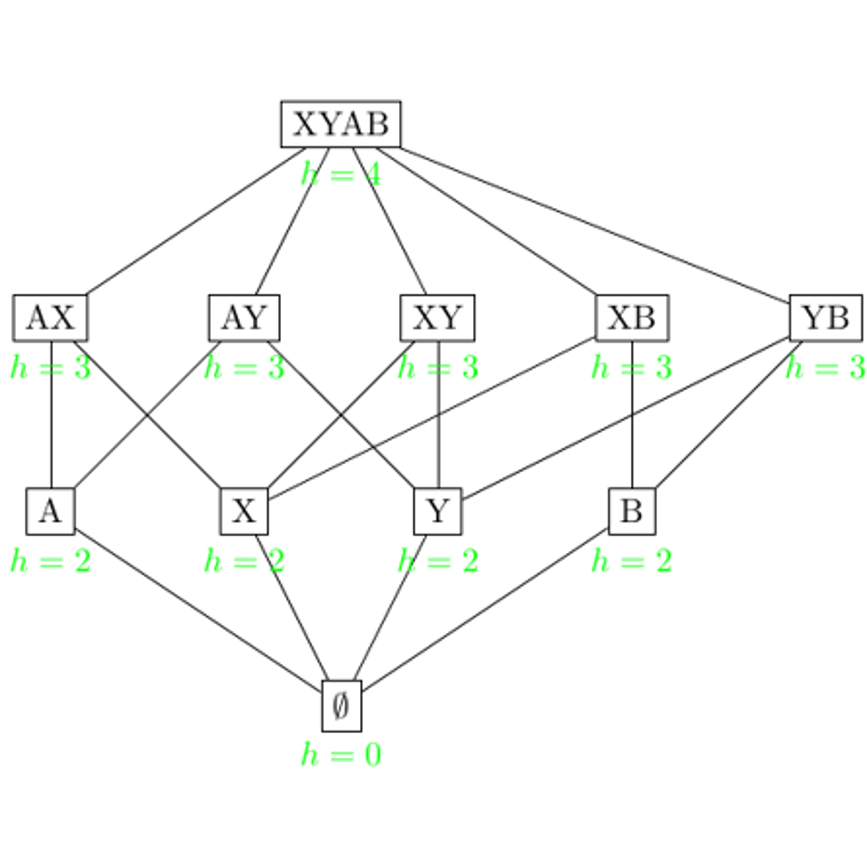}
    \midrule
    \exrowtwo{A flowchart consists of various colored shapes connected by arrows. At the top left, a red rounded rectangle labeled "Start" connects via a rightward arrow to a green parallelogram labeled "$\Psi_{\text{prep}}$". Below, a blue rectangle labeled "LDPC$_{\text{quantum}}$" connects leftward to another blue rectangle labeled "PAT Integration". This rectangle connect downward to a large yellow diamond labeled "Atmospheric Correction". From the diamond's right side, a rightward arrow leads to a blue rectangle labeled "Transmission", which connects downward to another blue rectangle labeled "Measurement". A downward arrow leads to a green parallelogram labeled "Information Decoding", which connects downward to a red rounded rectangle labeled "End". All arrows are black and connect the shapes in a logical sequence.}
        {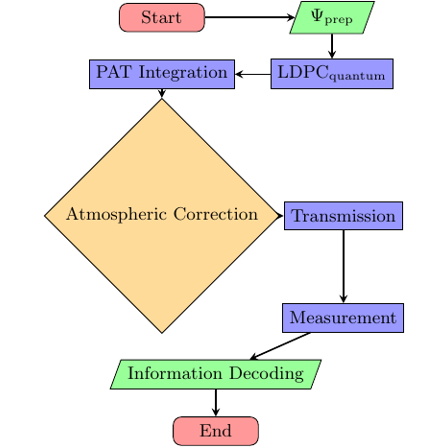}
        {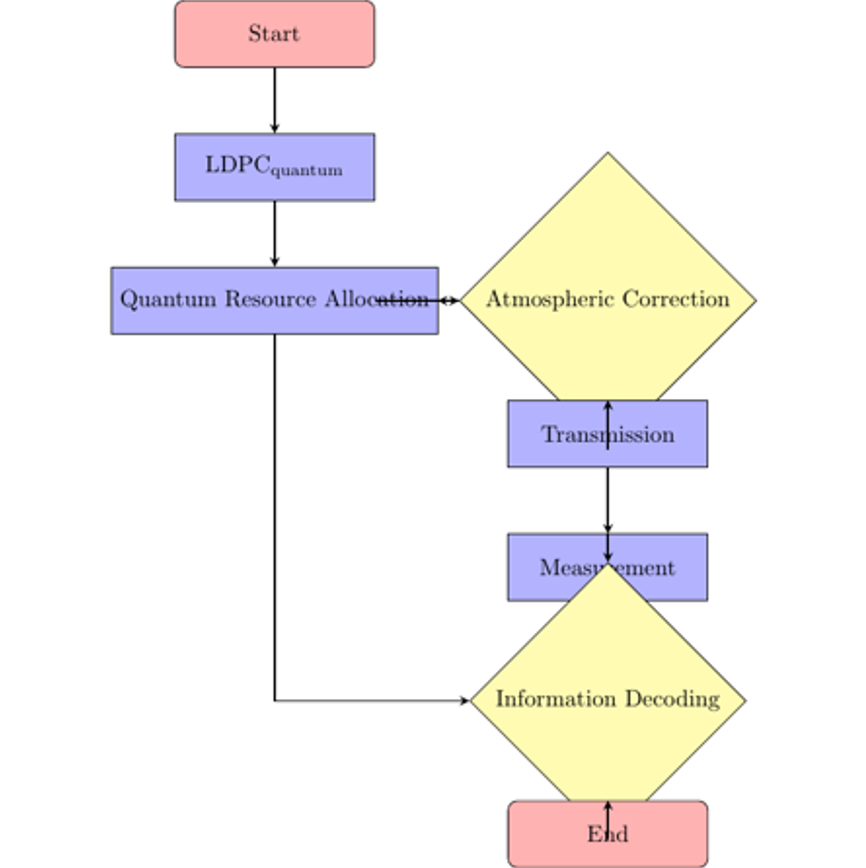}
        {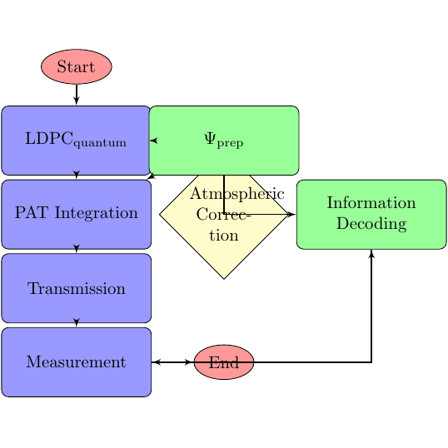}
        {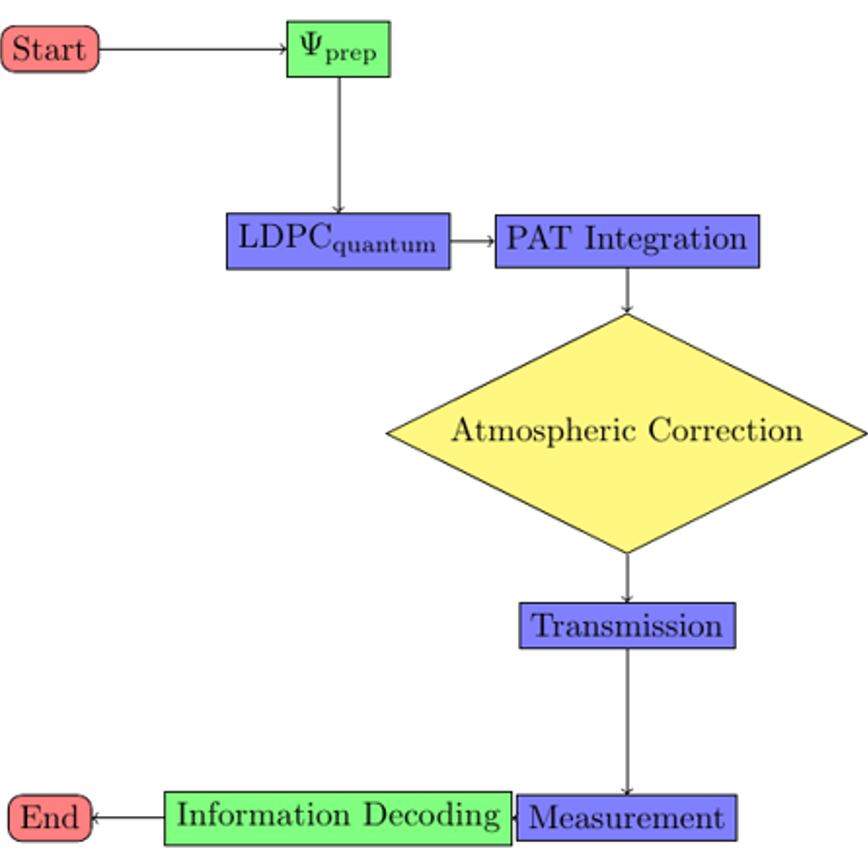}
    \midrule
    \exrowthree{A horizontal sequence of four rectangles is centered in the image. The first rectangle on the left is gray, labeled $y$ in black at its center, and has the label "Input" above it. To its right, a black arrow points from the center of the first rectangle to the center of the second rectangle, which is green and labeled $g_\theta(\cdot)$ in black. Below the green rectangle, the label "Encoder" is centered. A black arrow extends from the center of the green rectangle to the center of the third rectangle, which is purple and labeled $f_\theta(\cdot)$ in black. Below the purple rectangle, the label "Decoder" is centered. A black arrow points from the center of the purple rectangle to the center of a fourth gray rectangle on the right, labeled $\hat{y}$ in black. Above this fourth rectangle, the label "Output" is centered. The label $x$ is placed above the arrow between the green and purple rectangles.}
        {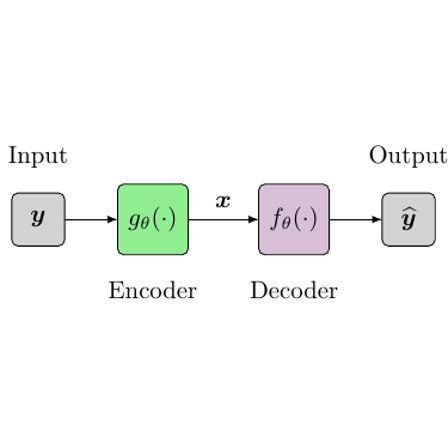}
        {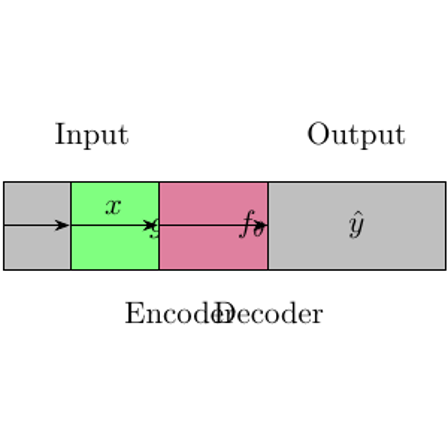}
        {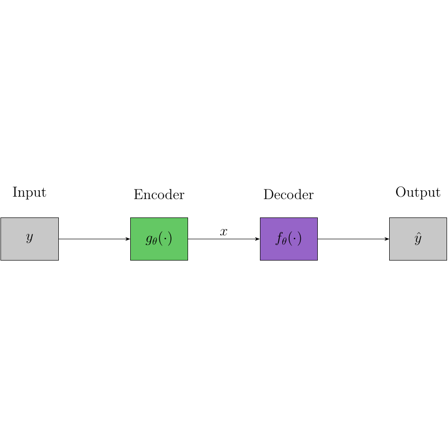}
        {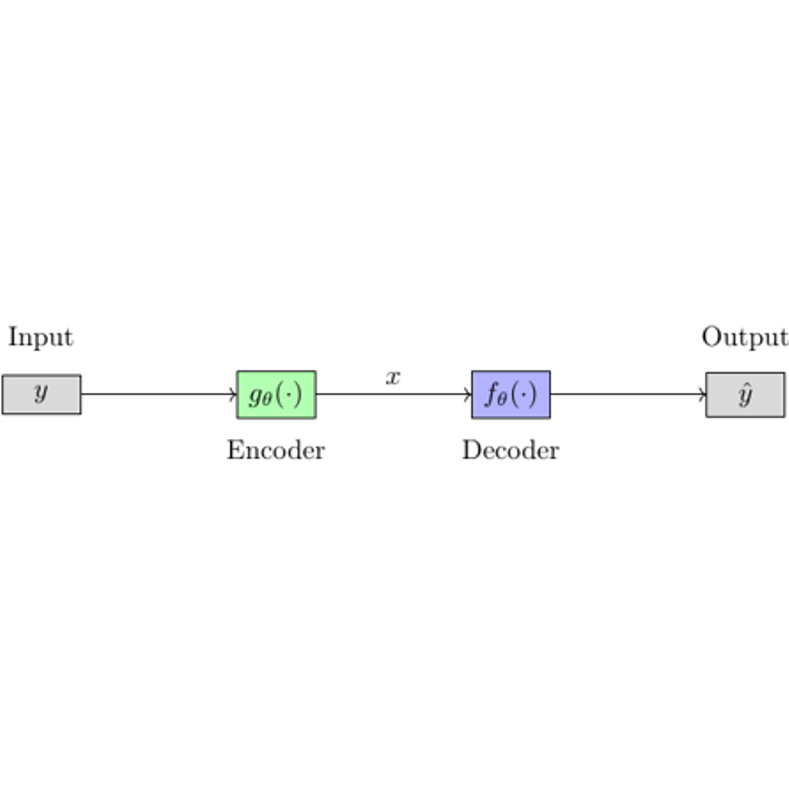}
    \midrule
    \exrowfour{A horizontal bar chart labeled 'SHAP Value' in the top right corner. Each bar is colored blue and corresponds to a specific feature, with the SHAP value displayed numerically at the end of each bar. The chart is sorted in ascending order of SHAP value, starting with 'Chemotherapy' at the top, which has the smallest value (11.07). This is followed by 'Gender' (27.23), 'dim z' (33.36), 'Surgery' (42.25), 'Age' (47.55), 'Count 2' (55.87), 'Count 0' (66.72), 'Count 1' (69.87), 'performance status' (77.32), 'Weight' (77.94), 'eGFR' (89.16), 'HPVstatus' (149.42), 'CenterId' (157.17) and 'Tobacco' with the largest value (221.44).}
        {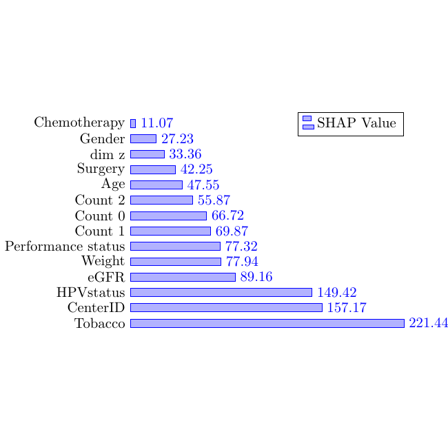}
        {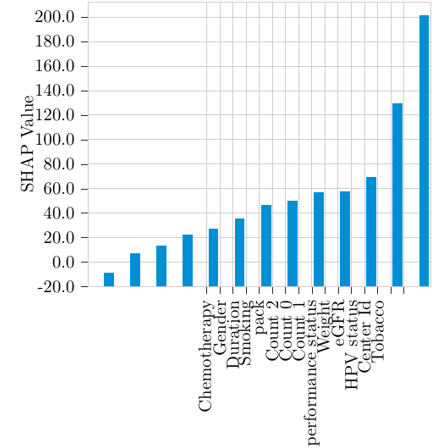}
        {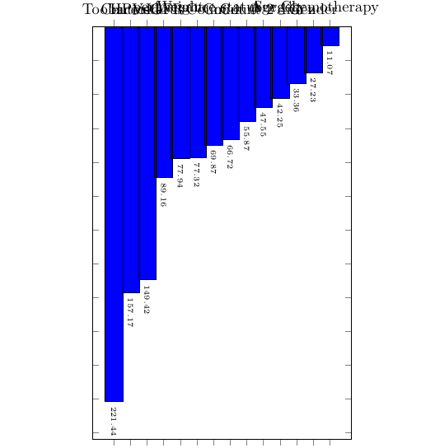}
        {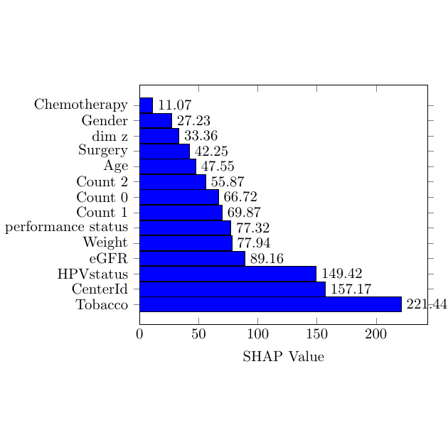}

    \bottomrule
\end{tabularx}
\end{table}

We address these limitations by constructing DaTikZ-V4, a dataset more than 1.5M instances larger than its predecessor, sourced from arXiv, GitHub, TeX StackExchange (TeX SE), and synthetic data. To improve data quality, we introduce an LLM-based debugging pipeline that repairs uncompilable TikZ code, and employ Vision Language Models (VLMs) to generate accurate figure descriptions. Building on DaTikZ-V4, we develop TikZilla, a family of small Qwen-based models (3B and 8B) trained with a two-stage pipeline: Supervised Finetuning (SFT) for syntax alignment, followed by Reinforcement Learning (RL) with a reward model trained on the Image-to-TikZ task beforehand. We find that this approach substantially improves Text-to-TikZ generation quality, where even models as small as 3B parameters outperform GPT-4o across automatic metrics and over 1,000 human judgments spanning four baseline LLMs. Table~\ref{tab:examples} shows examples with corresponding human ratings. We summarize our key contributions as follows:
\begin{itemize}
    \item \textbf{Caption Quality Analyisis:} We show that widely available captions are insufficient for reconstructing figures.
    \item \textbf{Scaling Dataset Size:} We introduce DaTikZ-V4 with over 2M unique TikZ samples, sourced from newer arXiv submissions and GitHub, quadrupling the scale of prior datasets.
    \item \textbf{Data Quality Enhancements:} We combine (1) improved rule-based filtering (e.g., dynamic package inclusion), (2) VLM-based scientific figure descriptions, and (3) an LLM debugging pipeline for uncompilable TikZ code.
    \item \textbf{Reward Model:} We finetune an image encoder on the Image-TikZ task using our larger TikZ corpus, providing more semantically meaningful rewards for RL optimization.
    \item \textbf{TikZilla Models:} We release TikZilla, a family of small open-source Qwen models (3B and 8B). TikZilla outperforms GPT-4o across automatic and human evaluation, and matches GPT-5 in image-based evaluation, despite operating at much smaller model sizes.
\end{itemize}

\section{Related Work}
\label{sec:Related Work}

\paragraph{Text-Guided Graphics Program Generation for Scientific Figures}
Generating vector graphics such as SVG or TikZ is essential in scientific publishing due to their fidelity and interpretability. Early approaches relied on handcrafted heuristics or neural sequence models to approximate images with path primitives~\citep{Lopes2019ALR, NEURIPS2020_bcf9d6bd}, but these struggled with complex scientific figures. More recently, LLM-based methods have emerged: AutomaTikZ~\citep{belouadi2024automatikz} finetunes on caption–TikZ pairs from arXiv and TeX SE, while StarVector~\citep{rodriguez2024starvectorgeneratingscalablevector} focuses on SVG generation with a dedicated benchmark. Yet for TikZ, dataset sparsity remains a bottleneck. TikZero~\citep{belouadi2025tikzerozeroshottextguidedgraphics} partially addresses this by combining an inverse-graphics model~\citep{belouadi2024detikzify} with a modality-bridging adapter~\citep{hu2023llmadaptersadapterfamilyparameterefficient}, distilling supervision from text–image pairs. However, TikZero still depends on noisy captions and cannot finetune its text decoder without paired graphics programs, limiting performance. In contrast, we construct a dataset over four times larger and of higher quality, pairing TikZ programs with VLM-generated descriptions, enabling small LLMs to be effectively finetuned for Text-to-TikZ.

\paragraph{Post-training with Reinforcement Learning}  
Advances in RL such as Group Relative Policy Optimization (GRPO)~\citep{deepseek-math} allow to more efficiently align LLMs either with human preferences~\citep{NEURIPS2022_b1efde53} or verifiable tasks~\citep{lambert2025tulu3pushingfrontiers}. For example, RLEF~\citep{gehring2025rlefgroundingcodellms} iteratively leverages execution feedback for code synthesis, ~\cite{yoshihara2025practicaltwostagerecipemathematical} enhance LLM reasoning on math benchmarks, and VisionR1~\citep{huang2025visionr1incentivizingreasoningcapability} extends reasoning capabilities to the multimodal domain. Closest to our setting, RLRF~\citep{rodriguez2025renderingawarereinforcementlearningvector} optimizes SVG code generation via composite rewards assessing code efficiency, semantic alignment, and visual fidelity. Our work differs in two ways: we focus on TikZ generation for scientific figures, and we introduce a domain-specific reward model, trained through inverse-graphics (Image–TikZ), which better captures semantics than general-purpose metrics such as CLIPScore~\citep{hessel-etal-2021-clipscore} or DreamSIM~\citep{10.5555/3666122.3668330}.

\section{Caption Quality Analysis}
\label{sec:Caption Quality Analysis}

\begin{figure*}[t]
  \centering
  \begin{tabular*}{\textwidth}{@{} p{0.52\textwidth} @{\hspace{1.5em}} p{0.4\textwidth} @{}}
    \begin{minipage}[t]{\linewidth}\vspace{0pt}\centering
      \includegraphics[width=\linewidth]{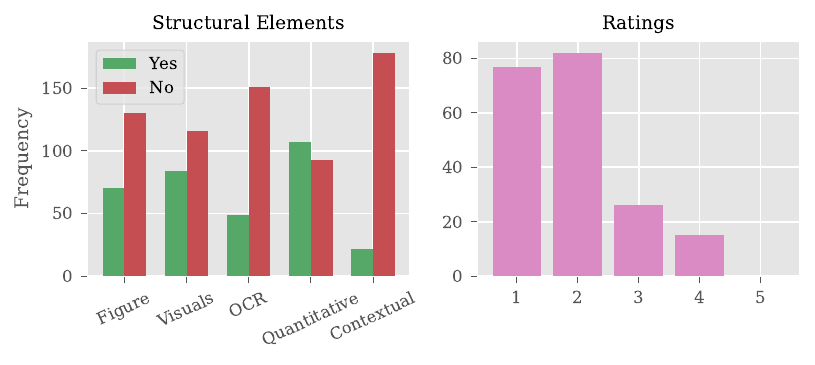}
    \end{minipage}
    &
    \begin{minipage}[t]{\linewidth}\vspace{10pt}\centering
      \tiny
      \begin{tabular}{lcccc}
        \toprule
        \textbf{Variant} & \textbf{BLEU-4}$\uparrow$ & \textbf{ROUGE-L}$\uparrow$ & \textbf{STS}$\uparrow$ & \textbf{Length} \\
        \midrule
        Captions       & 0.003 & 0.098 & 0.355 &  34.0 \\
        \midrule
        Qwen2.5-VL-7B  & 0.068 & 0.276 & 0.744 & 126.3 \\
        Qwen2.5-VL-32B & 0.047 & 0.242 & 0.719 & 177.8 \\
        InternVL3-8B   & 0.045 & 0.235 & 0.716 & 159.8 \\
        InternVL3-38B  & 0.057 & 0.264 & 0.743 & 141.6 \\
        \midrule
        GPT-4o-mini    & 0.073 & 0.281 & 0.761 & 140.9 \\
        GPT-4o         & \underline{0.089} & \underline{0.317} & \underline{0.777} & 123.5 \\
        \midrule
        Human          & \textbf{0.094} & \textbf{0.318} & \textbf{0.815} & 105.3 \\
        \bottomrule
      \end{tabular}
    \end{minipage}
  \end{tabular*}
  \caption{Left: human evaluation of caption quality by structural elements and usefulness ratings. Right: BLEU-4, ROUGE-L, STS, and average length for raw captions, VLM-generated descriptions, and human-written descriptions (using other human descriptions as references).}
  \label{fig:dataset_quality_eval}
\end{figure*}

Accurate Text-to-TikZ generation requires captions that specify objects, attributes, and spatial relations~\citep{zhang2025scimage}. To assess whether existing captions meet this need, we analyzed 200 samples from DaTikZ-V3 with three annotators (Figure~\ref{fig:dataset_quality_eval}, left). The annotators checked captions for missing structural elements (e.\ g.\ figure type, components, and labels) and judged usefulness on a 1–5 Likert scale. Two findings emerged: (i) key details such as figure types, components, and labels are often missing, and (ii) most captions received low usefulness scores (1–2). This indicates that raw captions are insufficient for faithfully reconstructing scientific figures.

To quantify this further, a human annotator wrote reference descriptions for all 200 figures. We then compared these human-written descriptions against both the original captions and VLM-generated descriptions using BLEU-4~\citep{papineni-etal-2002-bleu}, ROUGE-L~\citep{lin-2004-rouge}, and Semantic Textual Similarity (STS)~\citep{reimers-gurevych-2019-sentence} (Figure~\ref{fig:dataset_quality_eval}, right). Across multiple VLMs (Qwen2.5-VL 7B/32B~\citep{bai2025qwen25vltechnicalreport}, InternVL3 8B/38B~\citep{zhu2025internvl3exploringadvancedtraining}, GPT-4o and GPT-4o-mini~\citep{openai2024gpt4ocard}), results show that VLMs produce richer and more faithful descriptions than raw captions. For example, GPT-4o reaches 0.089 BLEU-4 compared to just 0.003 for captions, and comes close to human-human agreement (0.094 BLEU-4). VLM outputs are also substantially longer (120–170 vs.\ 34 characters), indicating that they capture additional detail necessary for figure reconstruction. These results motivate our use of VLM-generated descriptions in DaTikZ-V4. For additional information, we refer to the Appendix~\ref{app:caption_quality_analysis}.

\section{Dataset}
\label{sec:Dataset}

Building on DaTikZ-V3, we introduce DaTikZ-V4, a significantly expanded and refined dataset designed to support the training and evaluation of Text-to-TikZ models. The development of DaTikZ-V4 addresses the growing need for both larger and higher-quality datasets, which are critical for surpassing not only proprietary state-of-the-art models like GPT-5 but also increasingly more capable open-source LLMs such as Qwen3.

\setlength{\columnsep}{10pt}
\begin{wraptable}[11]{r}{0.48\columnwidth}
\vspace{-\baselineskip}
\centering
\footnotesize
\caption{Unique TikZ graphics across all DaTikZ versions.}
\label{tab:dataset_sizes_new}
\setlength{\tabcolsep}{3pt}
\begin{tabular}{lrrrr}
\toprule
\textbf{Source} & \textbf{DaTikZ} & \textbf{V2} & \textbf{V3} & \textbf{V4}\\
\midrule
arXiv     &  85{,}656 & 326{,}450 & 407{,}851 & 1{,}471{,}083\\
GitHub    &       0   &       0   &       0   &   413{,}178\\
TeX SE    &  29{,}238 &  30{,}609 &  42{,}654 &     97{,}909\\
Synthetic &   1{,}957 &   1{,}958 &   2{,}256 &     13{,}514\\
Curated   &     981   &   1{,}566 &   3{,}646 &      5{,}196\\
\midrule
\textbf{Total} & 117{,}832 & 360{,}583 & 456{,}407 & 2{,}000{,}880\\
\bottomrule
\end{tabular}
\end{wraptable}

\paragraph{Data Sourcing}
To enhance dataset scale, we first identify GitHub as a valuable large-scale source of high-quality graphics programs. With over one billion repositories, GitHub hosts a wealth of educational resources, tutorials, theses, books, and personal projects, many of which contain TikZ code. From this, we clone approximately 5,500 repositories containing \texttt{.tex} or \texttt{.pgf} files with TikZ content, resulting in over 400,000 unique TikZ samples. This GitHub-only subset is nearly as large as the entirety of DaTikZ-V3. To further expand coverage, we also extend sourcing from arXiv by including data post-2021 to mid 2025. The increasing amounts of arXiv submissions each year allows us to source 1M additional samples, resulting in over 2M TikZ graphics in total. Of these, 35.55\% originate from sources under permissive Creative Commons licenses (e.g., CC-BY, CC-BY-SA, CC0) and can be redistributed. 40.03\% originate from sources under Nonexclusive-Distribution licenses, and the remaining 24.43\% contain no explicit license information. A comparison of DaTikZ-V4 to previous releases is seen in Table~\ref{tab:dataset_sizes_new}.

\begin{figure}[h]
  \centering
  \includegraphics[width=\textwidth]{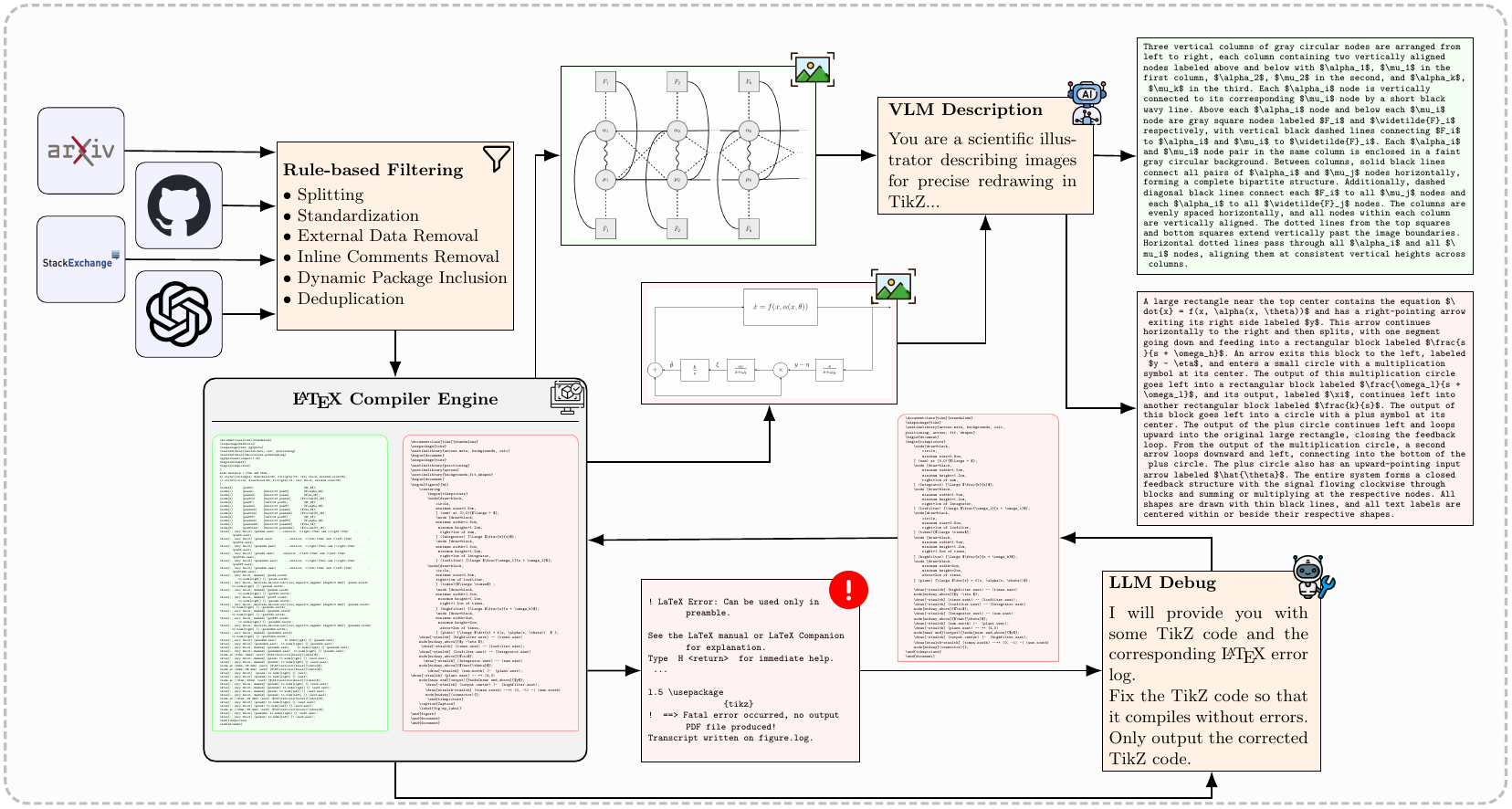}
  \caption{Overview of the data preprocessing workflow. We start by sourcing TikZ graphics programs primarily from arXiv, GitHub, TeX SE, as well as synthetic data. Next, rule-based filtering techniques are applied, and the TikZ code is rendered. Uncompilable code undergoes an iterative debugging process using LLMs alongside the error messages to attempt error correction. Finally, all compilable code images are described using VLMs.}
  \label{fig:tikz_workflow}
\end{figure}

\paragraph{Filtering}
Beyond traditional \texttt{tikzpicture} environments, we now extract from other environments such as \texttt{tikz-cd} (common in mathematical diagrams) and \texttt{circuitikz} (used in electronics). Since individual figures often contain multiple subfigures, we recursively split and extract all subfigure content. Furthermore, we enforce a standardized TikZ code by wrapping the code inside the \texttt{\textbackslash documentclass[tikz]\{standalone\}} environment. Additionally, we implement a dynamic package detection approach by using regular expressions to include necessary LaTeX packages (e.g., recognizing \texttt{circuitikz} from context such as \texttt{resistor}). We also remove any code that depends on external files (e.g., \texttt{\textbackslash input\{...\}}, \texttt{\textbackslash includegraphics\{...\}}), as well as all inline comments, to improve compilation rates and reduce noise. Lastly, we apply exact deduplication and dismiss all samples where the number of characters is both smaller than 100 and larger than 4000.

\paragraph{LLM Debugging}
Due to the low compilation success rate, especially from arXiv (success rate 31.3\%), we introduce an LLM-based debugging pipeline. Given a code snippet and its compiler error, an LLM is instructed to fix the TikZ code. Using Qwen-32B across our corpus of 1.3M uncompilable TikZ samples, we successfully repair 600K instances in the first pass. This approach substantially boosts the proportion of usable TikZ programs at scale.

\paragraph{VLM-based Image Description}
As shown in Section~\ref{sec:Caption Quality Analysis}, raw captions are often unhelpful for figure reproduction, potentially leading to severe hallucinations. To mitigate this, we employ VLMs to generate precise descriptions of TikZ figures. Using Qwen2.5-VL-7B-Instruct, we annotate around 1.3M compilable samples, producing the first large-scale dataset of TikZ paired with semantically rich textual descriptions, providing stronger supervision for downstream model training. An overview of our dataset construction is illustrated in Figure~\ref{fig:tikz_workflow}. For ablations and further details about prompts and frameworks, we refer to~\ref{app:dataset}.

\section{Method}
\label{sec:Method}

We train Text-to-TikZ models in two stages: SFT to ground models in TikZ syntax and task-specific token distributions, followed by RL for incorporating feedback from rendered images to enforce enhanced visual alignment~\citep{rodriguez2025renderingawarereinforcementlearningvector}. Similar two-stage paradigms have also proven effective in related domains such as code generation and mathematical reasoning, where surface-level syntax is complemented by execution-level accuracy~\citep{NEURIPS2022_8636419d, gehring2025rlefgroundingcodellms}.

\paragraph{Stage 1: Supervised Finetuning}
Given a figure description $x_{\text{desc}}$ and ground-truth TikZ sequence $x_{\text{tikz}} = (x_1,\dots,x_T)$, we minimize the standard autoregressive negative log-likelihood:
\begin{equation}
\mathcal{L}_{\text{SFT}}(\theta) = \mathbb{E}_{(x_{\text{desc}},\,x_{\text{tikz}}) \sim \mathcal{D}} \Bigg[ - \sum_{t=1}^{T} \log p_\theta(x_{t} \mid x_{<t},\, x_{\text{desc}}) \Bigg]
\label{eq:sft}
\end{equation}
This ensures syntactic validity and prompt alignment. At the same time, the model remains unaware of the rendered semantics of the figure, which leads to common errors such as loops, irrelevant content, or incorrect spatial relations.

\paragraph{Stage 2: Reinforcement Learning}
To address this, we reinterpret the SFT model $p_{\theta_{\text{SFT}}}$ as a stochastic policy and apply reinforcement learning with GRPO. For each description, $G$ rollouts $\{o_1,\ldots,o_G\} \sim p_{\theta_{\text{old}}}(\cdot \mid x_{\text{desc}})$ are sampled, each of which is assigned a scalar reward $\{r_1, \ldots, r_G\}$ scored by a reward model, and updated with group-centered advantages $A_i = \tfrac{r_i - \operatorname{mean}(\{r_j\})}{\operatorname{std}(\{r_j\})}$. The GRPO objective we maximize is:
\begin{align}
\mathcal{J}_{\text{GRPO}}(\theta) &=
\mathbb{E}_{\,x_{\text{desc}} \sim \mathcal{D}}
\Bigg[
  \frac{1}{LG}\sum_{i=1}^G\sum_{t=1}^{|o_i|}
  \min\!\Bigg(
    \frac{p_\theta(o_i \mid x_{\text{desc}})}{p_{\theta_{\text{old}}}(o_i \mid x_{\text{desc}})} A_i, \nonumber \\
&\qquad
    \operatorname{clip}\!\Big(
      \tfrac{p_\theta(o_i \mid x_{\text{desc}})}{p_{\theta_{\text{old}}}(o_i \mid x_{\text{desc}})},
      1-\epsilon_{\text{low}}, 1+\epsilon_{\text{high}}
    \Big) A_i
  \Bigg) \nonumber 
  - \beta\,D_{\text{KL}}\!\big(p_\theta \,\|\, p_{\theta_{\text{SFT}}}\big)
\Bigg]
\label{eq:grpo}
\end{align}
where $\beta$ regulates the KL penalty. We implement the “Dr.GRPO" variant~\citep{liu2025understandingr1zeroliketrainingcritical}, which replaces the response-level normalization by a token-level normalization with a constant divisor (the maximum completion length $L$). This removes the response length bias in TikZ sequences, where longer responses are under-penalized. Furthermore, we apply the “Clip-Higher" strategy from DAPO~\citep{yu2025dapoopensourcellmreinforcement}, which decouples the clipping threshold $\epsilon$ into $\epsilon_{\text{low}}$ and $\epsilon_{\text{high}}$. This allows more headroom for increasing the probability of low-probability exploration tokens (by raising $\epsilon_{\text{high}}$), while still preventing collapse of high-probability exploitation tokens (by keeping $\epsilon_{\text{low}}$ smaller). As in DAPO, we set $\epsilon_{low} = 0.2$ and $\epsilon_{high} = 0.28$. Additionally, we remove scaling the advantages by the standard deviation of the group rewards to not introduce a bias towards more or less difficult prompts, and mask all samples whose completion was cut by the length cap as we find that it increases training stability. Finally, we disable the KL coefficient ($\beta = 0$) and sample with \texttt{temperature=1.0} and \texttt{top\_p=0.9}.

\begin{figure}[h]
  \centering
  \includegraphics[width=\textwidth]{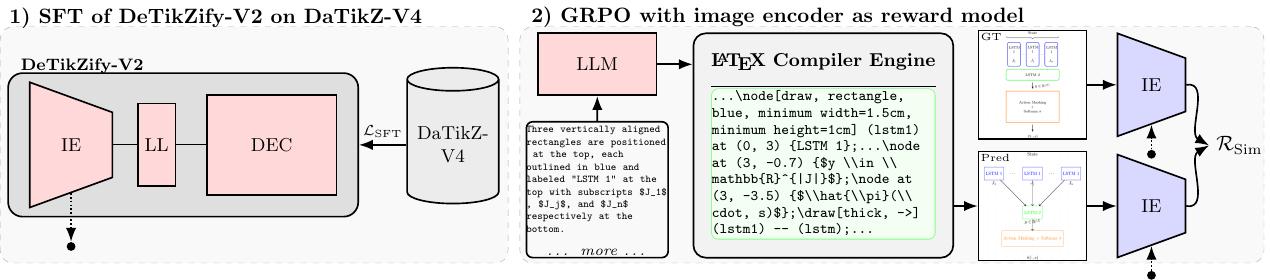}
  \caption{Overview of our post-SFT optimization steps. We first fully finetune DeTikZify-V2 consisting of an image encoder (IE), linear layer (LL) and LLM decoder (DEC) on our larger DaTikZ-V4 where we then use its enhanced IE to further finetune our LLMs based on the semantic similarity of the embeddings between ground truth and rendered image in an online RL setting using GRPO. The IE is kept frozen during RL optimization to mitigate reward hacking.}
  \label{fig:methodology_figure}
\end{figure}

\paragraph{Rewards}
Designing reward signals for Text-to-TikZ is challenging: they must capture faithfulness, scientific style, attributes, and spatial relations. Recent work has shown that metrics such as CLIPScore or DreamSim correlate poorly with human judgments as they fail to represent nuances in scientific figures~\citep{belouadi2025tikzerozeroshottextguidedgraphics} and are prone to reward hacking (e.g., embedding text into figures)~\citep{rodriguez2025renderingawarereinforcementlearningvector}. 

To the best of our knowledge, we propose the first domain-specific reward model for Text-to-TikZ. It builds on the image encoder of DeTikZify-V2~\citep{belouadi2024detikzify}, originally trained on DaTikZ-V3 for inverse graphics (image → TikZ). DeTikZify consists of an image encoder followed by a linear layer and an LLM decoder. By keeping the image encoder unfrozen during training, it incidentally learns to generate good low-dimensional representations of scientific figures in order to accurately reproduce the figure, allowing us to utilize it to measure semantic similarity between the embeddings of two scientific figures more accurately. With DaTikZ-V4 providing a much larger dataset, we retrain DeTikZify-V2 end-to-end, yielding a stronger encoder that produces richer, more generalizable embeddings of scientific diagrams. Subsequently, we use the retrained image encoder as our reward model in an online RL environment with GRPO. Both steps are illustrated in Figure~\ref{fig:methodology_figure}. Training details are provided in~\ref{app:method}.

For reward computation, pooled cosine similarity is not available since DeTikZify-V2 outputs patch-level embeddings. We therefore adopt an Earth Mover’s Distance (EMD)~\citep{710701, 10.5555/3045118.3045221} formulation, inspired by test-time scaling approaches in TikZero~\citep{belouadi2025tikzerozeroshottextguidedgraphics}. Given patch embeddings $\mathbf{x} = \{x_i\}_{i=1}^{|\mathbf{x}|}$ and $\mathbf{y} = \{y_j\}_{j=1}^{|\mathbf{y}|}$ from ground truth and predicted images, with distance matrix $D_{i,j} = 1 - \cos(x_i, y_j)$, the similarity reward is defined as
\begin{equation}
\mathcal{R}_{\text{Sim}}(\mathbf{x}, \mathbf{y}) \;=\;
1 - \frac{\sum_{i=1}^{|\mathbf{x}|} \sum_{j=1}^{|\mathbf{y}|} F_{i,j} D_{i,j}}
           {\sum_{i=1}^{|\mathbf{x}|} \sum_{j=1}^{|\mathbf{y}|} F_{i,j}},
\label{eq:sim}
\end{equation}
where $F \in \mathbb{R}^{|\mathbf{x}| \times |\mathbf{y}|}_{\geq 0}$ is the optimal flow matrix that minimizes the transport cost, subject to $\sum_i F_{i,j} = 1/|\mathbf{y}|$ and $\sum_j F_{i,j} = 1/|\mathbf{x}|$. This formulation yields a scalar reward in $[0,1]$ capturing semantic alignment. Finally, we add a format reward to ensure that the TikZ code starts and ends with valid document environments (i.e., \texttt{\textbackslash documentclass[tikz]\{standalone\}}, followed by \texttt{\textbackslash begin\{document\}}, and ending with \texttt{\textbackslash end\{document\}}). Non-conforming outputs receive a reward of zero.

\section{Experiments}
\label{sec:Experiments}

\paragraph{Experimental Setup}
For evaluation, we construct a contamination-free test set of 1,047 samples from DaTikZ-V4. To prevent overlap with training data, we (i) restrict to post–May 2025 samples, (ii) enforce per-source uniqueness (e.g., one figure per arXiv paper or GitHub repo, removing the rest from training), (iii) filter with n-gram matching~\citep{DBLP:journals/corr/abs-2303-08774}, and (iv) manual inspection to discard trivial or corrupted figures. To avoid model bias, all test descriptions are generated by GPT-4o. For RL-tuning, we create DaTikZ-V4-RL, a 160K-sample subset obtained by repairing uncompilable figures via a second LLM debugging step and re-describing them with Qwen2.5-VL-7B. This provides additional high-quality pairs beyond the training split.

\paragraph{Models}
We benchmark nine LLMs: (i) proprietary GPT-5\footnote{\url{https://openai.com/index/gpt-5-system-card/}} and GPT-4o~\citep{openai2024gpt4ocard}, (ii) open-source Qwen3 (32B, 8B), Qwen3-Coder-30B-A3B~\citep{yang2025qwen3technicalreport}, Qwen2.5 (14B, 3B)~\citep{qwen2025qwen25technicalreport}, TikZero-Plus-10B~\citep{belouadi2025tikzerozeroshottextguidedgraphics}, and Llama3.1-8B~\citep{grattafiori2024llama3herdmodels}, and (iii) our fine-tuned Qwen2.5-3B and Qwen3-8B models. We refer to our trained models as TikZilla, with the following variants: TikZilla-3B and TikZilla-8B (SFT only), and TikZilla-3B-RL and TikZilla-8B-RL (two-stage training). In addition, we also test RL-only training.

\paragraph{Evaluation Metrics}
We evaluate along four axes: (i) CLIPScore (CLIP)~\citep{hessel-etal-2021-clipscore} for text–image alignment, (ii) DreamSIM (DSim)~\citep{10.5555/3666122.3668330} for perceptual fidelity, (iii) TeX Edit Distance (TED)~\citep{10.5555/3045118.3045221} for code similarity, and (iv) Compilation Rate (CR) for executability. We also report average tokens (AT) for efficiency. To avoid reward–metric coupling in our RL ablations, we additionally report DINOScore (DINO)~\citep{caron2021emerging} and LPIPS~\citep{zhang2018unreasonable}, which are independent of our domain-specific reward model. An aggregate score (AVG) is computed as the mean of CLIP/DINO, DSim/LPIPS, and 1-TED (depending on the evaluation setting). Additional details are reported in~\ref{app:experiments}.

\section{Results}
\label{sec:Results}

\paragraph{Main Results}
Table~\ref{tab:main_results} reports results on automatic metrics. Our SFT+RL-tuned Qwen models achieve the best AVG performance, with TikZilla-3B-RL reaching 0.385 and TikZilla-8B-RL 0.384. Both surpass GPT-5 (0.365), despite it being recently released as one of the strongest reasoning LLMs, evaluated with no output length restrictions. Compared to the recently released TikZero-Plus-10B, TikZilla-3B-RL improves by +0.085 on CLIP and +0.334 on DSim, while achieving a 37\% higher compilation rate and requiring 261 fewer tokens on average. Similar improvements hold for TikZilla-8B-RL. These results highlight the effectiveness of our two-stage training process, combining high-quality data with a domain-specific reward model. For qualitative examples with TikZ code, we refer to the Appendix (Figure~\ref{fig:tikz_code_example_old_old},~\ref{fig:tikz_code_example_old},~\ref{fig:tikz_code_example}, and~\ref{fig:tikz_code_example_new}).

\begin{table}[t]
\centering
\caption{Results of all models on the evaluation subset of DaTikZ-V4. Both of our models trained with SFT and RL perform best, while GPT-5 and Qwen3-32B are the best proprietary and open-source LLMs respectively. \textbf{Bold} denotes best-performing while \underline{underline} is second-best.}
\label{tab:main_results}
\setlength{\tabcolsep}{4pt}
\begin{tabular}{lcccccc}
\toprule
\textbf{LLM} & 
\textbf{CLIP}$\uparrow$ & 
\textbf{DSim}$\uparrow$ &  
\textbf{TED}$\downarrow$ & 
\textbf{AVG}$\uparrow$ & 
\textbf{CR}$\uparrow$ & 
\textbf{AT}$\downarrow$ \\
\midrule
GPT-5 & 0.181 & 0.679 & \underline{0.765} & 0.365 & 88\% & 480 \\
GPT-4o & 0.147 & 0.580 & 0.767 & 0.320 & 78\% & 404 \\
\midrule
Qwen3-32B & 0.149 & 0.583 & \underline{0.765} & 0.322 & 79\% & 416 \\
Qwen3-Coder-30B-A3B & 0.140 & 0.566 & 0.778 & 0.309 & 77\% & 472 \\
Qwen2.5-14B & 0.132 & 0.511 & \underline{0.765} & 0.293 & 71\% & 376 \\
TikZero-Plus-10B & 0.104 & 0.397 & 0.807 & 0.231 & 61\% & 742 \\
Llama3.1-8B & 0.088 & 0.339 & 0.786 & 0.214 & 50\% & 529 \\
\midrule
Qwen2.5-3B & 0.081 & 0.315 & 0.789 & 0.202 & 52\% & 387 \\
Qwen2.5-3B (+RL) & 0.098 & 0.505 & 0.795 & 0.269 & \textbf{98\%} & \textbf{234} \\
TikZilla-3B & 0.161 & 0.613 & 0.802 & 0.324 & 89\% & 672 \\
TikZilla-3B-RL & \textbf{0.189} & \textbf{0.731} & 0.766 & \textbf{0.385} & \textbf{98\%} & 481 \\
\midrule
Qwen3-8B & 0.106 & 0.421 & 0.775 & 0.251 & 63\% & 412 \\
Qwen3-8B (+RL) & 0.169 & 0.669 & 0.768 & 0.357 & \textbf{98\%} & \underline{393} \\
TikZilla-8B & 0.158 & 0.602 & 0.793 & 0.322 & 86\% & 729 \\
TikZilla-8B-RL & \underline{0.185} & \underline{0.727} & \textbf{0.761} & \underline{0.384} & \underline{95\%} & 459 \\
\bottomrule
\end{tabular}
\end{table}

\subparagraph{Model Size and Training Regime}
Interestingly, the smaller Qwen2.5-3B not only closes the gap with Qwen3-8B but even slightly outperforms it once trained with SFT+RL. However, its low baseline (0.202) indicates that it strongly relies on SFT before RL, whereas Qwen3-8B benefits from RL directly (0.251 → 0.357). This suggests that SFT primarily provides syntax grounding for smaller models, while larger models already encode some TikZ knowledge that RL can amplify.

\subparagraph{Implicit Efficiency Effects}
RL consistently improves compilation rates to 95–98\% and reduces token length, indicating more efficient code generation. Unlike prior SVG studies~\citep{rodriguez2025renderingawarereinforcementlearningvector}, which required explicit code efficiency rewards, we observe a natural reduction in sequence length. We hypothesize this stems from our semantic reward model penalizing hallucinated or redundant elements, indirectly encouraging conciseness. A deeper comparison with explicit efficiency rewards is left for future work.

\paragraph{Human Evaluation}
We conduct a human evaluation with 9 expert annotators (6 PhD, 2 postdoc, 1 faculty member). Each annotator rated 30 randomized figures/descriptions across 4–5 models, using a 1–5 Likert scale (1 = uncompilable, 5 = publication-ready). Two criteria were considered: (i) textual alignment (does the output follow the provided description?) and (ii) image alignment (does the output match the original ground-truth figure?). Annotator agreement was strong (Cohen’s $\kappa = 0.814$ for text, 0.794 for image). Full details are provided in~\ref{app:results}.

\begin{figure}[h]
  \centering
  \includegraphics[width=0.8\textwidth]{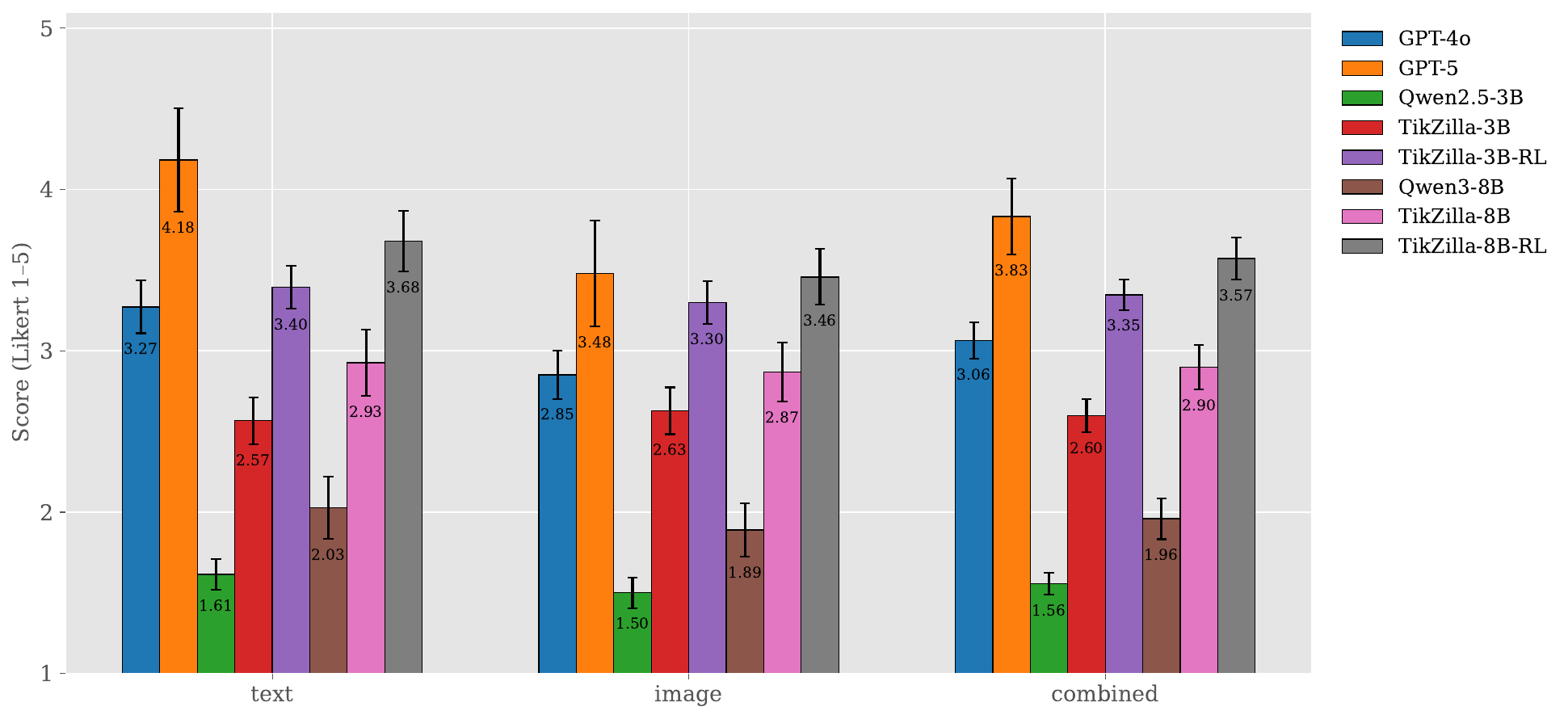}
  \caption{Average Likert-scale ratings (1–5, higher is better) with 95\% confidence intervals for eight LLMs, evaluated under two settings: (i) alignment with textual descriptions and (ii) alignment with ground-truth images. Combined scores are shown as the average of both settings.}
  \label{fig:human_scores}
\end{figure}

\subparagraph{Results}
Figure~\ref{fig:human_scores} shows that GPT-5 achieved the highest textual score (4.18) and tied with our TikZilla-8B-RL on image evaluation (3.48 vs. 3.46). TikZilla-3B-RL also performed competitively (3.40 text, 3.30 image). Reinforcement learning substantially boosted both Qwen models (+0.75 and +0.67 points), while base models lagged 1.5–2 points behind. Interestingly, most models (especially GPT-5) scored higher on the textual evaluation than on the image evaluation. We hypothesize two possible explanations: (i) if VLM-generated captions omit or misrepresent visual details, models may score highly on textual alignment (satisfying the description) but lower on image alignment (failing to match the true figure). (ii) Human annotators may apply stricter criteria when comparing against ground-truth images than when comparing against text. Disentangling these two factors remains an open question, which we leave for future work.

\subparagraph{Correlation with Metrics}
Finally, we compute correlations between automatic metrics and human scores using Spearman’s $\rho$. CLIP ($\rho_{CLIP}=0.260$) and TED ($\rho_{1-TED}=0.307$) show weak, DSim moderate ($\rho_{DSim}=0.586$), and our reward model strong ($\rho_{\mathcal{R}_{Sim}}=0.714$) correlation. This validates our design of a domain-specific reward model aligned with human judgment.

\begin{figure}[h]
  \centering
  \includegraphics[width=0.7\textwidth]{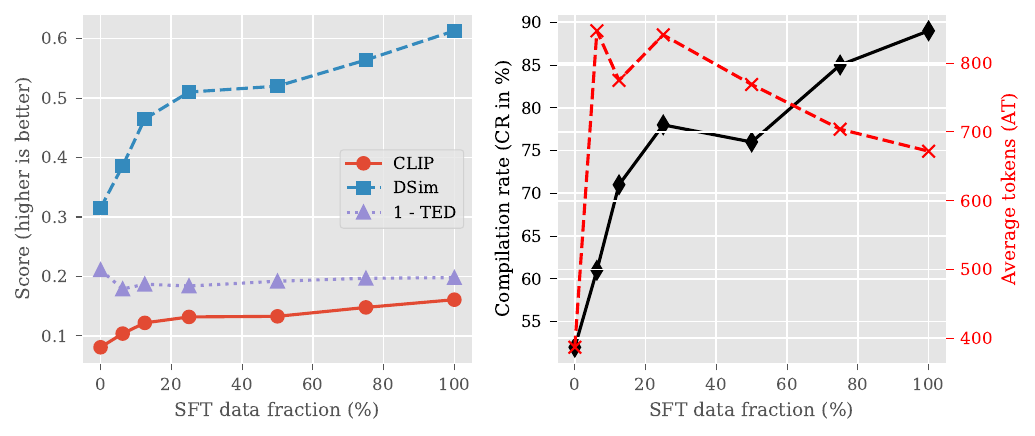}
  \caption{SFT on Qwen2.5-3B with different dataset scales (75\%, 50\%, 25\%, 12.5\%, and 6.25\%).}
  \label{fig:ablation_dataset_sizes}
\end{figure}

\paragraph{Ablations}
We perform a series of ablations to isolate the contribution of each component in our pipeline. These include: (i) the impact of input quality and LLM-based debugging (Table~\ref{tab:ablations_combined}), (ii) the effect of different reward models (Table~\ref{tab:ablation_reward_model}), and (iii) the influence of dataset scale (Figure~\ref{fig:ablation_dataset_sizes}). We additionally evaluate TikZilla in an out-of-distribution (OOD) setting using the SPIQA~\citep{pramanick2024spiqa} benchmark (Table~\ref{tab:main_results_spiqa_ood}).

\subparagraph{Captions vs. Descriptions} 
VLM-generated descriptions consistently outperform raw captions. At inference, GPT-4o achieves 0.315 AVG with descriptions versus 0.270 with captions, confirming our earlier analysis that captions are often unhelpful for figure reproduction. Examples are shown in Figure~\ref{fig:caption_versus_description} in the Appendix. For training, Qwen2.5-3B also benefits from descriptions (0.289 vs.\ 0.279), though the gap is smaller, likely due to the limited caption subset (468k samples). Mixing captions/descriptions (desc.\ $\vee$ cap.) and oversampling descriptions with captions (desc.\ + cap.) degrade performance, suggesting that low-quality captions dilute training even when more data is added.

\subparagraph{LLM-based Debugging}
Models trained only on first-try compilable code perform considerably worse than those trained on the full dataset (0.288 vs.\ 0.324), highlighting the necessity of our LLM-based debugging pipeline to increase the size of our usable TikZ corpus.

\begin{table}[t]
\centering
\caption{Ablations on input data quality and debugging. VLM-based descriptions consistently outperform captions, while mixing or oversampling captions brings no gains. Our LLM-based debugging step yield improvements.}
\label{tab:ablations_combined}
\setlength{\tabcolsep}{4pt}
\begin{tabular}{lcccccc}
\toprule
\textbf{LLM} & 
\textbf{CLIP}$\uparrow$ & 
\textbf{DSim}$\uparrow$ &  
\textbf{TED}$\downarrow$ & 
\textbf{AVG}$\uparrow$ & 
\textbf{CR}$\uparrow$ & 
\textbf{AT}$\downarrow$ \\
\midrule
GPT-4o\textsubscript{cap.} & 0.105 & 0.469 & \textbf{0.763} & 0.270 & 80\% & \textbf{337} \\
GPT-4o\textsubscript{desc.} & 0.143 & 0.568 & \underline{0.767} & 0.315 & 76\% & \underline{416} \\
\midrule
Qwen2.5-3B (+SFT\textsubscript{cap.}) & 0.134 & 0.511 & 0.809 & 0.279 & 79\% & 768 \\
Qwen2.5-3B (+SFT\textsubscript{desc.}) & 0.141 & 0.530 & 0.805 & 0.289 & \underline{85\%} & 651 \\
\midrule
Qwen2.5-3B (+SFT\textsubscript{desc. $\vee$ cap.}) & 0.154 & 0.589 & 0.804 & 0.313 & \underline{85\%} & 735 \\
Qwen2.5-3B (+SFT\textsubscript{desc. + cap.}) & \underline{0.157} & \underline{0.599} & 0.799 & \underline{0.319} & \textbf{89\%} & 787 \\
\midrule
Qwen2.5-3B (+SFT\textsubscript{no debug}) & 0.138 & 0.534 & 0.809 & 0.288 & 79\% & 762 \\
TikZilla-3B & \textbf{0.161} & \textbf{0.613} & 0.802 & \textbf{0.324} & \textbf{89\%} & 672 \\
\bottomrule
\end{tabular}
\end{table}

\subparagraph{Reward Model Training}
We compare our domain-specific reward $\mathcal{R}_{\text{Sim}}$ against CLIPScore (image–image) and DreamSIM (Table~\ref{tab:ablation_reward_model}). We find that all reward functions improve over the SFT baseline, but $\mathcal{R}_{\text{Sim}}$ achieves the strongest AVG performance. DreamSIM performs slightly better than CLIPScore. Retraining DeTikZify-V2 on DaTikZ-V4 yields a stronger reward model (0.389 vs.\ 0.375). Correlations with human judgments also improve ($\rho_{\mathcal{R}_{Sim}} = 0.714$ vs.\ 0.698), confirming that larger-scale scientific data produces more reliable image encoders for semantic evaluation.

\begin{table}[t]
\centering
\caption{Ablation of our domain-specific reward models compared to CLIP\textsubscript{Img} and DreamSIM. The DaTikZ-V4 trained encoder achieves the strongest AVG performance.}
\label{tab:ablation_reward_model}
\setlength{\tabcolsep}{4pt}
\begin{tabular}{lcccccc}
\toprule
\textbf{LLM} & 
\textbf{DINO}$\uparrow$ & 
\textbf{LPIPS}$\uparrow$ &  
\textbf{TED}$\downarrow$ & 
\textbf{AVG}$\uparrow$ & 
\textbf{CR}$\uparrow$ & 
\textbf{AT}$\downarrow$ \\
\midrule
Qwen2.5-3B (+SFT+RL\textsubscript{CLIP\textsubscript{Img.}}) & 0.751 & 0.418 & 0.779 & 0.463 & 97\% & 537 \\
Qwen2.5-3B (+SFT+RL\textsubscript{DSim}) & 0.759  & 0.439 & 0.777 & 0.474 & \textbf{99\%} & \underline{494} \\
Qwen2.5-3B (+SFT+RL)\textsubscript{$\mathcal{R}_{\text{Sim}}$(DaTikZ-V3)} & \underline{0.789} & \underline{0.440} & \underline{0.768} & \underline{0.487} & 97\% & 496 \\
TikZilla-3B-RL & \textbf{0.809} & \textbf{0.451} & \textbf{0.766} & \textbf{0.498} & \underline{98\%} & \textbf{481} \\
\bottomrule
\end{tabular}
\end{table}

\subparagraph{Dataset Sizes}
To understand how performance scales with data, we supervised fine-tune Qwen2.5-3B on subsets of DaTikZ-V4 at 75\%, 50\%, 25\%, 12.5\%, and 6.25\% of the full dataset (Figure~\ref{fig:ablation_dataset_sizes}). Performance increases sharply at small data scales (from 0–25\%), after which improvements become more gradual from 25\% up to the full dataset, suggesting that further data scaling (e.g., synthetic data) remains a promising direction for improving text-to-TikZ generation.

\subparagraph{OOD Data}
To assess TikZilla's robustness under distribution shift, we evaluate it on the SPIQA dataset. SPIQA figures are typically not generated in TikZ but originate from tools such as matplotlib, ggplot2, and MATLAB, often containing multi-panel layouts, overlays, and varied diagrammatic structures. This makes SPIQA a meaningful OOD benchmark from a structural-complexity perspective. For evaluation, we use all samples only including figures from the test-A and test-B splits and generate textual descriptions with GPT-4o, yielding 397 test cases. Since ground-truth TikZ code is unavailable, we omit TED in this evaluation. Relative to the DaTikZ-V4 test split (Table~\ref{tab:main_results}), SPIQA exhibits substantially longer sequences, lower compilation rates, and overall lower performance, as expected due to its non-TikZ origin and higher visual complexity. Notably, both TikZilla-8B-RL and especially TikZilla-3B RL outperform GPT-5 on this OOD benchmark.

\begin{table}[t]
\centering
\caption{Model performance on the SPIQA benchmark.}
\label{tab:main_results_spiqa_ood}
\setlength{\tabcolsep}{4pt}
\begin{tabular}{lccccc}
\toprule
\textbf{LLM} & 
\textbf{CLIP}$\uparrow$ & 
\textbf{DSim}$\uparrow$ &  
\textbf{CR}$\uparrow$ & 
\textbf{AT}$\downarrow$ \\
\midrule
GPT-5 & 0.115 & 0.432 & 60\% & 1239 \\
GPT-4o & 0.098 & 0.326 & 48\% & \textbf{748} \\
\midrule
Qwen3-Coder-30B-A3B & 0.102 & 0.349 & 58\% & 1000 \\
\midrule
Qwen2.5-3B & 0.038 & 0.117 & 24\% & 772 \\
TikZilla-3B & 0.114 & 0.374 & 64\% & 1170 \\
TikZilla-3B-RL & \textbf{0.193} & \textbf{0.637} & \textbf{97\%} & \underline{765} \\
\midrule
Qwen3-8B & 0.070 & 0.228 & 37\% & 781 \\
TikZilla-8B & 0.131 & 0.428 & 70\% & 1184 \\
TikZilla-8B-RL & \underline{0.174} & \underline{0.584} & \underline{90\%} & 809 \\
\bottomrule
\end{tabular}
\end{table}

\section{Conclusion, Limitations, and Future Work}
\label{sec:Conclusion, Limiatations, and Future Work}

We presented DaTikZ-V4, a large-scale, high-quality dataset for Text-to-TikZ, and a two-stage training framework combining SFT with RL. Our key contributions are a richer dataset sourced from arXiv and GitHub with LLM-based debugging to improve compilability, VLM-generated descriptions that overcome the low quality of raw captions, and a domain-specific reward model derived from an inverse-graphics image encoder, which correlates strongly with human judgments of figure quality. Building on these components, we introduced TikZilla, a family of small Qwen-based models that achieve near-perfect compilation rates and even surpass much larger commercial systems such as GPT-4o across automatic and human evaluation. Beyond technical performance, TikZilla demonstrates the feasibility of building reproducible and efficient text-to-image generation systems with small-scale open models, reducing reliance on costly proprietary solutions.

A key limitation is that our figure descriptions are generated automatically by VLMs, which may introduce omissions or hallucinations. This can bias training and, in rare cases, reward optimization may reinforce errors when descriptions diverge from figures. More reliable annotation methods and fine-grained reward functions are therefore crucial directions for future work. Beyond addressing these issues, future work should focus on designing automatic metrics with stronger alignment to human perception, and extending our approach to other structured generation tasks (e.g., LaTeX tables, CAD, or flowcharts), where programmatic fidelity is critical.

\bibliography{iclr2026_conference}

@misc{eger2025transformingsciencelargelanguage,
      title={Transforming Science with Large Language Models: A Survey on AI-assisted Scientific Discovery, Experimentation, Content Generation, and Evaluation}, 
      author={Steffen Eger and Yong Cao and Jennifer D'Souza and Andreas Geiger and Christian Greisinger and Stephanie Gross and Yufang Hou and Brigitte Krenn and Anne Lauscher and Yizhi Li and Chenghua Lin and Nafise Sadat Moosavi and Wei Zhao and Tristan Miller},
      year={2025},
      eprint={2502.05151},
      archivePrefix={arXiv},
      primaryClass={cs.CL},
      url={https://arxiv.org/abs/2502.05151}, 
}

@inproceedings{bi-etal-2024-ai,
    title = "{AI} for Science in the Era of Large Language Models",
    author = "Bi, Zhenyu  and
      Xu, Minghao  and
      Tang, Jian  and
      Wang, Xuan",
    editor = "Li, Jessy  and
      Liu, Fei",
    booktitle = "Proceedings of the 2024 Conference on Empirical Methods in Natural Language Processing: Tutorial Abstracts",
    month = nov,
    year = "2024",
    address = "Miami, Florida, USA",
    publisher = "Association for Computational Linguistics",
    url = "https://aclanthology.org/2024.emnlp-tutorials.5/",
    doi = "10.18653/v1/2024.emnlp-tutorials.5",
    pages = "32--38"
}

@misc{minaee2025largelanguagemodelssurvey,
      title={Large Language Models: A Survey}, 
      author={Shervin Minaee and Tomas Mikolov and Narjes Nikzad and Meysam Chenaghlu and Richard Socher and Xavier Amatriain and Jianfeng Gao},
      year={2025},
      eprint={2402.06196},
      archivePrefix={arXiv},
      primaryClass={cs.CL},
      url={https://arxiv.org/abs/2402.06196}, 
}

@misc{wu2023multimodallargelanguagemodels,
      title={Multimodal Large Language Models: A Survey}, 
      author={Jiayang Wu and Wensheng Gan and Zefeng Chen and Shicheng Wan and Philip S. Yu},
      year={2023},
      eprint={2311.13165},
      archivePrefix={arXiv},
      primaryClass={cs.AI},
      url={https://arxiv.org/abs/2311.13165}, 
}

@misc{gottweis2025aicoscientist,
      title={Towards an AI co-scientist}, 
      author={Juraj Gottweis and Wei-Hung Weng and Alexander Daryin and Tao Tu and Anil Palepu and Petar Sirkovic and Artiom Myaskovsky and Felix Weissenberger and Keran Rong and Ryutaro Tanno and Khaled Saab and Dan Popovici and Jacob Blum and Fan Zhang and Katherine Chou and Avinatan Hassidim and Burak Gokturk and Amin Vahdat and Pushmeet Kohli and Yossi Matias and Andrew Carroll and Kavita Kulkarni and Nenad Tomasev and Yuan Guan and Vikram Dhillon and Eeshit Dhaval Vaishnav and Byron Lee and Tiago R D Costa and José R Penadés and Gary Peltz and Yunhan Xu and Annalisa Pawlosky and Alan Karthikesalingam and Vivek Natarajan},
      year={2025},
      eprint={2502.18864},
      archivePrefix={arXiv},
      primaryClass={cs.AI},
      url={https://arxiv.org/abs/2502.18864}, 
}

@misc{lu2024aiscientistfullyautomated,
      title={The AI Scientist: Towards Fully Automated Open-Ended Scientific Discovery}, 
      author={Chris Lu and Cong Lu and Robert Tjarko Lange and Jakob Foerster and Jeff Clune and David Ha},
      year={2024},
      eprint={2408.06292},
      archivePrefix={arXiv},
      primaryClass={cs.AI},
      url={https://arxiv.org/abs/2408.06292}, 
}

@article{rodriguez2023figgen,
  title={Figgen: Text to scientific figure generation},
  author={Rodriguez, Juan A and Vazquez, David and Laradji, Issam and Pedersoli, Marco and Rodriguez, Pau},
  journal={arXiv preprint arXiv:2306.00800},
  year={2023}
}

@article{zou2024vgbench,
  title={Vgbench: Evaluating large language models on vector graphics understanding and generation},
  author={Zou, Bocheng and Cai, Mu and Zhang, Jianrui and Lee, Yong Jae},
  journal={arXiv preprint arXiv:2407.10972},
  year={2024}
}

@inproceedings{belouadi2024automatikz,
  title={{AutomaTikZ}: Text-Guided Synthesis of Scientific Vector Graphics with {TikZ}},
  author={Jonas Belouadi and Anne Lauscher and Steffen Eger},
  booktitle={The Twelfth International Conference on Learning Representations},
  year={2024},
  url={https://openreview.net/forum?id=v3K5TVP8kZ}
}

@inproceedings{belouadi2024detikzify,
    title={{DeTikZify}: Synthesizing Graphics Programs for Scientific Figures and Sketches with {TikZ}},
    author={Jonas Belouadi and Simone Paolo Ponzetto and Steffen Eger},
    booktitle={The Thirty-eighth Annual Conference on Neural Information Processing Systems},
    year={2024},
    url={https://openreview.net/forum?id=bcVLFQCOjc}
}

@misc{belouadi2025tikzerozeroshottextguidedgraphics,
      title={TikZero: Zero-Shot Text-Guided Graphics Program Synthesis}, 
      author={Jonas Belouadi and Eddy Ilg and Margret Keuper and Hideki Tanaka and Masao Utiyama and Raj Dabre and Steffen Eger and Simone Paolo Ponzetto},
      year={2025},
      eprint={2503.11509},
      archivePrefix={arXiv},
      primaryClass={cs.CL},
      url={https://arxiv.org/abs/2503.11509}, 
}

@article{Lopes2019ALR,
  title={A Learned Representation for Scalable Vector Graphics},
  author={Raphael Gontijo Lopes and David R Ha and Douglas Eck and Jonathon Shlens},
  journal={2019 IEEE/CVF International Conference on Computer Vision (ICCV)},
  year={2019},
  pages={7929-7938},
  url={https://api.semanticscholar.org/CorpusID:102353397}
}

@inproceedings{NEURIPS2020_bcf9d6bd,
 author = {Carlier, Alexandre and Danelljan, Martin and Alahi, Alexandre and Timofte, Radu},
 booktitle = {Advances in Neural Information Processing Systems},
 editor = {H. Larochelle and M. Ranzato and R. Hadsell and M.F. Balcan and H. Lin},
 pages = {16351--16361},
 publisher = {Curran Associates, Inc.},
 title = {DeepSVG: A Hierarchical Generative Network for Vector Graphics Animation},
 url = {https://proceedings.neurips.cc/paper_files/paper/2020/file/bcf9d6bd14a2095866ce8c950b702341-Paper.pdf},
 volume = {33},
 year = {2020}
}

@misc{rodriguez2024starvectorgeneratingscalablevector,
      title={StarVector: Generating Scalable Vector Graphics Code from Images and Text}, 
      author={Juan A. Rodriguez and Abhay Puri and Shubham Agarwal and Issam H. Laradji and Pau Rodriguez and Sai Rajeswar and David Vazquez and Christopher Pal and Marco Pedersoli},
      year={2024},
      eprint={2312.11556},
      archivePrefix={arXiv},
      primaryClass={cs.CV},
      url={https://arxiv.org/abs/2312.11556}, 
}

@misc{openai2024gpt4ocard,
      title={GPT-4o System Card}, 
      author={OpenAI and : and Aaron Hurst and Adam Lerer and Adam P. Goucher and Adam Perelman and Aditya Ramesh and Aidan Clark and AJ Ostrow and Akila Welihinda and Alan Hayes and Alec Radford and Aleksander Mądry and Alex Baker-Whitcomb and Alex Beutel and Alex Borzunov and Alex Carney and Alex Chow and Alex Kirillov and Alex Nichol and Alex Paino and Alex Renzin and Alex Tachard Passos and Alexander Kirillov and Alexi Christakis and Alexis Conneau and Ali Kamali and Allan Jabri and Allison Moyer and Allison Tam and Amadou Crookes and Amin Tootoochian and Amin Tootoonchian and Ananya Kumar and Andrea Vallone and Andrej Karpathy and Andrew Braunstein and Andrew Cann and Andrew Codispoti and Andrew Galu and Andrew Kondrich and Andrew Tulloch and Andrey Mishchenko and Angela Baek and Angela Jiang and Antoine Pelisse and Antonia Woodford and Anuj Gosalia and Arka Dhar and Ashley Pantuliano and Avi Nayak and Avital Oliver and Barret Zoph and Behrooz Ghorbani and Ben Leimberger and Ben Rossen and Ben Sokolowsky and Ben Wang and Benjamin Zweig and Beth Hoover and Blake Samic and Bob McGrew and Bobby Spero and Bogo Giertler and Bowen Cheng and Brad Lightcap and Brandon Walkin and Brendan Quinn and Brian Guarraci and Brian Hsu and Bright Kellogg and Brydon Eastman and Camillo Lugaresi and Carroll Wainwright and Cary Bassin and Cary Hudson and Casey Chu and Chad Nelson and Chak Li and Chan Jun Shern and Channing Conger and Charlotte Barette and Chelsea Voss and Chen Ding and Cheng Lu and Chong Zhang and Chris Beaumont and Chris Hallacy and Chris Koch and Christian Gibson and Christina Kim and Christine Choi and Christine McLeavey and Christopher Hesse and Claudia Fischer and Clemens Winter and Coley Czarnecki and Colin Jarvis and Colin Wei and Constantin Koumouzelis and Dane Sherburn and Daniel Kappler and Daniel Levin and Daniel Levy and David Carr and David Farhi and David Mely and David Robinson and David Sasaki and Denny Jin and Dev Valladares and Dimitris Tsipras and Doug Li and Duc Phong Nguyen and Duncan Findlay and Edede Oiwoh and Edmund Wong and Ehsan Asdar and Elizabeth Proehl and Elizabeth Yang and Eric Antonow and Eric Kramer and Eric Peterson and Eric Sigler and Eric Wallace and Eugene Brevdo and Evan Mays and Farzad Khorasani and Felipe Petroski Such and Filippo Raso and Francis Zhang and Fred von Lohmann and Freddie Sulit and Gabriel Goh and Gene Oden and Geoff Salmon and Giulio Starace and Greg Brockman and Hadi Salman and Haiming Bao and Haitang Hu and Hannah Wong and Haoyu Wang and Heather Schmidt and Heather Whitney and Heewoo Jun and Hendrik Kirchner and Henrique Ponde de Oliveira Pinto and Hongyu Ren and Huiwen Chang and Hyung Won Chung and Ian Kivlichan and Ian O'Connell and Ian O'Connell and Ian Osband and Ian Silber and Ian Sohl and Ibrahim Okuyucu and Ikai Lan and Ilya Kostrikov and Ilya Sutskever and Ingmar Kanitscheider and Ishaan Gulrajani and Jacob Coxon and Jacob Menick and Jakub Pachocki and James Aung and James Betker and James Crooks and James Lennon and Jamie Kiros and Jan Leike and Jane Park and Jason Kwon and Jason Phang and Jason Teplitz and Jason Wei and Jason Wolfe and Jay Chen and Jeff Harris and Jenia Varavva and Jessica Gan Lee and Jessica Shieh and Ji Lin and Jiahui Yu and Jiayi Weng and Jie Tang and Jieqi Yu and Joanne Jang and Joaquin Quinonero Candela and Joe Beutler and Joe Landers and Joel Parish and Johannes Heidecke and John Schulman and Jonathan Lachman and Jonathan McKay and Jonathan Uesato and Jonathan Ward and Jong Wook Kim and Joost Huizinga and Jordan Sitkin and Jos Kraaijeveld and Josh Gross and Josh Kaplan and Josh Snyder and Joshua Achiam and Joy Jiao and Joyce Lee and Juntang Zhuang and Justyn Harriman and Kai Fricke and Kai Hayashi and Karan Singhal and Katy Shi and Kavin Karthik and Kayla Wood and Kendra Rimbach and Kenny Hsu and Kenny Nguyen and Keren Gu-Lemberg and Kevin Button and Kevin Liu and Kiel Howe and Krithika Muthukumar and Kyle Luther and Lama Ahmad and Larry Kai and Lauren Itow and Lauren Workman and Leher Pathak and Leo Chen and Li Jing and Lia Guy and Liam Fedus and Liang Zhou and Lien Mamitsuka and Lilian Weng and Lindsay McCallum and Lindsey Held and Long Ouyang and Louis Feuvrier and Lu Zhang and Lukas Kondraciuk and Lukasz Kaiser and Luke Hewitt and Luke Metz and Lyric Doshi and Mada Aflak and Maddie Simens and Madelaine Boyd and Madeleine Thompson and Marat Dukhan and Mark Chen and Mark Gray and Mark Hudnall and Marvin Zhang and Marwan Aljubeh and Mateusz Litwin and Matthew Zeng and Max Johnson and Maya Shetty and Mayank Gupta and Meghan Shah and Mehmet Yatbaz and Meng Jia Yang and Mengchao Zhong and Mia Glaese and Mianna Chen and Michael Janner and Michael Lampe and Michael Petrov and Michael Wu and Michele Wang and Michelle Fradin and Michelle Pokrass and Miguel Castro and Miguel Oom Temudo de Castro and Mikhail Pavlov and Miles Brundage and Miles Wang and Minal Khan and Mira Murati and Mo Bavarian and Molly Lin and Murat Yesildal and Nacho Soto and Natalia Gimelshein and Natalie Cone and Natalie Staudacher and Natalie Summers and Natan LaFontaine and Neil Chowdhury and Nick Ryder and Nick Stathas and Nick Turley and Nik Tezak and Niko Felix and Nithanth Kudige and Nitish Keskar and Noah Deutsch and Noel Bundick and Nora Puckett and Ofir Nachum and Ola Okelola and Oleg Boiko and Oleg Murk and Oliver Jaffe and Olivia Watkins and Olivier Godement and Owen Campbell-Moore and Patrick Chao and Paul McMillan and Pavel Belov and Peng Su and Peter Bak and Peter Bakkum and Peter Deng and Peter Dolan and Peter Hoeschele and Peter Welinder and Phil Tillet and Philip Pronin and Philippe Tillet and Prafulla Dhariwal and Qiming Yuan and Rachel Dias and Rachel Lim and Rahul Arora and Rajan Troll and Randall Lin and Rapha Gontijo Lopes and Raul Puri and Reah Miyara and Reimar Leike and Renaud Gaubert and Reza Zamani and Ricky Wang and Rob Donnelly and Rob Honsby and Rocky Smith and Rohan Sahai and Rohit Ramchandani and Romain Huet and Rory Carmichael and Rowan Zellers and Roy Chen and Ruby Chen and Ruslan Nigmatullin and Ryan Cheu and Saachi Jain and Sam Altman and Sam Schoenholz and Sam Toizer and Samuel Miserendino and Sandhini Agarwal and Sara Culver and Scott Ethersmith and Scott Gray and Sean Grove and Sean Metzger and Shamez Hermani and Shantanu Jain and Shengjia Zhao and Sherwin Wu and Shino Jomoto and Shirong Wu and Shuaiqi and Xia and Sonia Phene and Spencer Papay and Srinivas Narayanan and Steve Coffey and Steve Lee and Stewart Hall and Suchir Balaji and Tal Broda and Tal Stramer and Tao Xu and Tarun Gogineni and Taya Christianson and Ted Sanders and Tejal Patwardhan and Thomas Cunninghman and Thomas Degry and Thomas Dimson and Thomas Raoux and Thomas Shadwell and Tianhao Zheng and Todd Underwood and Todor Markov and Toki Sherbakov and Tom Rubin and Tom Stasi and Tomer Kaftan and Tristan Heywood and Troy Peterson and Tyce Walters and Tyna Eloundou and Valerie Qi and Veit Moeller and Vinnie Monaco and Vishal Kuo and Vlad Fomenko and Wayne Chang and Weiyi Zheng and Wenda Zhou and Wesam Manassra and Will Sheu and Wojciech Zaremba and Yash Patil and Yilei Qian and Yongjik Kim and Youlong Cheng and Yu Zhang and Yuchen He and Yuchen Zhang and Yujia Jin and Yunxing Dai and Yury Malkov},
      year={2024},
      eprint={2410.21276},
      archivePrefix={arXiv},
      primaryClass={cs.CL},
      url={https://arxiv.org/abs/2410.21276}, 
}

@misc{qwen2025qwen25technicalreport,
      title={Qwen2.5 Technical Report}, 
      author={Qwen and : and An Yang and Baosong Yang and Beichen Zhang and Binyuan Hui and Bo Zheng and Bowen Yu and Chengyuan Li and Dayiheng Liu and Fei Huang and Haoran Wei and Huan Lin and Jian Yang and Jianhong Tu and Jianwei Zhang and Jianxin Yang and Jiaxi Yang and Jingren Zhou and Junyang Lin and Kai Dang and Keming Lu and Keqin Bao and Kexin Yang and Le Yu and Mei Li and Mingfeng Xue and Pei Zhang and Qin Zhu and Rui Men and Runji Lin and Tianhao Li and Tianyi Tang and Tingyu Xia and Xingzhang Ren and Xuancheng Ren and Yang Fan and Yang Su and Yichang Zhang and Yu Wan and Yuqiong Liu and Zeyu Cui and Zhenru Zhang and Zihan Qiu},
      year={2025},
      eprint={2412.15115},
      archivePrefix={arXiv},
      primaryClass={cs.CL},
      url={https://arxiv.org/abs/2412.15115}, 
}

@misc{yang2025qwen3technicalreport,
      title={Qwen3 Technical Report}, 
      author={An Yang and Anfeng Li and Baosong Yang and Beichen Zhang and Binyuan Hui and Bo Zheng and Bowen Yu and Chang Gao and Chengen Huang and Chenxu Lv and Chujie Zheng and Dayiheng Liu and Fan Zhou and Fei Huang and Feng Hu and Hao Ge and Haoran Wei and Huan Lin and Jialong Tang and Jian Yang and Jianhong Tu and Jianwei Zhang and Jianxin Yang and Jiaxi Yang and Jing Zhou and Jingren Zhou and Junyang Lin and Kai Dang and Keqin Bao and Kexin Yang and Le Yu and Lianghao Deng and Mei Li and Mingfeng Xue and Mingze Li and Pei Zhang and Peng Wang and Qin Zhu and Rui Men and Ruize Gao and Shixuan Liu and Shuang Luo and Tianhao Li and Tianyi Tang and Wenbiao Yin and Xingzhang Ren and Xinyu Wang and Xinyu Zhang and Xuancheng Ren and Yang Fan and Yang Su and Yichang Zhang and Yinger Zhang and Yu Wan and Yuqiong Liu and Zekun Wang and Zeyu Cui and Zhenru Zhang and Zhipeng Zhou and Zihan Qiu},
      year={2025},
      eprint={2505.09388},
      archivePrefix={arXiv},
      primaryClass={cs.CL},
      url={https://arxiv.org/abs/2505.09388}, 
}

@misc{deepseek-math,
  author = {Zhihong Shao and Peiyi Wang and Qihao Zhu and Runxin Xu and Junxiao Song and Mingchuan Zhang and Y.K. Li and Y. Wu and Daya Guo},
  title = {DeepSeekMath: Pushing the Limits of Mathematical Reasoning in Open Language Models},
  journal = {CoRR},
  volume = {abs/2402.03300},
  year = {2024},
  url = {https://arxiv.org/abs/2402.03300},
}

@misc{gehring2025rlefgroundingcodellms,
      title={RLEF: Grounding Code LLMs in Execution Feedback with Reinforcement Learning}, 
      author={Jonas Gehring and Kunhao Zheng and Jade Copet and Vegard Mella and Quentin Carbonneaux and Taco Cohen and Gabriel Synnaeve},
      year={2025},
      eprint={2410.02089},
      archivePrefix={arXiv},
      primaryClass={cs.CL},
      url={https://arxiv.org/abs/2410.02089}, 
}

@inproceedings{NEURIPS2022_8636419d,
 author = {Le, Hung and Wang, Yue and Gotmare, Akhilesh Deepak and Savarese, Silvio and Hoi, Steven Chu Hong},
 booktitle = {Advances in Neural Information Processing Systems},
 editor = {S. Koyejo and S. Mohamed and A. Agarwal and D. Belgrave and K. Cho and A. Oh},
 pages = {21314--21328},
 publisher = {Curran Associates, Inc.},
 title = {CodeRL: Mastering Code Generation through Pretrained Models and Deep Reinforcement Learning},
 url = {https://proceedings.neurips.cc/paper_files/paper/2022/file/8636419dea1aa9fbd25fc4248e702da4-Paper-Conference.pdf},
 volume = {35},
 year = {2022}
}

@misc{yoshihara2025practicaltwostagerecipemathematical,
      title={A Practical Two-Stage Recipe for Mathematical LLMs: Maximizing Accuracy with SFT and Efficiency with Reinforcement Learning}, 
      author={Hiroshi Yoshihara and Taiki Yamaguchi and Yuichi Inoue},
      year={2025},
      eprint={2507.08267},
      archivePrefix={arXiv},
      primaryClass={cs.LG},
      url={https://arxiv.org/abs/2507.08267}, 
}

@misc{huang2025visionr1incentivizingreasoningcapability,
      title={Vision-R1: Incentivizing Reasoning Capability in Multimodal Large Language Models}, 
      author={Wenxuan Huang and Bohan Jia and Zijie Zhai and Shaosheng Cao and Zheyu Ye and Fei Zhao and Zhe Xu and Yao Hu and Shaohui Lin},
      year={2025},
      eprint={2503.06749},
      archivePrefix={arXiv},
      primaryClass={cs.CV},
      url={https://arxiv.org/abs/2503.06749}, 
}

@misc{rodriguez2025renderingawarereinforcementlearningvector,
      title={Rendering-Aware Reinforcement Learning for Vector Graphics Generation}, 
      author={Juan A. Rodriguez and Haotian Zhang and Abhay Puri and Aarash Feizi and Rishav Pramanik and Pascal Wichmann and Arnab Mondal and Mohammad Reza Samsami and Rabiul Awal and Perouz Taslakian and Spandana Gella and Sai Rajeswar and David Vazquez and Christopher Pal and Marco Pedersoli},
      year={2025},
      eprint={2505.20793},
      archivePrefix={arXiv},
      primaryClass={cs.CV},
      url={https://arxiv.org/abs/2505.20793}, 
}

@inproceedings{NEURIPS2022_b1efde53,
 author = {Ouyang, Long and Wu, Jeffrey and Jiang, Xu and Almeida, Diogo and Wainwright, Carroll and Mishkin, Pamela and Zhang, Chong and Agarwal, Sandhini and Slama, Katarina and Ray, Alex and Schulman, John and Hilton, Jacob and Kelton, Fraser and Miller, Luke and Simens, Maddie and Askell, Amanda and Welinder, Peter and Christiano, Paul F and Leike, Jan and Lowe, Ryan},
 booktitle = {Advances in Neural Information Processing Systems},
 editor = {S. Koyejo and S. Mohamed and A. Agarwal and D. Belgrave and K. Cho and A. Oh},
 pages = {27730--27744},
 publisher = {Curran Associates, Inc.},
 title = {Training language models to follow instructions with human feedback},
 url = {https://proceedings.neurips.cc/paper_files/paper/2022/file/b1efde53be364a73914f58805a001731-Paper-Conference.pdf},
 volume = {35},
 year = {2022}
}

@misc{lambert2025tulu3pushingfrontiers,
      title={Tulu 3: Pushing Frontiers in Open Language Model Post-Training}, 
      author={Nathan Lambert and Jacob Morrison and Valentina Pyatkin and Shengyi Huang and Hamish Ivison and Faeze Brahman and Lester James V. Miranda and Alisa Liu and Nouha Dziri and Shane Lyu and Yuling Gu and Saumya Malik and Victoria Graf and Jena D. Hwang and Jiangjiang Yang and Ronan Le Bras and Oyvind Tafjord and Chris Wilhelm and Luca Soldaini and Noah A. Smith and Yizhong Wang and Pradeep Dasigi and Hannaneh Hajishirzi},
      year={2025},
      eprint={2411.15124},
      archivePrefix={arXiv},
      primaryClass={cs.CL},
      url={https://arxiv.org/abs/2411.15124}, 
}

@inproceedings{zhang2025scimage,
title={ScImage: How good are multimodal large language models at scientific text-to-image generation?},
author={Leixin Zhang and Steffen Eger and Yinjie Cheng and WEIHE ZHAI and Jonas Belouadi and Fahimeh Moafian and Zhixue Zhao},
booktitle={The Thirteenth International Conference on Learning Representations},
year={2025},
url={https://openreview.net/forum?id=ugyqNEOjoU}
}

@inproceedings{kwon2023efficient,
  title={Efficient Memory Management for Large Language Model Serving with PagedAttention},
  author={Woosuk Kwon and Zhuohan Li and Siyuan Zhuang and Ying Sheng and Lianmin Zheng and Cody Hao Yu and Joseph E. Gonzalez and Hao Zhang and Ion Stoica},
  booktitle={Proceedings of the ACM SIGOPS 29th Symposium on Operating Systems Principles},
  year={2023}
}

@misc{bai2025qwen25vltechnicalreport,
      title={Qwen2.5-VL Technical Report}, 
      author={Shuai Bai and Keqin Chen and Xuejing Liu and Jialin Wang and Wenbin Ge and Sibo Song and Kai Dang and Peng Wang and Shijie Wang and Jun Tang and Humen Zhong and Yuanzhi Zhu and Mingkun Yang and Zhaohai Li and Jianqiang Wan and Pengfei Wang and Wei Ding and Zheren Fu and Yiheng Xu and Jiabo Ye and Xi Zhang and Tianbao Xie and Zesen Cheng and Hang Zhang and Zhibo Yang and Haiyang Xu and Junyang Lin},
      year={2025},
      eprint={2502.13923},
      archivePrefix={arXiv},
      primaryClass={cs.CV},
      url={https://arxiv.org/abs/2502.13923}, 
}

@misc{zhu2025internvl3exploringadvancedtraining,
      title={InternVL3: Exploring Advanced Training and Test-Time Recipes for Open-Source Multimodal Models}, 
      author={Jinguo Zhu and Weiyun Wang and Zhe Chen and Zhaoyang Liu and Shenglong Ye and Lixin Gu and Hao Tian and Yuchen Duan and Weijie Su and Jie Shao and Zhangwei Gao and Erfei Cui and Xuehui Wang and Yue Cao and Yangzhou Liu and Xingguang Wei and Hongjie Zhang and Haomin Wang and Weiye Xu and Hao Li and Jiahao Wang and Nianchen Deng and Songze Li and Yinan He and Tan Jiang and Jiapeng Luo and Yi Wang and Conghui He and Botian Shi and Xingcheng Zhang and Wenqi Shao and Junjun He and Yingtong Xiong and Wenwen Qu and Peng Sun and Penglong Jiao and Han Lv and Lijun Wu and Kaipeng Zhang and Huipeng Deng and Jiaye Ge and Kai Chen and Limin Wang and Min Dou and Lewei Lu and Xizhou Zhu and Tong Lu and Dahua Lin and Yu Qiao and Jifeng Dai and Wenhai Wang},
      year={2025},
      eprint={2504.10479},
      archivePrefix={arXiv},
      primaryClass={cs.CV},
      url={https://arxiv.org/abs/2504.10479}, 
}

@inproceedings{papineni-etal-2002-bleu,
    title = "{B}leu: a Method for Automatic Evaluation of Machine Translation",
    author = "Papineni, Kishore  and
      Roukos, Salim  and
      Ward, Todd  and
      Zhu, Wei-Jing",
    editor = "Isabelle, Pierre  and
      Charniak, Eugene  and
      Lin, Dekang",
    booktitle = "Proceedings of the 40th Annual Meeting of the Association for Computational Linguistics",
    month = jul,
    year = "2002",
    address = "Philadelphia, Pennsylvania, USA",
    publisher = "Association for Computational Linguistics",
    url = "https://aclanthology.org/P02-1040/",
    doi = "10.3115/1073083.1073135",
    pages = "311--318"
}

@inproceedings{lin-2004-rouge,
    title = "{ROUGE}: A Package for Automatic Evaluation of Summaries",
    author = "Lin, Chin-Yew",
    booktitle = "Text Summarization Branches Out",
    month = jul,
    year = "2004",
    address = "Barcelona, Spain",
    publisher = "Association for Computational Linguistics",
    url = "https://aclanthology.org/W04-1013/",
    pages = "74--81"
}

@inproceedings{reimers-gurevych-2019-sentence,
    title = "Sentence-{BERT}: Sentence Embeddings using {S}iamese {BERT}-Networks",
    author = "Reimers, Nils  and
      Gurevych, Iryna",
    editor = "Inui, Kentaro  and
      Jiang, Jing  and
      Ng, Vincent  and
      Wan, Xiaojun",
    booktitle = "Proceedings of the 2019 Conference on Empirical Methods in Natural Language Processing and the 9th International Joint Conference on Natural Language Processing (EMNLP-IJCNLP)",
    month = nov,
    year = "2019",
    address = "Hong Kong, China",
    publisher = "Association for Computational Linguistics",
    url = "https://aclanthology.org/D19-1410/",
    doi = "10.18653/v1/D19-1410",
    pages = "3982--3992"
}

@misc{liu2025understandingr1zeroliketrainingcritical,
      title={Understanding R1-Zero-Like Training: A Critical Perspective}, 
      author={Zichen Liu and Changyu Chen and Wenjun Li and Penghui Qi and Tianyu Pang and Chao Du and Wee Sun Lee and Min Lin},
      year={2025},
      eprint={2503.20783},
      archivePrefix={arXiv},
      primaryClass={cs.LG},
      url={https://arxiv.org/abs/2503.20783}, 
}

@misc{yu2025dapoopensourcellmreinforcement,
      title={DAPO: An Open-Source LLM Reinforcement Learning System at Scale}, 
      author={Qiying Yu and Zheng Zhang and Ruofei Zhu and Yufeng Yuan and Xiaochen Zuo and Yu Yue and Weinan Dai and Tiantian Fan and Gaohong Liu and Lingjun Liu and Xin Liu and Haibin Lin and Zhiqi Lin and Bole Ma and Guangming Sheng and Yuxuan Tong and Chi Zhang and Mofan Zhang and Wang Zhang and Hang Zhu and Jinhua Zhu and Jiaze Chen and Jiangjie Chen and Chengyi Wang and Hongli Yu and Yuxuan Song and Xiangpeng Wei and Hao Zhou and Jingjing Liu and Wei-Ying Ma and Ya-Qin Zhang and Lin Yan and Mu Qiao and Yonghui Wu and Mingxuan Wang},
      year={2025},
      eprint={2503.14476},
      archivePrefix={arXiv},
      primaryClass={cs.LG},
      url={https://arxiv.org/abs/2503.14476}, 
}

@misc{beyer2024paligemmaversatile3bvlm,
      title={PaliGemma: A versatile 3B VLM for transfer}, 
      author={Lucas Beyer and Andreas Steiner and André Susano Pinto and Alexander Kolesnikov and Xiao Wang and Daniel Salz and Maxim Neumann and Ibrahim Alabdulmohsin and Michael Tschannen and Emanuele Bugliarello and Thomas Unterthiner and Daniel Keysers and Skanda Koppula and Fangyu Liu and Adam Grycner and Alexey Gritsenko and Neil Houlsby and Manoj Kumar and Keran Rong and Julian Eisenschlos and Rishabh Kabra and Matthias Bauer and Matko Bošnjak and Xi Chen and Matthias Minderer and Paul Voigtlaender and Ioana Bica and Ivana Balazevic and Joan Puigcerver and Pinelopi Papalampidi and Olivier Henaff and Xi Xiong and Radu Soricut and Jeremiah Harmsen and Xiaohua Zhai},
      year={2024},
      eprint={2407.07726},
      archivePrefix={arXiv},
      primaryClass={cs.CV},
      url={https://arxiv.org/abs/2407.07726}, 
}

@misc{zhai2023sigmoidlosslanguageimage,
      title={Sigmoid Loss for Language Image Pre-Training}, 
      author={Xiaohua Zhai and Basil Mustafa and Alexander Kolesnikov and Lucas Beyer},
      year={2023},
      eprint={2303.15343},
      archivePrefix={arXiv},
      primaryClass={cs.CV},
      url={https://arxiv.org/abs/2303.15343}, 
}

@INPROCEEDINGS{710701,
  author={Rubner, Y. and Tomasi, C. and Guibas, L.J.},
  booktitle={Sixth International Conference on Computer Vision (IEEE Cat. No.98CH36271)}, 
  title={A metric for distributions with applications to image databases}, 
  year={1998},
  volume={},
  number={},
  pages={59-66},
  doi={10.1109/ICCV.1998.710701}
}

@inproceedings{10.5555/3045118.3045221,
author = {Kusner, Matt J. and Sun, Yu and Kolkin, Nicholas I. and Weinberger, Kilian Q.},
title = {From word embeddings to document distances},
year = {2015},
publisher = {JMLR.org},
booktitle = {Proceedings of the 32nd International Conference on International Conference on Machine Learning - Volume 37},
pages = {957–966},
numpages = {10},
location = {Lille, France},
series = {ICML'15}
}

@inproceedings{10.5555/3666122.3668330,
author = {Fu, Stephanie and Tamir, Netanel Y. and Sundaram, Shobhita and Chai, Lucy and Zhang, Richard and Dekel, Tali and Isola, Phillip},
title = {DreamSim: learning new dimensions of human visual similarity using synthetic data},
year = {2023},
publisher = {Curran Associates Inc.},
address = {Red Hook, NY, USA},
booktitle = {Proceedings of the 37th International Conference on Neural Information Processing Systems},
articleno = {2208},
numpages = {27},
location = {New Orleans, LA, USA},
series = {NIPS '23}
}

@inproceedings{hessel-etal-2021-clipscore,
    title = "{CLIPS}core: A Reference-free Evaluation Metric for Image Captioning",
    author = "Hessel, Jack  and
      Holtzman, Ari  and
      Forbes, Maxwell  and
      Le Bras, Ronan  and
      Choi, Yejin",
    editor = "Moens, Marie-Francine  and
      Huang, Xuanjing  and
      Specia, Lucia  and
      Yih, Scott Wen-tau",
    booktitle = "Proceedings of the 2021 Conference on Empirical Methods in Natural Language Processing",
    month = nov,
    year = "2021",
    address = "Online and Punta Cana, Dominican Republic",
    publisher = "Association for Computational Linguistics",
    url = "https://aclanthology.org/2021.emnlp-main.595/",
    doi = "10.18653/v1/2021.emnlp-main.595",
    pages = "7514--7528"
}

@article{DBLP:journals/corr/abs-2303-08774,
  author = {OpenAI},
  bibsource = {dblp computer science bibliography, https://dblp.org},
  doi = {10.48550/ARXIV.2303.08774},
  eprint = {2303.08774},
  eprinttype = {arXiv},
  journal = {CoRR},
  title = {{GPT-4} Technical Report},
  url = {https://doi.org/10.48550/arXiv.2303.08774},
  volume = {abs/2303.08774},
  year = 2023
}

@inproceedings{tang-etal-2023-vistext,
    title = "{V}is{T}ext: A Benchmark for Semantically Rich Chart Captioning",
    author = "Tang, Benny  and
      Boggust, Angie  and
      Satyanarayan, Arvind",
    editor = "Rogers, Anna  and
      Boyd-Graber, Jordan  and
      Okazaki, Naoaki",
    booktitle = "Proceedings of the 61st Annual Meeting of the Association for Computational Linguistics (Volume 1: Long Papers)",
    month = jul,
    year = "2023",
    address = "Toronto, Canada",
    publisher = "Association for Computational Linguistics",
    url = "https://aclanthology.org/2023.acl-long.401/",
    doi = "10.18653/v1/2023.acl-long.401",
    pages = "7268--7298"
}

@inproceedings{10.1145/3613905.3650738,
author = {Hsu, Ting-Yao and Huang, Chieh-Yang and Huang, Shih-Hong and Rossi, Ryan and Kim, Sungchul and Yu, Tong and Giles, C Lee and Huang, Ting-Hao Kenneth},
title = {SciCapenter: Supporting Caption Composition for Scientific Figures with Machine-Generated Captions and Ratings},
year = {2024},
isbn = {9798400703317},
publisher = {Association for Computing Machinery},
address = {New York, NY, USA},
url = {https://doi.org/10.1145/3613905.3650738},
doi = {10.1145/3613905.3650738},
booktitle = {Extended Abstracts of the CHI Conference on Human Factors in Computing Systems},
articleno = {284},
numpages = {9},
location = {Honolulu, HI, USA},
series = {CHI EA '24}
}

@misc{hu2023llmadaptersadapterfamilyparameterefficient,
      title={LLM-Adapters: An Adapter Family for Parameter-Efficient Fine-Tuning of Large Language Models}, 
      author={Zhiqiang Hu and Lei Wang and Yihuai Lan and Wanyu Xu and Ee-Peng Lim and Lidong Bing and Xing Xu and Soujanya Poria and Roy Ka-Wei Lee},
      year={2023},
      eprint={2304.01933},
      archivePrefix={arXiv},
      primaryClass={cs.CL},
      url={https://arxiv.org/abs/2304.01933}, 
}

@misc{grattafiori2024llama3herdmodels,
      title={The Llama 3 Herd of Models}, 
      author={Aaron Grattafiori and Abhimanyu Dubey and Abhinav Jauhri and Abhinav Pandey and Abhishek Kadian and Ahmad Al-Dahle and Aiesha Letman and Akhil Mathur and Alan Schelten and Alex Vaughan and Amy Yang and Angela Fan and Anirudh Goyal and Anthony Hartshorn and Aobo Yang and Archi Mitra and Archie Sravankumar and Artem Korenev and Arthur Hinsvark and Arun Rao and Aston Zhang and Aurelien Rodriguez and Austen Gregerson and Ava Spataru and Baptiste Roziere and Bethany Biron and Binh Tang and Bobbie Chern and Charlotte Caucheteux and Chaya Nayak and Chloe Bi and Chris Marra and Chris McConnell and Christian Keller and Christophe Touret and Chunyang Wu and Corinne Wong and Cristian Canton Ferrer and Cyrus Nikolaidis and Damien Allonsius and Daniel Song and Danielle Pintz and Danny Livshits and Danny Wyatt and David Esiobu and Dhruv Choudhary and Dhruv Mahajan and Diego Garcia-Olano and Diego Perino and Dieuwke Hupkes and Egor Lakomkin and Ehab AlBadawy and Elina Lobanova and Emily Dinan and Eric Michael Smith and Filip Radenovic and Francisco Guzmán and Frank Zhang and Gabriel Synnaeve and Gabrielle Lee and Georgia Lewis Anderson and Govind Thattai and Graeme Nail and Gregoire Mialon and Guan Pang and Guillem Cucurell and Hailey Nguyen and Hannah Korevaar and Hu Xu and Hugo Touvron and Iliyan Zarov and Imanol Arrieta Ibarra and Isabel Kloumann and Ishan Misra and Ivan Evtimov and Jack Zhang and Jade Copet and Jaewon Lee and Jan Geffert and Jana Vranes and Jason Park and Jay Mahadeokar and Jeet Shah and Jelmer van der Linde and Jennifer Billock and Jenny Hong and Jenya Lee and Jeremy Fu and Jianfeng Chi and Jianyu Huang and Jiawen Liu and Jie Wang and Jiecao Yu and Joanna Bitton and Joe Spisak and Jongsoo Park and Joseph Rocca and Joshua Johnstun and Joshua Saxe and Junteng Jia and Kalyan Vasuden Alwala and Karthik Prasad and Kartikeya Upasani and Kate Plawiak and Ke Li and Kenneth Heafield and Kevin Stone and Khalid El-Arini and Krithika Iyer and Kshitiz Malik and Kuenley Chiu and Kunal Bhalla and Kushal Lakhotia and Lauren Rantala-Yeary and Laurens van der Maaten and Lawrence Chen and Liang Tan and Liz Jenkins and Louis Martin and Lovish Madaan and Lubo Malo and Lukas Blecher and Lukas Landzaat and Luke de Oliveira and Madeline Muzzi and Mahesh Pasupuleti and Mannat Singh and Manohar Paluri and Marcin Kardas and Maria Tsimpoukelli and Mathew Oldham and Mathieu Rita and Maya Pavlova and Melanie Kambadur and Mike Lewis and Min Si and Mitesh Kumar Singh and Mona Hassan and Naman Goyal and Narjes Torabi and Nikolay Bashlykov and Nikolay Bogoychev and Niladri Chatterji and Ning Zhang and Olivier Duchenne and Onur Çelebi and Patrick Alrassy and Pengchuan Zhang and Pengwei Li and Petar Vasic and Peter Weng and Prajjwal Bhargava and Pratik Dubal and Praveen Krishnan and Punit Singh Koura and Puxin Xu and Qing He and Qingxiao Dong and Ragavan Srinivasan and Raj Ganapathy and Ramon Calderer and Ricardo Silveira Cabral and Robert Stojnic and Roberta Raileanu and Rohan Maheswari and Rohit Girdhar and Rohit Patel and Romain Sauvestre and Ronnie Polidoro and Roshan Sumbaly and Ross Taylor and Ruan Silva and Rui Hou and Rui Wang and Saghar Hosseini and Sahana Chennabasappa and Sanjay Singh and Sean Bell and Seohyun Sonia Kim and Sergey Edunov and Shaoliang Nie and Sharan Narang and Sharath Raparthy and Sheng Shen and Shengye Wan and Shruti Bhosale and Shun Zhang and Simon Vandenhende and Soumya Batra and Spencer Whitman and Sten Sootla and Stephane Collot and Suchin Gururangan and Sydney Borodinsky and Tamar Herman and Tara Fowler and Tarek Sheasha and Thomas Georgiou and Thomas Scialom and Tobias Speckbacher and Todor Mihaylov and Tong Xiao and Ujjwal Karn and Vedanuj Goswami and Vibhor Gupta and Vignesh Ramanathan and Viktor Kerkez and Vincent Gonguet and Virginie Do and Vish Vogeti and Vítor Albiero and Vladan Petrovic and Weiwei Chu and Wenhan Xiong and Wenyin Fu and Whitney Meers and Xavier Martinet and Xiaodong Wang and Xiaofang Wang and Xiaoqing Ellen Tan and Xide Xia and Xinfeng Xie and Xuchao Jia and Xuewei Wang and Yaelle Goldschlag and Yashesh Gaur and Yasmine Babaei and Yi Wen and Yiwen Song and Yuchen Zhang and Yue Li and Yuning Mao and Zacharie Delpierre Coudert and Zheng Yan and Zhengxing Chen and Zoe Papakipos and Aaditya Singh and Aayushi Srivastava and Abha Jain and Adam Kelsey and Adam Shajnfeld and Adithya Gangidi and Adolfo Victoria and Ahuva Goldstand and Ajay Menon and Ajay Sharma and Alex Boesenberg and Alexei Baevski and Allie Feinstein and Amanda Kallet and Amit Sangani and Amos Teo and Anam Yunus and Andrei Lupu and Andres Alvarado and Andrew Caples and Andrew Gu and Andrew Ho and Andrew Poulton and Andrew Ryan and Ankit Ramchandani and Annie Dong and Annie Franco and Anuj Goyal and Aparajita Saraf and Arkabandhu Chowdhury and Ashley Gabriel and Ashwin Bharambe and Assaf Eisenman and Azadeh Yazdan and Beau James and Ben Maurer and Benjamin Leonhardi and Bernie Huang and Beth Loyd and Beto De Paola and Bhargavi Paranjape and Bing Liu and Bo Wu and Boyu Ni and Braden Hancock and Bram Wasti and Brandon Spence and Brani Stojkovic and Brian Gamido and Britt Montalvo and Carl Parker and Carly Burton and Catalina Mejia and Ce Liu and Changhan Wang and Changkyu Kim and Chao Zhou and Chester Hu and Ching-Hsiang Chu and Chris Cai and Chris Tindal and Christoph Feichtenhofer and Cynthia Gao and Damon Civin and Dana Beaty and Daniel Kreymer and Daniel Li and David Adkins and David Xu and Davide Testuggine and Delia David and Devi Parikh and Diana Liskovich and Didem Foss and Dingkang Wang and Duc Le and Dustin Holland and Edward Dowling and Eissa Jamil and Elaine Montgomery and Eleonora Presani and Emily Hahn and Emily Wood and Eric-Tuan Le and Erik Brinkman and Esteban Arcaute and Evan Dunbar and Evan Smothers and Fei Sun and Felix Kreuk and Feng Tian and Filippos Kokkinos and Firat Ozgenel and Francesco Caggioni and Frank Kanayet and Frank Seide and Gabriela Medina Florez and Gabriella Schwarz and Gada Badeer and Georgia Swee and Gil Halpern and Grant Herman and Grigory Sizov and Guangyi and Zhang and Guna Lakshminarayanan and Hakan Inan and Hamid Shojanazeri and Han Zou and Hannah Wang and Hanwen Zha and Haroun Habeeb and Harrison Rudolph and Helen Suk and Henry Aspegren and Hunter Goldman and Hongyuan Zhan and Ibrahim Damlaj and Igor Molybog and Igor Tufanov and Ilias Leontiadis and Irina-Elena Veliche and Itai Gat and Jake Weissman and James Geboski and James Kohli and Janice Lam and Japhet Asher and Jean-Baptiste Gaya and Jeff Marcus and Jeff Tang and Jennifer Chan and Jenny Zhen and Jeremy Reizenstein and Jeremy Teboul and Jessica Zhong and Jian Jin and Jingyi Yang and Joe Cummings and Jon Carvill and Jon Shepard and Jonathan McPhie and Jonathan Torres and Josh Ginsburg and Junjie Wang and Kai Wu and Kam Hou U and Karan Saxena and Kartikay Khandelwal and Katayoun Zand and Kathy Matosich and Kaushik Veeraraghavan and Kelly Michelena and Keqian Li and Kiran Jagadeesh and Kun Huang and Kunal Chawla and Kyle Huang and Lailin Chen and Lakshya Garg and Lavender A and Leandro Silva and Lee Bell and Lei Zhang and Liangpeng Guo and Licheng Yu and Liron Moshkovich and Luca Wehrstedt and Madian Khabsa and Manav Avalani and Manish Bhatt and Martynas Mankus and Matan Hasson and Matthew Lennie and Matthias Reso and Maxim Groshev and Maxim Naumov and Maya Lathi and Meghan Keneally and Miao Liu and Michael L. Seltzer and Michal Valko and Michelle Restrepo and Mihir Patel and Mik Vyatskov and Mikayel Samvelyan and Mike Clark and Mike Macey and Mike Wang and Miquel Jubert Hermoso and Mo Metanat and Mohammad Rastegari and Munish Bansal and Nandhini Santhanam and Natascha Parks and Natasha White and Navyata Bawa and Nayan Singhal and Nick Egebo and Nicolas Usunier and Nikhil Mehta and Nikolay Pavlovich Laptev and Ning Dong and Norman Cheng and Oleg Chernoguz and Olivia Hart and Omkar Salpekar and Ozlem Kalinli and Parkin Kent and Parth Parekh and Paul Saab and Pavan Balaji and Pedro Rittner and Philip Bontrager and Pierre Roux and Piotr Dollar and Polina Zvyagina and Prashant Ratanchandani and Pritish Yuvraj and Qian Liang and Rachad Alao and Rachel Rodriguez and Rafi Ayub and Raghotham Murthy and Raghu Nayani and Rahul Mitra and Rangaprabhu Parthasarathy and Raymond Li and Rebekkah Hogan and Robin Battey and Rocky Wang and Russ Howes and Ruty Rinott and Sachin Mehta and Sachin Siby and Sai Jayesh Bondu and Samyak Datta and Sara Chugh and Sara Hunt and Sargun Dhillon and Sasha Sidorov and Satadru Pan and Saurabh Mahajan and Saurabh Verma and Seiji Yamamoto and Sharadh Ramaswamy and Shaun Lindsay and Shaun Lindsay and Sheng Feng and Shenghao Lin and Shengxin Cindy Zha and Shishir Patil and Shiva Shankar and Shuqiang Zhang and Shuqiang Zhang and Sinong Wang and Sneha Agarwal and Soji Sajuyigbe and Soumith Chintala and Stephanie Max and Stephen Chen and Steve Kehoe and Steve Satterfield and Sudarshan Govindaprasad and Sumit Gupta and Summer Deng and Sungmin Cho and Sunny Virk and Suraj Subramanian and Sy Choudhury and Sydney Goldman and Tal Remez and Tamar Glaser and Tamara Best and Thilo Koehler and Thomas Robinson and Tianhe Li and Tianjun Zhang and Tim Matthews and Timothy Chou and Tzook Shaked and Varun Vontimitta and Victoria Ajayi and Victoria Montanez and Vijai Mohan and Vinay Satish Kumar and Vishal Mangla and Vlad Ionescu and Vlad Poenaru and Vlad Tiberiu Mihailescu and Vladimir Ivanov and Wei Li and Wenchen Wang and Wenwen Jiang and Wes Bouaziz and Will Constable and Xiaocheng Tang and Xiaojian Wu and Xiaolan Wang and Xilun Wu and Xinbo Gao and Yaniv Kleinman and Yanjun Chen and Ye Hu and Ye Jia and Ye Qi and Yenda Li and Yilin Zhang and Ying Zhang and Yossi Adi and Youngjin Nam and Yu and Wang and Yu Zhao and Yuchen Hao and Yundi Qian and Yunlu Li and Yuzi He and Zach Rait and Zachary DeVito and Zef Rosnbrick and Zhaoduo Wen and Zhenyu Yang and Zhiwei Zhao and Zhiyu Ma},
      year={2024},
      eprint={2407.21783},
      archivePrefix={arXiv},
      primaryClass={cs.AI},
      url={https://arxiv.org/abs/2407.21783}, 
}

@inproceedings{10.5555/3495724.3495883,
author = {Brown, Tom B. and Mann, Benjamin and Ryder, Nick and Subbiah, Melanie and Kaplan, Jared and Dhariwal, Prafulla and Neelakantan, Arvind and Shyam, Pranav and Sastry, Girish and Askell, Amanda and Agarwal, Sandhini and Herbert-Voss, Ariel and Krueger, Gretchen and Henighan, Tom and Child, Rewon and Ramesh, Aditya and Ziegler, Daniel M. and Wu, Jeffrey and Winter, Clemens and Hesse, Christopher and Chen, Mark and Sigler, Eric and Litwin, Mateusz and Gray, Scott and Chess, Benjamin and Clark, Jack and Berner, Christopher and McCandlish, Sam and Radford, Alec and Sutskever, Ilya and Amodei, Dario},
title = {Language models are few-shot learners},
year = {2020},
isbn = {9781713829546},
publisher = {Curran Associates Inc.},
address = {Red Hook, NY, USA},
booktitle = {Proceedings of the 34th International Conference on Neural Information Processing Systems},
articleno = {159},
numpages = {25},
location = {Vancouver, BC, Canada},
series = {NIPS '20}
}

@misc{loshchilov2019decoupledweightdecayregularization,
      title={Decoupled Weight Decay Regularization}, 
      author={Ilya Loshchilov and Frank Hutter},
      year={2019},
      eprint={1711.05101},
      archivePrefix={arXiv},
      primaryClass={cs.LG},
      url={https://arxiv.org/abs/1711.05101}, 
}

@inproceedings{caron2021emerging,
  title={Emerging Properties in Self-Supervised Vision Transformers},
  author={Caron, Mathilde and Touvron, Hugo and Misra, Ishan and J\'egou, Herv\'e  and Mairal, Julien and Bojanowski, Piotr and Joulin, Armand},
  booktitle={Proceedings of the International Conference on Computer Vision (ICCV)},
  year={2021}
}

@inproceedings{zhang2018unreasonable,
  title={The unreasonable effectiveness of deep features as a perceptual metric},
  author={Zhang, Richard and Isola, Phillip and Efros, Alexei A and Shechtman, Eli and Wang, Oliver},
  booktitle={Proceedings of the IEEE conference on computer vision and pattern recognition},
  pages={586--595},
  year={2018}
}

@article{pramanick2024spiqa,
  title={Spiqa: A dataset for multimodal question answering on scientific papers},
  author={Pramanick, Shraman and Chellappa, Rama and Venugopalan, Subhashini},
  journal={Advances in Neural Information Processing Systems},
  volume={37},
  pages={118807--118833},
  year={2024}
}
\bibliographystyle{plainnat}

\appendix

\section{Appendix}
\label{sec:Appendix}

\subsection{Caption Quality Analysis}
\label{app:caption_quality_analysis}
\begin{figure}[h]
  \centering
  \includegraphics[width=\textwidth]{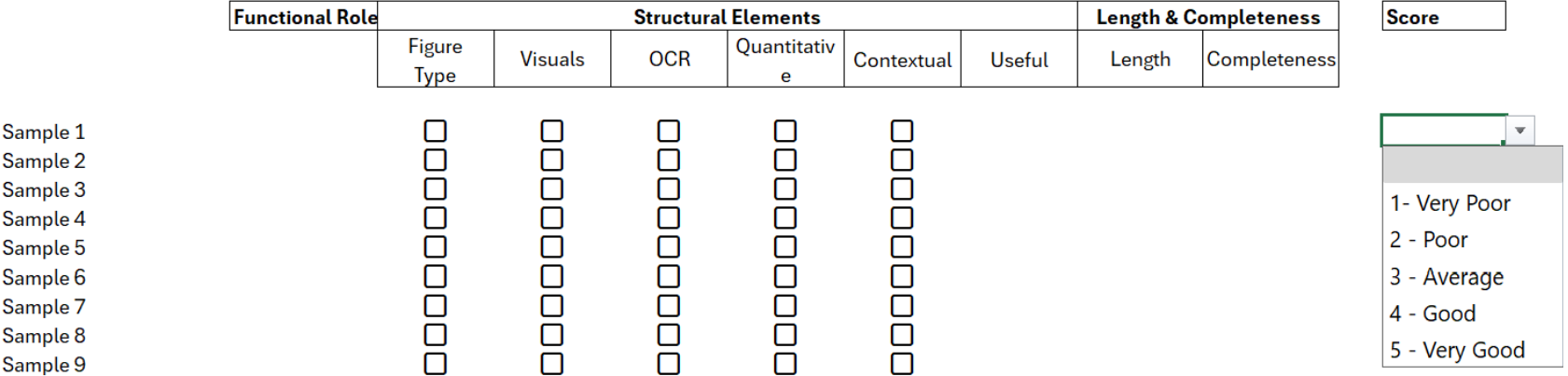}
  \caption{Screenshot of our evaluation form for the first nine scientific figures.}
  \label{fig:eval_sheet_caption_annotation}
\end{figure}

Our caption quality analysis involved three annotators: one bachelor’s student, one PhD student, and one faculty member (all male). From our subset of DaTikZ-V3, 74\% of samples originate from arXiv and 26\% from TeX SE. One annotator completed the evaluation sheet in Figure~\ref{fig:eval_sheet_caption_annotation}, based on the taxonomy in Table~\ref{tab:caption_analysis_taxonomy}. This annotator also manually described all 200 scientific figures, which we subsequently used as reference descriptions to compute BLEU-4, ROUGE-L, and STS with the \texttt{all-mpnet-base-v2} sentence encoder between human descriptions and VLM-generated descriptions.

The other two annotators each described 30 figures to measure agreement, yielding unweighted $\kappa=0.35$ and weighted $\kappa=0.63$. The structural elements for scientific figure captions were adapted from best practices in academic writing and prior research taxonomies~\citep{tang-etal-2023-vistext, 10.1145/3613905.3650738}.

\begin{table}[h!]
  \centering
  \scriptsize
  \setlength{\tabcolsep}{4pt}
  \renewcommand{\arraystretch}{1.05}
  \caption{Caption analysis taxonomy for structural elements and usefulness scores.}
  \label{tab:caption_analysis_taxonomy}
  \begin{tabularx}{\linewidth}{>{\raggedright\arraybackslash}p{0.19\linewidth} >{\raggedright\arraybackslash}X}
    \toprule
    \textbf{Structural Elements} &
      \textbf{Figure type:} names the high-level type (e.g., graph, tree, workflow). \\
      & \textbf{Visual details:} mentions colors, shapes, axes, layout/spatial relations. \\
      & \textbf{OCR:} includes textual elements visible in the figure (axis labels, annotations, math), aiding correct labeling. \\
      & \textbf{Contextual reference:} points outside the figure (e.g., “see Sec. 3”). Useful but reduces standalone utility. \\
      & \textbf{Quantitative content:} numbers, formulas, code. Adds technical substance (often paired with OCR). \\
    \midrule
    \textbf{Usefulness Scores} &
      \textbf{Very Poor:} not meaningfully descriptive. May be only a label or irrelevant text. \\
      & \textbf{Poor:} somewhat relevant but vague/incomplete. Mentions topic/elements without adequate clarity or context. \\
      & \textbf{Average:} describes the main content but lacks depth/specifics. States what it is without highlighting key details. \\
      & \textbf{Good:} clear, specific, and near-complete. Covers important visual/quantitative details and structure. \\
      & \textbf{Very Good:} precise, insightful, and largely self-contained. Explains key elements so the figure is almost unnecessary. \\
    \bottomrule
  \end{tabularx}
\end{table}

\subsection{Dataset}
\label{app:dataset}
To create synthetic data, we follow a strategy similar to ScImage~\citep{zhang2025scimage}. We first generate 2,000 templates with varied terms, each used to produce 10 queries that generate TikZ code. All steps are performed using GPT-4o with minimal human intervention.

\paragraph{LLM Debugging}
For LLM-based debugging, we use the prompt in Figure~\ref{fig:debug_prompt}. We first tested this on a subset of 753 samples spanning all sources, manually evaluating the percentage of compilable, non-empty, and non-corrupted outputs. As shown in Table~\ref{tab:debug_ablation}, Qwen3-32B (non-thinking) was the best-performing model, recovering 49.40\% of errors in a single pass and 59.04\% after three repair rounds. Smaller Qwen variants and Qwen2.5-7B-Instruct~\citep{qwen2025qwen25technicalreport} performed considerably worse. We therefore applied Qwen3-32B for large-scale debugging, which took 14 days on 4 × A100 40GB GPUs using the vLLM framework~\citep{kwon2023efficient}. Examples of the debugging process are shown in Figure~\ref{fig:example_debug} and \ref{fig:example_debug_new}.

\refstepcounter{figure}
\begin{PromptBox}{LLM Debug Prompt}
I will provide you with some TikZ code and the corresponding LaTeX error log. Fix the TikZ code so that it compiles without errors. Only output the corrected TikZ code.\textbackslash n

Original TikZ Code:
\{tikz\_code\}\textbackslash n

Compilation Error Log:
\{log\_message\}
\end{PromptBox}
\label{fig:debug_prompt}

\begin{table}[t]
  \centering
  \caption{Accuracy of different LLMs in debugging TikZ code from error logs over three refinement iterations. Bold indicates the best-performing model.}
  \label{tab:debug_ablation}
  \begin{tabular}{lrrr}
    \toprule
    \textbf{LLM} & \textbf{Iteration 1} & \textbf{Iteration 2} & \textbf{Iteration 3} \\
    \midrule
    Qwen2.5-7B  & 17.49\% & 28.03\% & 34.08\% \\
    Qwen3-4B    & 14.17\% & 18.56\% & 21.98\% \\
    Qwen3-8B    & 35.11\% & 39.36\% & 41.49\% \\
    Qwen3-32B   & \textbf{49.40\%} & \textbf{55.42\%} & \textbf{59.04\%} \\
    \midrule
    GPT-4o-mini & 36.82\% & 41.36\% & 43.62\% \\
    GPT-4o      & \underline{48.10\%} & \underline{53.73\%} & \underline{58.89\%} \\
    \bottomrule
  \end{tabular}
\end{table}

\begin{figure*}[htbp]
\centering
\begin{tcolorbox}[
    width=\textwidth,
    colback=gray!3,
    colframe=black!50,
    title=LLM-based TikZ Debugging,
    enhanced,
    sharp corners=south,
    ]
    \textbf{Original TikZ Code:}
    \begin{lstlisting}[style=mystyle]
    \documentclass[tikz]{standalone}
    \usepackage[utf8]{inputenc}
    \usepackage{circuitikz}
    \usepackage{float}
    \usepackage{calc}
    \begin{document}
    \begin{circuitikz}[american, straight voltages]
        \draw (-1,0)
        to [american voltage source, v=$V_P$, invert, voltage shift=1] (-1,4)
        to [R, R=$R_p$, i^>=$i_p$] (2,4)
        to [R=$R_L$] (4,4)
        to [L, l_=$L$, v^<=$v_L$, i=$i_L$, voltage shift=1.5] (7,4)
        to [Tnigbt,bodydiode] (10,4)
        to [short] (12,4)
        to [american voltage source, v^<=$V_{out}$, voltage shift=1] (12,0)
        to [short] (-1,0)
        (2.0,4) to [R=$R_Ci$, i=$i_{Ci}$] (2.0,1.5)
        to [C, l_=$C_i$, v^<=$v_{Ci}$] (2.0,0)
        (7.2,4) to [Tnigbt,bodydiode, invert] (7.2,0)
        (10.0,4) to [R=$R_Co$, i=$i_{Co}$] (10.0,1.5)
        to [C, l_=$C_o$, v^<=$v_{Co}$] (10.0,0)
        (8.5,5) node[align=center]{$G_2$}
        (6.1,2) node[align=center]{$G_1$}
        (7.2,0) node[circ, scale=1.5]{$1$}
        (7.2,4) node[circ, scale=1.5]
        (2,0) node[circ, scale=1.5]
        (2,4) node[circ, color=red, scale=1.5]
        (10,4) node[circ, color=red, scale=1.5]
        (10,0) node[circ, color=red, scale=1.5]
        ;
    \end{circuitikz}
    \end{document}
    \end{lstlisting}

    \medskip
    \textbf{Compiler Error Log:}
    \begin{lstlisting}[style=mystyle]
    ! Package tikz Error: A node must have a (possibly empty) label text.
    See the tikz package documentation for explanation.
    Type  H <return>  for immediate help.
     ...                                                                   
    l.26     (2,0) node[circ, scale=1.5]
    !  ==> Fatal error occurred, no output PDF file produced!
    Transcript written on figure.log.
    \end{lstlisting}

    \medskip
    \textbf{Corrected TikZ Code (Changed Parts):}
    \begin{lstlisting}[style=mystyle]
    ...
    (7.2,4) node[circ, scale=1.5]{}
    (2,0) node[circ, scale=1.5]{} 
    (2,4) node[circ, color=red, scale=1.5]{} 
    (10,4) node[circ, color=red, scale=1.5]{} 
    (10,0) node[circ, color=red, scale=1.5]{} 
    ...
    \end{lstlisting}

    \medskip
    \includegraphics[width=0.6\linewidth]{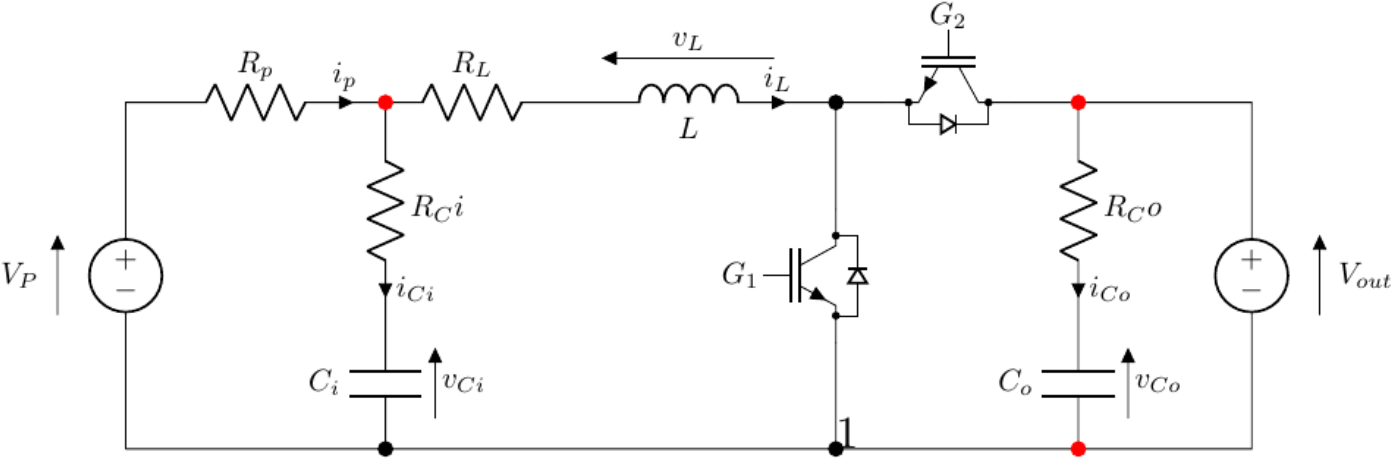}
\end{tcolorbox}
\caption{An example of the LLM debugging pipeline. The original TikZ code failed to compile. The compiler error log was passed to the LLM, which generated corrected TikZ code. The fixed code produces the valid figure shown above.}
\label{fig:example_debug}
\end{figure*}

\begin{figure*}[htbp]
\centering
\begin{tcolorbox}[
    width=\textwidth,
    colback=gray!3,
    colframe=black!50,
    title=LLM-based TikZ Debugging,
    enhanced,
    sharp corners=south,
    ]
    \textbf{Original TikZ Code:}
    \begin{lstlisting}[style=mystyle]
    \documentclass[tikz]{standalone}
    \usepackage{tikz}
    \usetikzlibrary{automata,shapes.geometric}
    \usepackage{array}
    \begin{document}
    \begin{figure}[h]
    \begin{tabular}{*{2}{>{\centering\arraybackslash}b{\dimexpr0.5\textwidth-2\tabcolsep\relax}}}
    \legend{Weighted, complete graph $K_H$}
    \begin{tikzpicture}[state/.append style={minimum size=5mm}]
        \node [state] (0) at (-2, 3) [label=left:E] {};
        \node [state] (1) at ( 2, 3) [label=right:B]{};
        \node [state] (2) at (-1.25, 0.75)[label=left:D] {};
        \node [state] (3) at ( 1.25, 0.75) [label=right:C]{};
        \node [state] (4) at ( 0, 4.5) [label=above: A]{};
        \draw  (0) to (4);
        \draw (4) to (1);
        \draw (1) to (3);
        \draw (0) to (2);
        \draw (3) to (2);
        \draw (2) to (4);
        \draw (4) to (3);
        \draw (0) to (1);
        \draw (0) to (3);
        \draw (2) to (1);
    \end{tikzpicture} &
    \renewcommand{\arraystretch}{1.3}
    \legend{\textbf{THIS IS TABLE LEGEND}}
    \begin{tabular}{c|ccccc}
     & A  & B & C & D & E  \\
    \hline
    A & --  & 4 & 7 & 6 & 12 \\
    B & 4  & -- & 3 & 5 & 8  \\
    C & 7  & 3 & -- & 2 & 5  \\
    D & 6  & 5 & 2 & -- & 9  \\
    E & 12 & 8 & 5 & 9 & --
    \end{tabular}
    \end{tabular}
    \end{figure}
    \end{document}
    \end{lstlisting}

    \medskip
    \textbf{Compiler Error Log:}
    \begin{lstlisting}[style=mystyle]
    ! LaTeX Error: Not allowed in LR mode.
    See the LaTeX manual or LaTeX Companion for explanation.
    Type  H <return>  for immediate help.
     ...                                              
    l.6 \begin{figure}[h]
    !  ==> Fatal error occurred, no output PDF file produced!
    Transcript written on figure.log.
    \end{lstlisting}

    \medskip
    \textbf{Corrected TikZ Code (Changed Parts):}
    \begin{lstlisting}[style=mystyle]
    ...
    \begin{document}
    \begin{tabular}{*{2}{>{\centering\arraybackslash}b{\dimexpr0.5\textwidth-2\tabcolsep\relax}}}
    \begin{tikzpicture}[state/.append style={minimum size=5mm}]
    ...
    \renewcommand{\arraystretch}{1.3}
    \begin{tabular}{c|ccccc}
    ...
    \end{tabular}
    \end{tabular}
    \end{document}
    \end{lstlisting}

    \medskip
    \includegraphics[width=0.3\linewidth]{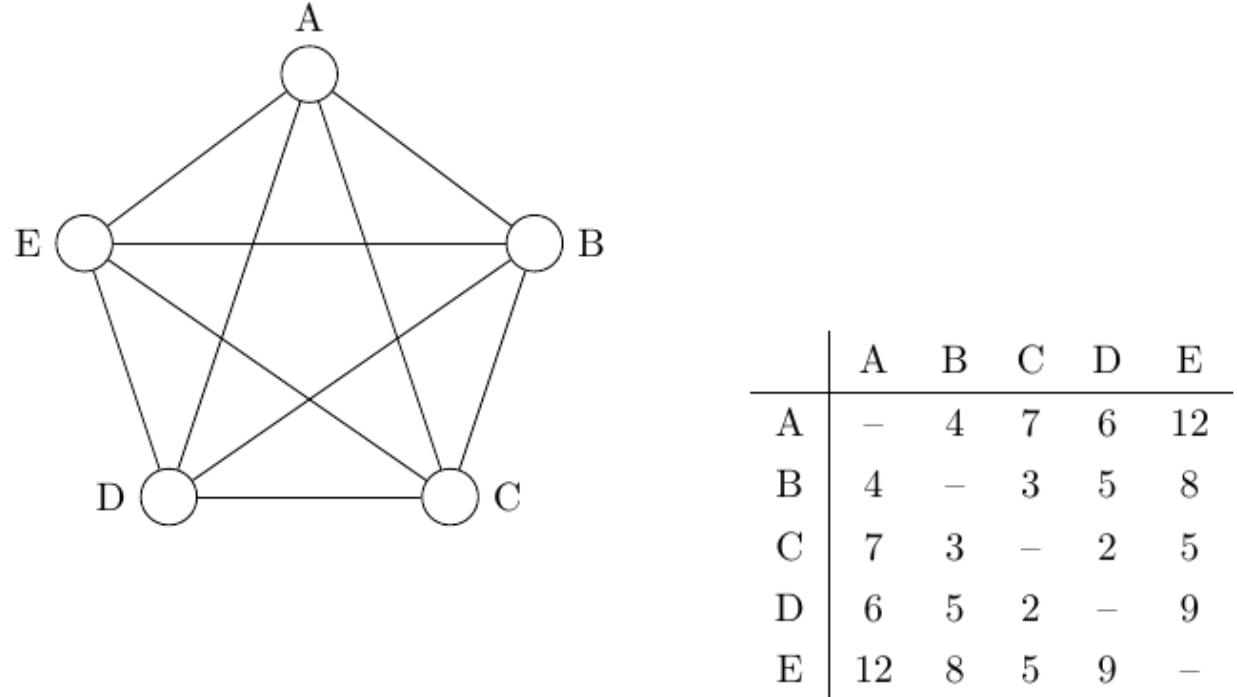}
\end{tcolorbox}
\caption{An example of the LLM debugging pipeline. The original TikZ code failed to compile. The compiler error log was passed to the LLM, which generated corrected TikZ code. The fixed code produces the valid figure shown above.} 
\label{fig:example_debug_new}
\end{figure*}

\paragraph{VLM-based Image Description}
The prompt for image description is shown in Figure~\ref{fig:describe_prompt}. We use few-shot in-context learning~\citep{10.5555/3495724.3495883} with two high-STS human descriptions as exemplars. We run Qwen2.5-VL-7B-Instruct, which was the strongest open-source VLM in our evaluation, to describe all figures in DaTikZ-V4. Processing required 2 days on 4 × A100 40GB GPUs.

\refstepcounter{figure}
\begin{PromptBox}{VLM Description Prompt}
You are a scientific illustrator describing images for precise redrawing in TikZ.\textbackslash n

Your task is to describe the image in precise, continuous prose without bullet points, lists, or line breaks.\textbackslash n

Start directly with the main object or scene. Avoid introductory phrases like 'Certainly!', 'The image depicts...', 'Here is a precise description.'.\textbackslash n

Use clear, active language focused on geometry, labels, colors, spatial relationships, coordinates, and other visible properties.\textbackslash n

Describe all visible elements such as shapes, lines, arrows, and labels, including their relative or absolute positions, dimensions, and orientation.\textbackslash n

Use consistent, minimal naming for objects (e.g., 'circle A', 'line L1') and specify label positions relative to shapes precisely.\textbackslash n

Only describe exact, concrete visual elements that enable precise image reconstruction in TikZ.\textbackslash n

Avoid vague, interpretive, or inferential language, and exclude summaries, conclusions, or commentary about the image's meaning, function, or aesthetics.\textbackslash n

Here are a few examples:\textbackslash n

A thin black horizontal line centered in the middle, containing nine evenly spaced black dots, and labeled $x_2$ at the left. Each dot is connected by a thin black line in an alternating pattern to either $x_0$ (placed at the top middle) or $x_1$ (placed at the bottom middle).\textbackslash n

A line chart has different instruction scales of 1/10, 1/4, 1/2, and 1 on the x-axis. On the y-axis it shows BLEU scores between 20 and 50, with steps of 5. The chart contains three lines with Zh-En in blue, De-En in red, and Fr-En in brown. All BLEU scores are initially 20 at the lowest instruction scale. As the instruction scale increases, BLEU scores improve for all pairs. De-En is the highest, closely followed by Fr-En and then Zh-En far below. The increase is largest from 1/10 to 1/4 and only marginally above an instruction scale of 1/4. The legend is placed inside the chart at the top left.\textbackslash n

Write a description in this exact style for the given image.
\end{PromptBox}
\label{fig:describe_prompt}

\subsection{Method}
\label{app:method}
For finetuning DeTikZify-V2~\citep{belouadi2024detikzify}, which is a SigLIP~\citep{zhai2023sigmoidlosslanguageimage} vision encoder of \texttt{PaliGemma-3b-mix-448}~\citep{beyer2024paligemmaversatile3bvlm}), we use the training split of DaTikZ-V4 consisting of 1.3M Image–TikZ pairs. Inputs are 448×448-pixel images with a maximum output length of 2048 tokens. Training runs for two epochs with a learning rate of 5e-5, AdamW~\citep{loshchilov2019decoupledweightdecayregularization}, cosine scheduler, and 3\% linear warmup. The batch size is 128, trained on 4 × H200 140GB GPUs for 12 days.

\subsection{Experiments}
\label{app:experiments}
The prompt template for all models is shown in Figure~\ref{fig:llm_prompt}. We also experimented with templates without the standalone environment but found that this reduced performance and compilation rates. 

\paragraph{Models}
Except for GPT-5, decoding uses \texttt{temperature=1.0}, \texttt{top\_p=0.9}, and max length 2048. For GPT-5, we set \texttt{reasoning=medium}, \texttt{verbosity=medium}, and evaluate a random subset of 100 samples due to cost. For TikZero, trained on caption–TikZ pairs, we only provide the figure description as prompt. For SFT, Qwen2.5-3B is finetuned on DaTikZ-V4 for two days with a learning rate of 1e-4, warmup ratio 3\%, cosine scheduler, and batch size 128. Qwen3-8B is trained for four days with a reduced learning rate of 5e-5. For RL on DaTikZ-V4-RL, TikZilla-3B is trained with GRPO for 4,000 iterations (batch size 144, 8 rollouts) using learning rate 5e-6 and weight decay 1e-2. TikZilla-8B uses learning rate 2e-6. RL-only runs were also tested. Training took 5 days for TikZilla-3B and 10 days for TikZilla-8B, all on 4 × H200 140GB GPUs.

\paragraph{Metrics}
CLIPScore (\textbf{CLIP}) is computed with \texttt{siglip-so400m-patch14-384}. DreamSIM (\textbf{DSim}) uses CLIP, DINO, and OpenCLIP (\texttt{ViT-B/16}). TeX Edit Distance (\textbf{TED}) uses \texttt{TexLexer}. Average tokens (\textbf{AT}) are measured with \texttt{o200k\_base} tokenizer. DINOScore (\textbf{DINO}) is calculated using the cosine similarity of the patch embeddings produced by \texttt{dino-vits16}. Learned Perceptual Image Patch Similarity (\textbf{LPIPS}) uses the \texttt{alex} network.

\refstepcounter{figure}
\begin{PromptBox}{Prompt Template}
Generate a complete LaTeX document that contains a TikZ figure according to the following requirements: \\
\{figure\_description\} \\
Wrap your code using \texttt{\textbackslash documentclass[tikz]\{standalone\}}, and include \texttt{\textbackslash begin\{document\}}...\texttt{\textbackslash end\{document\}}. Only output valid LaTeX code with no extra text.
\end{PromptBox}
\label{fig:llm_prompt}

\subsection{Results}
\label{app:results}

\paragraph{Human Evaluation}
We split 9 annotators (6 male, 3 female) into two groups. Group 1 (5 annotators) evaluated GPT-5, GPT-4o, Qwen2.5-3B, TikZilla-3B, and TikZilla-3B-RL. Group 2 (4 annotators) evaluated GPT-5, GPT-4o, Qwen3-8B, TikZilla-8B, and TikZilla-8B-RL. Each annotator received two Excel sheets (textual vs. image alignment), each with 30 randomized samples. We ensured at least five overlapping samples for inter-annotator agreement and five GPT-5 samples (scarcer due to cost). Annotation interfaces are shown in Figures~\ref{fig:t2i} and \ref{fig:i2i}. Likert scale definitions are shown below:
\begin{itemize}
    \item \textbf{5 Excellent}: Figure fulfills all requirements. Few minor issues (e.g., slightly imperfect layout, one or two mislabeled/extra elements) are acceptable. Think about it as publication or almost publication ready where only small tweaks needs to be made.
    \item \textbf{4 Good}: Figure broadly fulfills the requirements and contains no major errors, but it is clearly not perfect. Typical cases include multiple minor flaws (e.g., clutter, small inaccuracies, awkward design) or one moderate issue.
    \item \textbf{3 Fair}: The figure has about one to two major issues (e.g., important elements missing, wrong trends in charts, ...) and/or some minor issues. It is still usable with corrections as parts of the figure are clearly correct.
    \item \textbf{2 Poor}: Several major issues and/or many minor ones. The figure no longer meaningfully reflects the description or GT image (e.g., severe overlaps, high amounts of hallucinated content, ...).
    \item \textbf{1 Failed}: Non-compilable code (already auto-assigned).
\end{itemize}

\begin{figure}[h!]
    \centering
    \includegraphics[width=\textwidth]{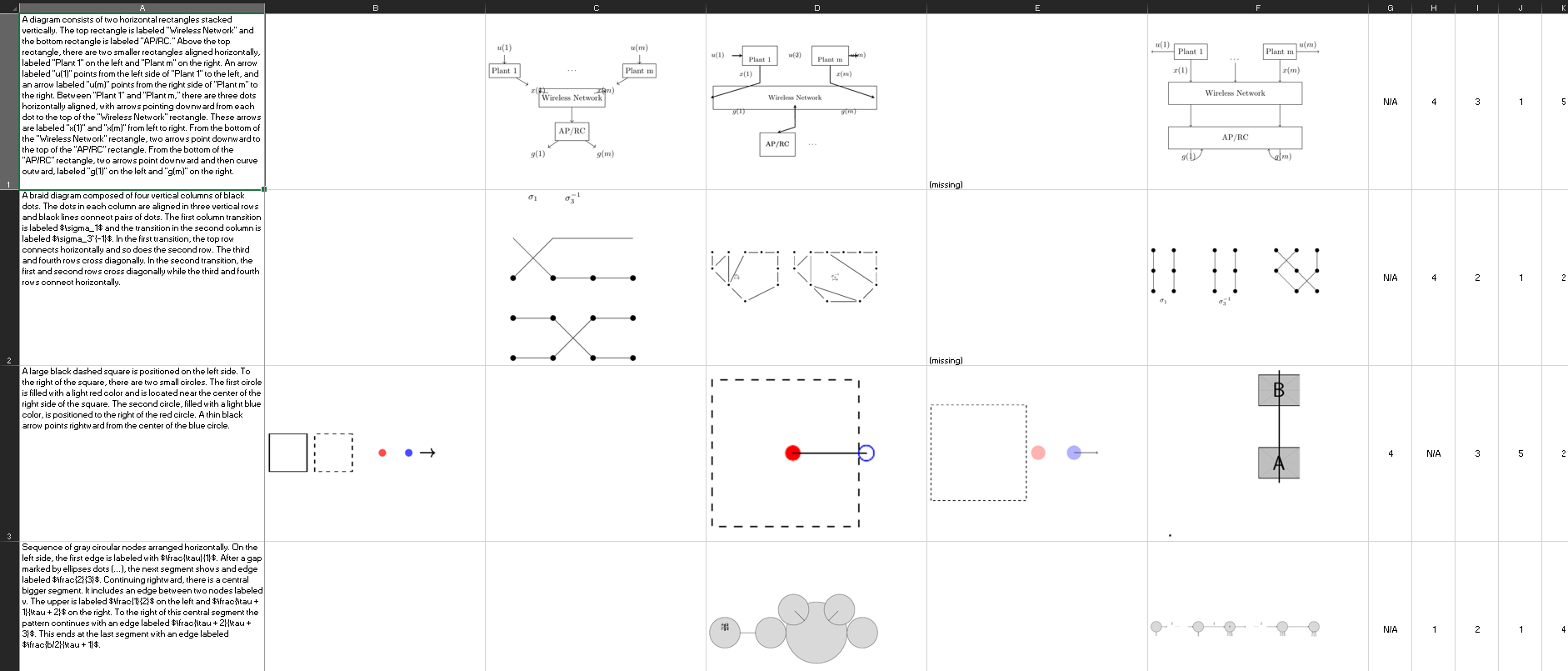}
    \caption{Example of text-image annotations.}
    \label{fig:t2i}
\end{figure}

\begin{figure}[h!]
    \centering
    \includegraphics[width=\textwidth]{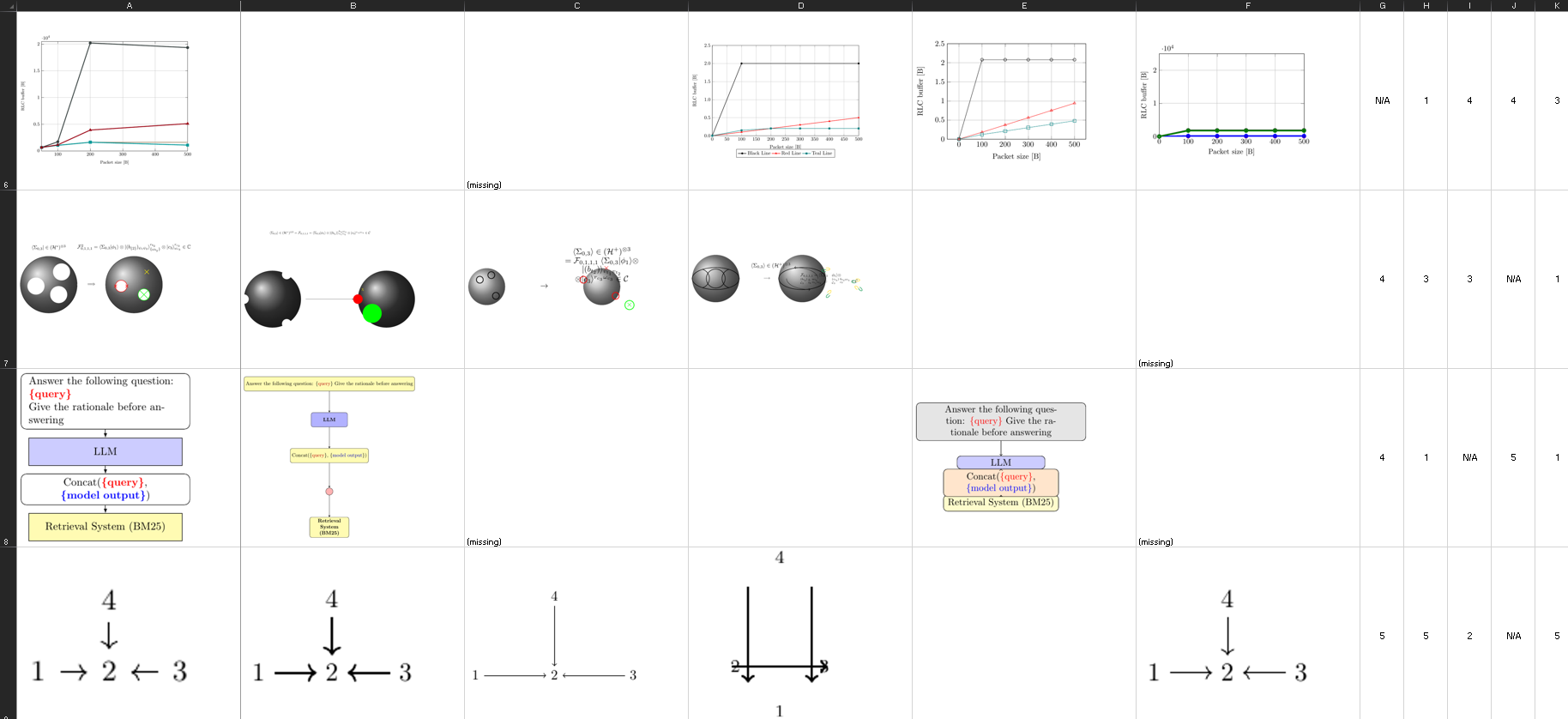}
    \caption{Example of image-image annotations.}
    \label{fig:i2i}
\end{figure}

\paragraph{Ablations}
For inference, we ablate input quality by comparing GPT-4o on the evaluation subset where captions are available (GPT-4o\textsubscript{cap.}) versus the same subset with VLM-generated descriptions instead (GPT-4o\textsubscript{desc.}). For training, we finetune Qwen2.5-3B on different input variants: (i) Qwen2.5-3B (SFT\textsubscript{cap.}), using only caption–TikZ pairs (468k samples), (ii) Qwen2.5-3B (SFT\textsubscript{desc.}), using the same subset but replacing captions with VLM descriptions, (iii) Qwen2.5-3B (SFT\textsubscript{desc.~$\vee$~cap.}), using the full DaTikZ-V4 dataset, but preferring captions whenever they exist, and (iv) Qwen2.5-3B (SFT\textsubscript{desc.~+~cap.}), oversampling by including both descriptions and captions for all samples with paired captions. This setup isolates whether captions add robustness or simply dilute supervision from richer descriptions.

Table~\ref{tab:ablation_data_scources} shows that arXiv data alone achieves strong results (0.305 AVG). Adding GitHub yields further gains (0.320), while TeX SE and synthetic data provide marginal benefits. This highlights that large-scale, naturally occurring TikZ from arXiv and GitHub are the most valuable sources.

\begin{table}[h!]
\centering
\caption{Ablation study of different data sources. Using only data from arXiv already leads to very good performances and arXiv + GitHub almost reaches its full potential.}
\label{tab:ablation_data_scources}
\setlength{\tabcolsep}{4pt}
\begin{tabular}{lcccccc}
\toprule
\textbf{Source} & 
\textbf{CLIP}$\uparrow$ & 
\textbf{DSim}$\uparrow$ &  
\textbf{TED}$\downarrow$ & 
\textbf{AVG}$\uparrow$ & 
\textbf{CR}$\uparrow$ & 
\textbf{AT}$\downarrow$ \\
\midrule
arXiv & 0.152 & 0.568 & \underline{0.805} & 0.305 & 84\% & 550 \\
+ GitHub & 0.158 & 0.605 & \textbf{0.802} & \underline{0.320} & \underline{88\%} & \underline{548} \\
+ TeX SE & \underline{0.159} & \underline{0.608} & 0.806 & \underline{0.320} & \underline{88\%} & 569 \\
All & \textbf{0.161} & \textbf{0.613} & \textbf{0.802} & \textbf{0.324} & \textbf{89\%} & \textbf{529} \\
\bottomrule
\end{tabular}
\end{table}

\subparagraph{Failure Cases}
We conducted a manual inspection of randomly sampled outputs to compare the typical error patterns of TikZilla and GPT-5. Our observations reveal several systematic differences:

\begin{itemize}
    \item \textbf{Compilation.} GPT-5 frequently produces TikZ code that fails to compile for complex scientific figures, often due to missing library imports, incorrect macro nesting, or hallucinated commands. In contrast, our RL-tuned TikZilla models almost always generate syntactically valid code.

    \item \textbf{Code structure.} TikZilla generally relies on basic primitives such as \texttt{\textbackslash node}, \texttt{\textbackslash draw}, and \texttt{\textbackslash fill}, leading to code that is easy to interpret and modify. GPT-5 tends to generate more elaborate constructs (e.g., macros, loops, nested coordinate definitions), which are more compact but also more brittle and error-prone.

    \item \textbf{Category-specific strengths.} TikZilla performs best on diagrams with strong geometric or mathematical constraints—charts, function plots, schematics, and commutative diagrams (\texttt{tikzcd}). GPT-5 performs better on high-level conceptual figures and network style diagrams where spatial layout is loosely specified.
    
    \item \textbf{Effect of RL tuning.} TikZilla without RL exhibits similar structural strengths but more frequent spatial misalignments (e.g., misplaced labels or arrows). RL substantially improves geometric coherence and spatial consistency.
\end{itemize}

\begin{figure*}[htbp]
\centering
\begin{minipage}{\textwidth}
    \begin{tcolorbox}[
        width=\textwidth,
        colback=pink!5,
        colframe=red!40,
        title=Captions vs. Descriptions,
        enhanced,
        sharp corners=south,
    ]
    \textbf{Caption:} Outline of our algorithm for enumerating Williamson sequences of order n. The boxes on the left correspond to the preprocessing which encodes and decomposes the original problem into SAT instances. The boxes on the right correspond to an SMT-like setup where the system that computes the discrete Fourier transform takes on the role of the theory solver.
    
    \textbf{Description:} A block diagram illustrating with several components. There are four main labeled rectangular blocks connected by arrows indicating the direction. At the bottom left, there is an input labeled n entering a rectangular block titled 'Driver script', which sends an arrow labeled 'External call' upward to a block titled 'Diophantine solver / Fourier transform'. From this block another arrow labeled 'Result' points downwards back to the 'Driver script'. From the 'Driver script' a horizontal black arrow point to the right and is labeled 'SAT instances' connected to a block titled 'Programmatic SAT solver'. It outputs a horizontal black arrow labeled 'Enumeration in order n' pointing to the right out of the diagram. Above the 'Programmatic SAT solver' is another block labeled 'Fourier transform' and connected with an upward arrow labeled 'Partial assignment' and a downward arrow labeled 'Conflict clause'. A dashed arrow labeled 'Encoding information' points from the 'Driver script' block back to the \sx{'Diophantine solver / Fourier transform'} \se{to the 'Fourier transform'}.
    
    \vspace{0.5em}
    \begin{center}
    \begin{tabular}{ccc}
    \textbf{Ground Truth} & \textbf{Caption (GPT-4o)} & \textbf{Description (GPT-4o)} \\
    \includegraphics[height=3.5cm]{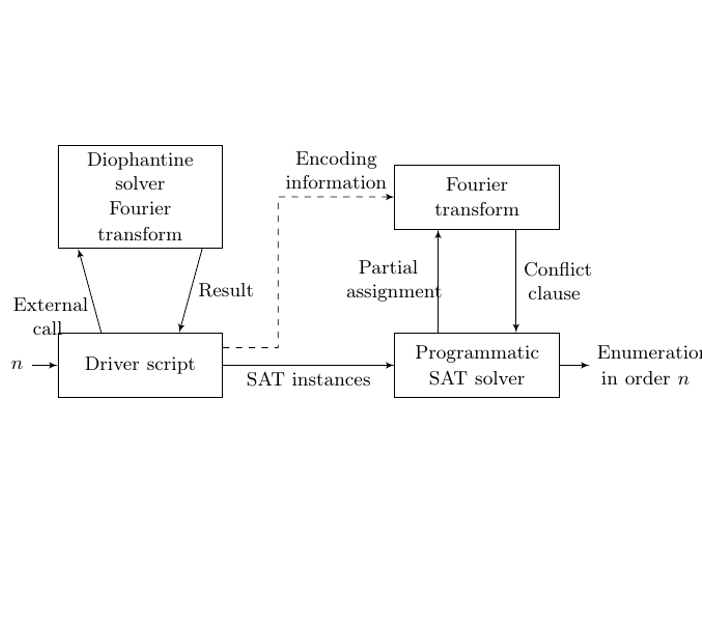} &
    \includegraphics[height=3.5cm]{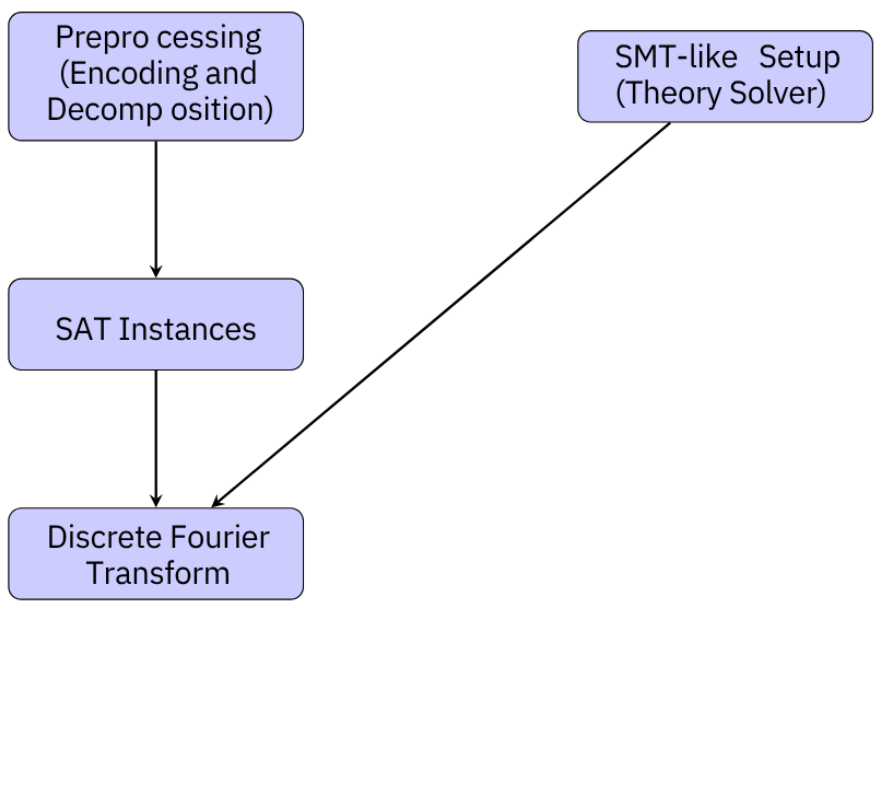} &
    \includegraphics[height=3.5cm]{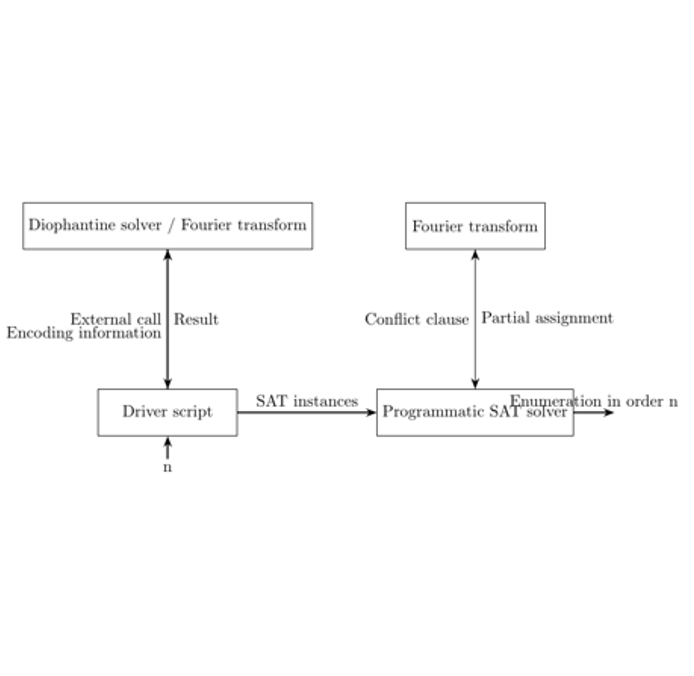} \\
    \bottomrule
    \end{tabular}
    \end{center}

    \textbf{Caption:} A set \lstinline|$\sigma \in PW+$| inside a rectangle R. The blue region \lstinline|$\frac{R}{(\sigma \cup \partial R)}$| can always be triangulated.
    
    \textbf{Description:} A blue rectangle labeled R in the top-left corner. Inside the rectangle, there are two black geometric figures. At the lower-left side, is a layered square pattern composed of three squares, a small black square at the center, surrounded by a blue square matching the background color of the rectangle, surrounded by a larger black square. Diagonally toward the upper-right is an irregular black polygon labeled \lstinline|$\sigma$|. Inside the polygon two shapes have the \sx{black} background color \se{of the rectangle}, one is hexagonal \se{at the top} and the other is diamond shaped \se{at the bottom}.
    
    \vspace{0.5em}
    \begin{center}
    \begin{tabular}{ccc}
    \textbf{Ground Truth} & \textbf{Caption (GPT-4o)} & \textbf{Description (GPT-4o)} \\
    \includegraphics[height=3.5cm]{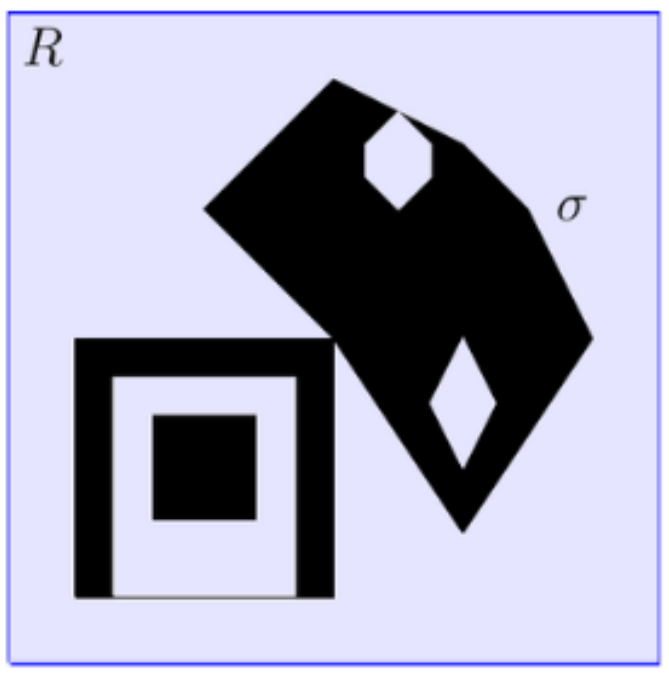} &
    \includegraphics[height=3.5cm]{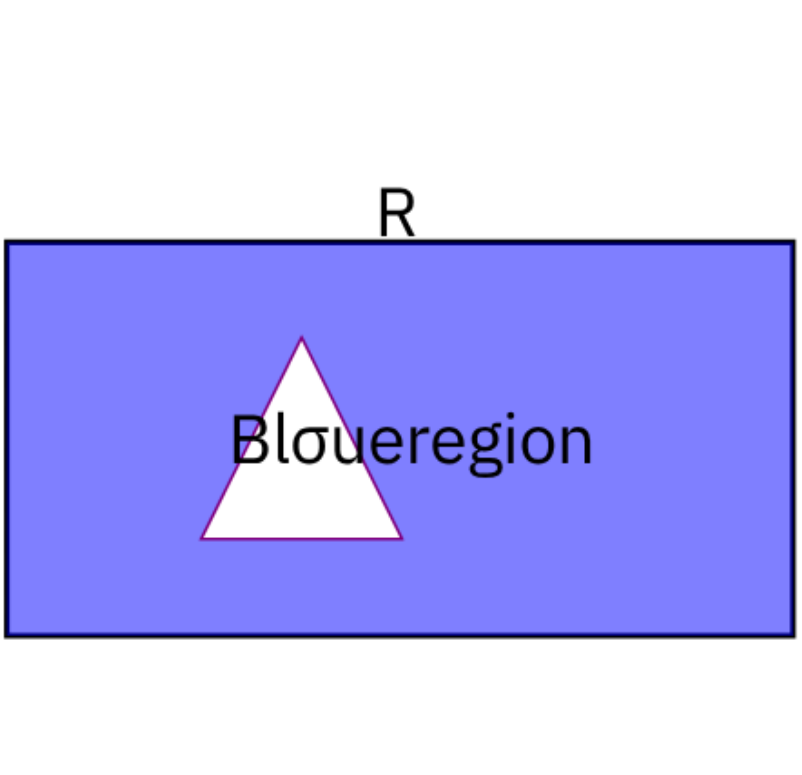} &
    \includegraphics[height=3.5cm]{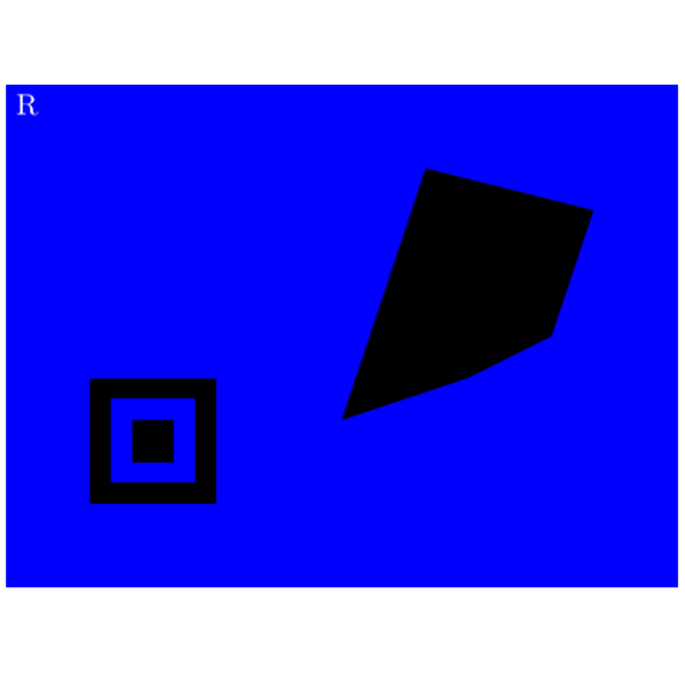} \\
    \end{tabular}
    \end{center}
    \end{tcolorbox}
\end{minipage}
\caption{Captions versus VLM-generated descriptions and their produced figures by GPT-4o. \sx{Striked out} text indicates incorrect VLM-descriptions and \se{red} text indicates improved descriptions made by human annotators. We observe that VLMs most frequently omit low-level stylistic details (e.g., small decorative elements), while hallucinations or omission of key structural elements occur less frequently.}
\label{fig:caption_versus_description}
\end{figure*}

\begin{figure*}[htbp]
\centering
\begin{tcolorbox}[
    width=\textwidth,
    colback=magenta!5,
    colframe=purple!40,
    title=Examples with TikZ Code,
    enhanced,
    sharp corners=south,
    ]
    
    \textbf{Description \& GT Image:}
    \begin{lstlisting}[style=mystyle]
    A large circle centered around the point (0,0). Two points labeled $z^*$ and $w^*$ are placed on 
    the circle near the top. A straight black horizontal line connects both points and is labeled 
    $\theta$ and $2 \sin\left(\frac{\theta}{2}\right)$ and its also exactly aligned with the vertical 
    axes of the circle. From each point $z^*$ and $w^*$ a straight solid line extends downward to the 
    center of the circle. These lines form an isosceles triangle with the lines labeled 1. A vertical 
    dashed line, which is the vertical axes of both the triangle and circle goes from the point (0,0) 
    upwards to the circles circumference and is labeled $\theta$.
    \end{lstlisting}
    \includegraphics[width=0.19\linewidth]{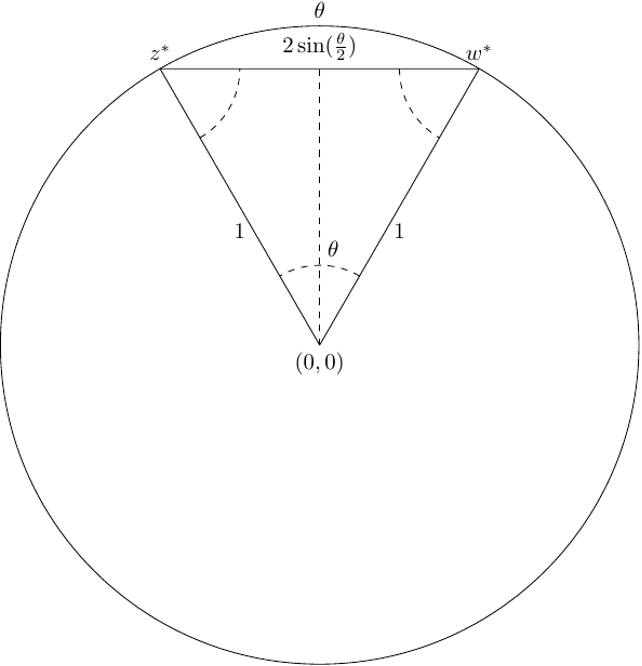}
    
    \textbf{GPT-5:}
    \begin{lstlisting}[style=mystyle]
    \documentclass[tikz]{standalone}
    \usepackage{amsmath}
    \usetikzlibrary{calc}
    \begin{document}
    \begin{tikzpicture}[scale=1]
    \def\\R{3}
    \def\\thetadeg{60}
    \coordinate (O) at (0,0);
    \coordinate (Z) at ({\\R*cos(90-\\thetadeg/2)},{\\R*sin(90-\\thetadeg/2)});
    \coordinate (W) at ({\\R*cos(90+\\thetadeg/2)},{\\R*sin(90+\\thetadeg/2)});
    \coordinate (M) at ($(Z)!0.5!(W)$);
    \draw[thick] (O) circle (\\R);
    \draw[dashed] (O) -- (0,\\R) node[midway, right] {$\\theta$};
    \draw[thick] (O) -- (Z) node[midway, sloped, above] {$1$};
    \draw[thick] (O) -- (W) node[midway, sloped, above] {$1$};
    \draw[thick] (Z) -- (W);
    \node[above] at (M) {$\\theta$};
    \node[below] at (M) {$2\\sin\\!\\left(\\frac{\\theta}{2}\\right)$};
    \fill (Z) circle (2pt) node[above right] {$z^{*}$};
    \fill (W) circle (2pt) node[above left] {$w^{*}$};
    \end{tikzpicture}
    \end{document}
    \end{lstlisting}
    \includegraphics[width=0.19\linewidth]{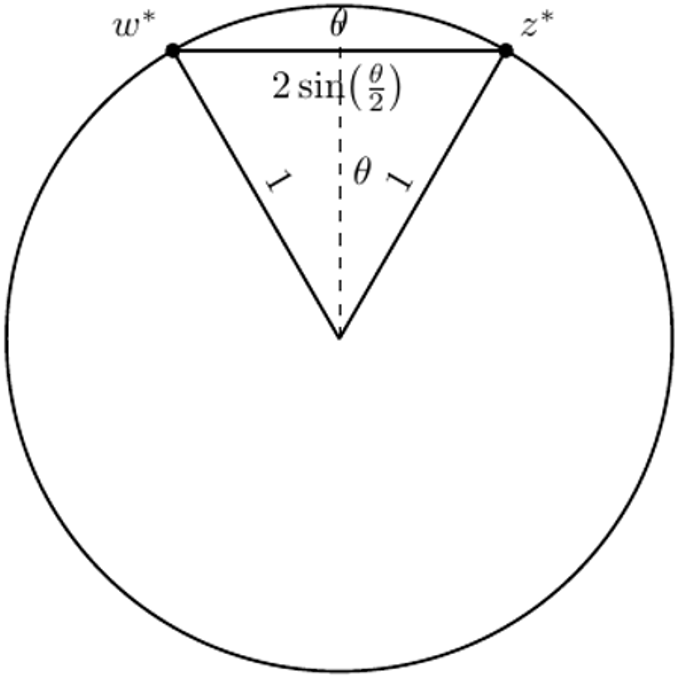}
    
    \medskip
    \textbf{TikZilla-8B-RL:}
    \begin{lstlisting}[style=mystyle]
    \documentclass[tikz]{standalone}
    \usepackage{tikz}
    \usepackage{pgfplots}
    \usetikzlibrary{arrows.meta, decorations.pathreplacing, positioning, shapes, calc}
    \begin{document}
    \begin{tikzpicture}
    \draw (0,0) circle (3cm);
    \fill (0,0) circle (0.05cm) node[below left] {$0$};
    \fill (1.5, 2.598) circle (0.05cm) node[above right] {$z^*$};
    \fill (-1.5, 2.598) circle (0.05cm) node[above left] {$w^*$};
    \draw[thick] (0,0) -- (1.5, 2.598) node[midway, above left] {$1$};
    \draw[thick] (0,0) -- (-1.5, 2.598) node[midway, above right] {$1$};
    \draw[dashed] (0,0) -- (0, 3) node[midway, right] {$\\theta$};
    \node at (0.8, 3.2) {$2\\sin\\left(\\frac{\\theta}{2}\\right)$};
    \end{tikzpicture}
    \end{document}
    \end{lstlisting}
    \medskip
    \includegraphics[width=0.19\linewidth]{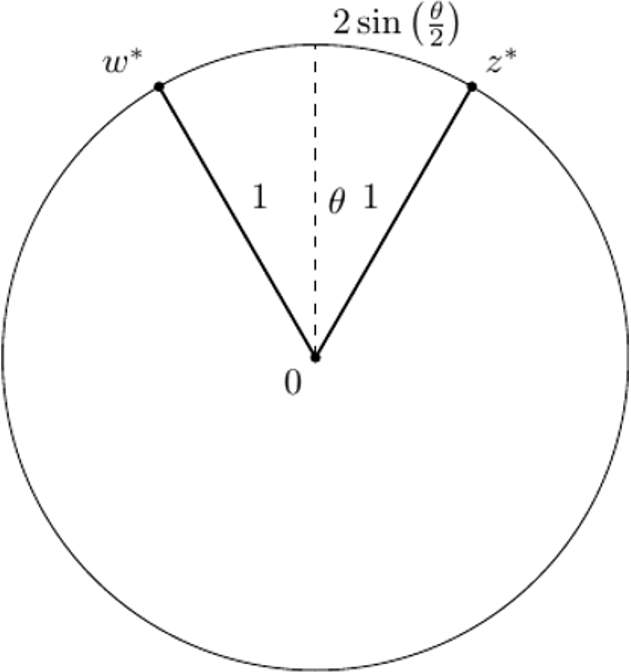}
    
\end{tcolorbox}
\caption{TikZ code and rendered figures shown for GPT-5 and TikZilla-3B-RL using the description above. The code produced by GPT-5 more accurately follows the description compared to TikZilla-3B-RL.}
\label{fig:tikz_code_example_old_old}
\end{figure*}

\begin{figure*}[htbp]
\centering
\begin{tcolorbox}[
    width=\textwidth,
    colback=magenta!5,
    colframe=purple!40,
    title=Examples with TikZ Code,
    enhanced,
    sharp corners=south,
    ]
    
    \textbf{Description \& GT Image:}
    \begin{lstlisting}[style=mystyle]
    A sequence of green circles labeled $X_0$, $X_1$, $X_2$, and $X_{T-1}$ is arranged horizontally 
    from left to right. Each circle is connected to the next by a rightward-pointing arrow labeled 
    $A$. Below each circle, there is a corresponding blue square labeled $O_0$, $O_1$, $O_2$, and 
    $O_{T-1}$, respectively. Each circle is connected to its corresponding square by a vertical black 
    arrow labeled $B$. A dashed red horizontal line runs across the image, intersecting the vertical 
    arrows. The sequence continues with ellipses between $X_2$ and $X_{T-1}$, and after $X_{T-1}$, 
    indicating continuation.
    \end{lstlisting}
    \includegraphics[width=0.45\linewidth]{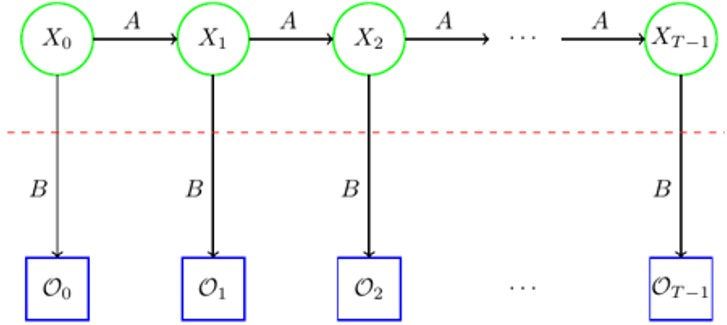}
    
    \textbf{GPT-4o:}
    \begin{lstlisting}[style=mystyle]
    \documentclass[tikz]{standalone}
    \usepackage{amsmath}
    \begin{document}
    \begin{tikzpicture}[node distance=2cm, auto, >=stealth']
    \node[circle, draw=green!50!black, fill=green!20, minimum size=10mm] (X0) {$X_0$};
    \node[circle, draw=green!50!black, fill=green!20, minimum size=10mm, right of=X0] (X1) {$X_1$};
    \node[circle, draw=green!50!black, fill=green!20, minimum size=10mm, right of=X1] (X2) {$X_2$};
    \node[circle, draw=green!50!black, fill=green!20, minimum size=10mm, right of=X2, node distance=3cm] (XT1) {$X_{T-1}$};
    \node[right of=X2, node distance=1.5cm] (dots) {$\\cdots$};
    \node[right of=XT1, node distance=1.5cm] (dots2) {$\\cdots$};
    \node[rectangle, draw=blue!50!black, fill=blue!20, minimum size=10mm, below of=X0] (O0) {$O_0$};
    \node[rectangle, draw=blue!50!black, fill=blue!20, minimum size=10mm, below of=X1] (O1) {$O_1$};
    \node[rectangle, draw=blue!50!black, fill=blue!20, minimum size=10mm, below of=X2] (O2) {$O_2$};
    \node[rectangle, draw=blue!50!black, fill=blue!20, minimum size=10mm, below of=XT1] (OT1) {$O_{T-1}$};
    \draw[->] (X0) -- node[midway, above] {$A$} (X1);
    \draw[->] (X1) -- node[midway, above] {$A$} (X2);
    \draw[->] (X2) -- node[midway, above] {$A$} (dots);
    \draw[->] (dots) -- node[midway, above] {$A$} (XT1);
    \draw[->, thick] (X0) -- node[right] {$B$} (O0);
    \draw[->, thick] (X1) -- node[right] {$B$} (O1);
    \draw[->, thick] (X2) -- node[right] {$B$} (O2);
    \draw[->, thick] (XT1) -- node[right] {$B$} (OT1);
    \draw[dashed, red] ($(O0)!0.5!(O1)$) -- ($(OT1)!0.5!(dots2)$);
    \end{tikzpicture}
    \end{document}
    \end{lstlisting}
    
    \medskip
    \textbf{TikZilla-3B-RL:}
    \begin{lstlisting}[style=mystyle]
    \documentclass[tikz]{standalone}
    \usepackage{tikz, pgfplots}
    \usetikzlibrary{arrows.meta, decorations.pathreplacing, positioning, shapes, calc}
    \begin{document}
    \begin{tikzpicture}
    \node[draw, circle, fill=green!30, minimum size=1cm] (x0) at (0, 0) {$X_0$};
    \node[draw, circle, fill=green!30, minimum size=1cm] (x1) at (2.5, 0) {$X_1$};
    \node[draw, circle, fill=green!30, minimum size=1cm] (x2) at (5, 0) {$X_2$};
    \node at (7.5, 0) {$\\cdots$};
    \node[draw, circle, fill=green!30, minimum size=1cm] (xT) at (10, 0) {$X_{T-1}$};
    \node[draw, rectangle, fill=blue!30, minimum size=0.8cm] (o0) at (0, -2) {$O_0$};
    \node[draw, rectangle, fill=blue!30, minimum size=0.8cm] (o1) at (2.5, -2) {$O_1$};
    \node[draw, rectangle, fill=blue!30, minimum size=0.8cm] (o2) at (5, -2) {$O_2$};
    \node at (7.5, -2) {$\\cdots$};
    \node[draw, rectangle, fill=blue!30, minimum size=0.8cm] (oT) at (10, -2) {$O_{T-1}$};
    \draw[->] (x0) -- (x1) node[midway, above] {$A$};
    \draw[->] (x1) -- (x2) node[midway, above] {$A$};
    \draw[->] (x2) -- (7, 0) node[midway, above] {$A$};
    \draw[->] (8, 0) -- (xT) node[midway, above] {$A$};
    \draw[->] (x0) -- (o0) node[midway, left] {$B$};
    \draw[->] (x1) -- (o1) node[midway, left] {$B$};
    \draw[->] (x2) -- (o2) node[midway, left] {$B$};
    \draw[->] (xT) -- (oT) node[midway, left] {$B$};
    \draw[dashed, red] (-1, -1) -- (11, -1);
    \end{tikzpicture}
    \end{document}
    \end{lstlisting}
    \medskip
    \includegraphics[width=0.55\linewidth]{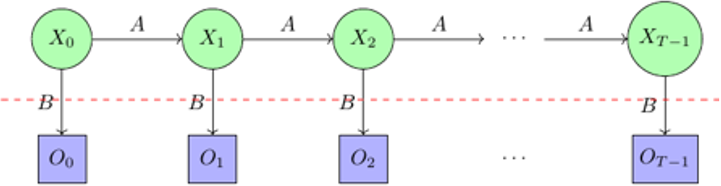}
    
\end{tcolorbox}
\caption{TikZ code and rendered figures shown for GPT-4o and TikZilla-3B-RL using the description above. The code produced by GPT-4o could not be compiled while TikZilla-3B-RL produced correct code.}
\label{fig:tikz_code_example_old}
\end{figure*}

\begin{figure*}[htbp]
\centering
\begin{tcolorbox}[
    width=\textwidth,
    colback=magenta!5,
    colframe=purple!40,
    title=Examples with TikZ Code,
    enhanced,
    sharp corners=south,
    ]
    
    \textbf{Description \& GT Image:}
    \begin{lstlisting}[style=mystyle]
    A sequence of four colored circles is arranged horizontally, each connected by bidirectional 
    arrows. Circle 1 is green and positioned on the far left, labeled with the number "1" centered 
    inside. To its right, circle 2 is yellow, labeled "2" inside. A bidirectional arrow labeled 
    "a" above and "b" below connects circle 1 and circle 2. Next, circle 3 is red, labeled "3" inside, 
    connected to circle 2 by a bidirectional arrow labeled "c" above and "d" below. Finally, circle 4 
    is  blue, labeled "4" inside, connected to circle 3 by a bidirectional arrow labeled "e" above 
    and "f" below. The circles are evenly spaced along the horizontal axis.
    \end{lstlisting}
    \includegraphics[width=0.5\linewidth]{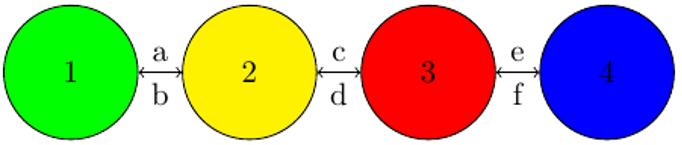}
    
    \textbf{TikZilla-8B:}
    \begin{lstlisting}[style=mystyle]
    \documentclass[tikz{standalone}
    \usepackage{amsmath}
    \usepackage{amssymb}
    \usepackage{colortbl}
    \usepackage{tikz}
    \usetikzlibrary{calc,matrix}
    \usetikzlibrary{decorations.pathmorphing,decorations.markings,decorations.pathreplacing,arrows,shapes}
    \usetikzlibrary{knots}
    \usetikzlibrary{arrows.meta}
    \usepackage[bookmarks=true, bookmarksopen=true]{hyperref}
    \begin{document}
    \begin{tikzpicture}[baseline=-0.5ex]
    \node (1) at (0,0) [] {$1$};
    \node (2) at (2,0) [] {$2$};
    \node (3) at (4,0) [] {$3$};
    \node (4) at (6,0) [] {$4$};
    \draw [->] (1) -- node [above] {$a$} (2);
    \draw [->] (2) -- node [above] {$c$} (3);
    \draw [->] (3) -- node [above] {$e$} (4);
    \draw [<-] (1) -- node [below] {$b$} (2);
    \draw [<-] (2) -- node [below] {$d$} (3);
    \draw [<-] (3) -- node [below] {$f$} (4);
    \draw [fill=green, opacity=0.5] (1) circle (0.25);
    \draw [fill=yellow, opacity=0.5] (2) circle (0.25);
    \draw [fill=red, opacity=0.5] (3) circle (0.25);
    \draw [fill=blue, opacity=0.5] (4) circle (0.25);
    \end{tikzpicture}
    \end{document}
    \end{lstlisting}
    \includegraphics[width=0.6\linewidth]{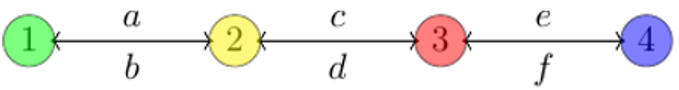}
    
    \medskip
    \textbf{TikZilla-8B-RL:}
    \begin{lstlisting}[style=mystyle]
    \documentclass[tikz]{standalone}
    \usepackage{tikz, amsmath, amssymb, pgfplots, pgffor}
    \usetikzlibrary{automata, positioning, backgrounds, external}
    \tikzset{every state/.style={minimum size=2pt}}
    \begin{document}
    \begin{tikzpicture}
    \node[circle, draw, fill=green!30] (1) at (0,0) {$1$};
    \node[circle, draw, fill=yellow!30] (2) at (2,0) {$2$};
    \node[circle, draw, fill=red!30] (3) at (4,0) {$3$};
    \node[circle, draw, fill=blue!30] (4) at (6,0) {$4$};
    \draw[->] (1) -- node[above] {$a$} (2);
    \draw[->] (2) -- node[above] {$c$} (3);
    \draw[->] (3) -- node[above] {$e$} (4);
    \draw[<-] (1) -- node[below] {$b$} (2);
    \draw[<-] (2) -- node[below] {$d$} (3);
    \draw[<-] (3) -- node[below] {$f$} (4);
    \end{tikzpicture}
    \end{document}
    \end{lstlisting}
    \medskip
    \includegraphics[width=0.6\linewidth]{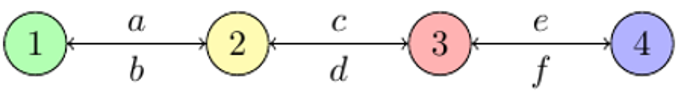}
    
\end{tcolorbox}
\caption{TikZ code and rendered figures shown for TikZilla-8B and TikZilla-8B-RL using the description above. Despite both figures being correct, the code produced by TikZilla-8B-RL is much shorter compared to TikZilla-8B.} 
\label{fig:tikz_code_example}
\end{figure*}

\begin{figure*}[htbp]
\centering
\begin{tcolorbox}[
    width=\textwidth,
    colback=magenta!5,
    colframe=purple!40,
    title=Examples with TikZ Code,
    enhanced,
    sharp corners=south,
    ]
    
    \textbf{Description \& GT Image:}
    \begin{lstlisting}[style=mystyle]
    A Cartesian coordinate system with x-axis labeled 'x' and y-axis labeled 'y' contains three 
    distinct curves. The x-axis ranges from -3 to 3, and the y-axis ranges from -2 to 2, with grid 
    lines at each integer. The first curve, labeled 'y = \text{ReLU}(x)' in red, is a piecewise 
    linear function starting at the origin, extending horizontally along the x-axis for negative 
    x-values, and rising linearly with a slope of 1 for positive x-values. The second curve, 
    labeled 'y = \tanh x' in blue, is an S-shaped curve passing through the origin, asymptotically 
    approaching y = 1 as x approaches positive infinity and y = -1 as x approaches negative 
    infinity. The third curve, labeled 'y = \sigma(x)' in orange, is another S-shaped curve, similar 
    to the blue curve but with a more gradual slope, also passing through the origin and 
    asymptotically approaching y = 1 and y = 0 for positive and negative x-values, respectively. 
    The labels for the curves are positioned to the right of the graph, with the red label at the 
    top, followed by the blue label, and the orange label at the bottom.
    \end{lstlisting}
    \includegraphics[width=0.45\linewidth]{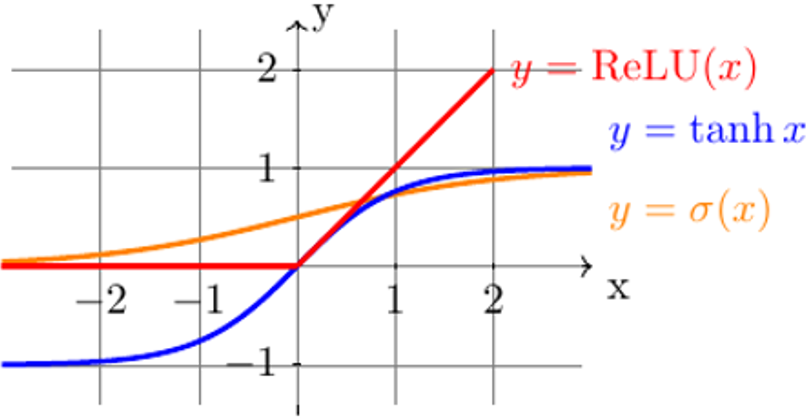}
    
    \textbf{TikZilla-8B:}
    \begin{lstlisting}[style=mystyle]
    \documentclass[tikz]{standalone}
    \usepackage[utf8{inputenc}
    \usepackage{amsmath}
    \usepackage{amssymb}
    \usepackage[dvipsnames,table,xcdraw]{xcolor}
    \usepackage{tikz}
    \usetikzlibrary{cd, positioning, arrows.meta, backgrounds, calc, 3d}
    \usepackage[many,breakable]{tcolorbox}
    \begin{document}
    \begin{tikzpicture}[domain=-2.75:2.75]
    \draw[very thin,color=gray] (-2.9,-1.2) grid (3,2.9);
    \draw[->] (-3,0) -- (3,0) node[right] {$x$};
    \draw[->] (0,-2.2) -- (0,3) node[above] {$y$};
    \draw[color=RedOrange,thick]    plot (\\x,\\x) node[anchor=south west,text width=1.5cm] {$y=\\operatorname{ReLU} (x)$};
    \draw[color=Blue,thick]   plot (\\x,{0.5*(exp(\\x)-exp(-\\x))}) node[below right,text width=1.5cm] {$y=\\tanh x$};
    \draw[color=Orange,thick]   plot (\\x,{0.5*(1+sgn(\\x))}) node[below right,text width=1.5cm] {$y=\\sigma(x)$};
    \end{tikzpicture}
    \end{document}
    \end{lstlisting}
    
    \medskip
    \textbf{TikZilla-8B-RL:}
    \begin{lstlisting}[style=mystyle]
    \documentclass[tikz]{standalone}
    \usepackage{tikz, pgfplots, amsmath}
    \usetikzlibrary{arrows.meta, decorations.pathreplacing, positioning, shapes, calc}
    \begin{document}
    \begin{tikzpicture}
    \draw[->] (-3, 0) -- (3, 0) node[right] {$x$};
    \draw[->] (0, -2) -- (0, 2) node[above] {$y$};
    \draw[help lines, gray!30] (-3,-2) grid (3,2);
    \draw[thick, red] (-3, 0) -- (0, 0) -- (3, 3);
    \node[red, right] at (3.2, 1.5) {$y = \\text{ReLU}(x)$};
    \draw[thick, blue] plot[domain=-3:3, samples=100] (\\x, {tanh(\\x)});
    \node[blue, right] at (3.2, -0.5) {$y = \\tanh x$};
    \draw[thick, orange] plot[domain=-3:3, samples=100] (\\x, {1/(1 + exp(-\\x))});
    \node[orange, right] at (3.2, 0.5) {$y = \\sigma(x)$};
    \node[below] at (-2, 0) {-2};
    \node[below] at (-1, 0) {-1};
    \node[below] at (1, 0) {1};
    \node[below] at (2, 0) {2};
    \node[left] at (0, -1) {-1};
    \node[left] at (0, 1) {1};
    \end{tikzpicture}
    \end{document}
    \end{lstlisting}
    \medskip
    \includegraphics[width=0.45\linewidth]{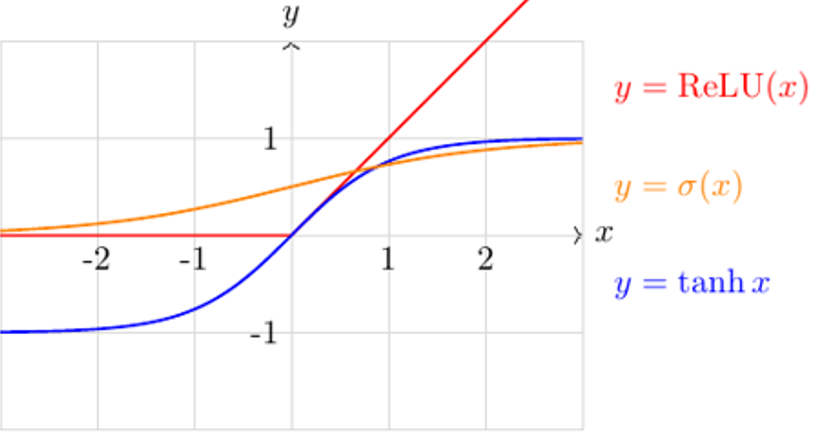}
    
\end{tcolorbox}
\caption{TikZ code and rendered figures shown for TikZilla-8B and TikZilla-8B-RL using the description above. The code produced by TikZilla-8B could not be compiled while TikZilla-8B-RL produced correct code.}
\label{fig:tikz_code_example_new}
\end{figure*}

\setlength{\aboverulesep}{0pt}
\setlength{\belowrulesep}{0pt}
\setlength{\tabcolsep}{4pt}
\renewcommand{\arraystretch}{1.03}
\setlength{\abovetopsep}{0pt}

\newcommand{\exrowappone}[5]{%
  \desc{#1} & \imgcell{#2} & \imgcellgood{#3} & \imgcell{#4} & \imgcellverygood{#5} \\
}
\newcommand{\exrowapptwo}[5]{%
  \desc{#1} & \imgcell{#2} & \imgcellgood{#3} & \imgcellokay{#4} & \imgcellverygood{#5} \\
}
\newcommand{\exrowappthree}[5]{%
  \desc{#1} & \imgcell{#2} & \imgcellokay{#3} & \imgcell{#4} & \imgcellgood{#5} \\
}
\newcommand{\exrowappfour}[5]{%
  \desc{#1} & \imgcell{#2} & \imgcell{#3} & \imgcellbad{#4} & \imgcellokay{#5} \\
}
\newcommand{\exrowappfive}[5]{%
  \desc{#1} & \imgcell{#2} & \imgcellgood{#3} & \imgcellbad{#4} & \imgcellokay{#5} \\
}
\newcommand{\exrowappsix}[5]{%
  \desc{#1} & \imgcell{#2} & \imgcellokay{#3} & \imgcellokay{#4} & \imgcellgood{#5} \\
}
\newcommand{\exrowappseven}[5]{%
  \desc{#1} & \imgcell{#2} & \imgcell{#3} & \imgcellgood{#4} & \imgcellgood{#5} \\
}

\begin{table}[t]
\centering
\caption{Exemplary scientific TikZ figures produced by one baseline LLM (GPT-4o) and two of our finetuned LLMs (TikZilla-3B and TikZilla-3B-RL) using the prompts from the first column which have been VLM augmented based on the Ground Truth figures in the second column. \legendbox{green}-boxed figures have been rated as very good, \legendbox{yellow} as good, \legendbox{orange} as bad, and \legendbox{red} as very bad by human annotators. Empty cells indicate non-compilable TikZ code.}
\label{tab:examples_1}
\begin{tabularx}{\textwidth}{YCCCC}
    \toprule
    \headercell{Prompt} & \headercell{Ground Truth} & \headercell{GPT-4o} & \headercell{TikZilla-3B} & \headercell{TikZilla-3B-RL} \\
    \midrule

    \exrowappone{A series of black lines connect two vertical columns of elements. The left column contains labels $x_1$, $x_2$, $x_3$, $x_4$, and $x_n$, arranged vertically from top to bottom with equal spacing. The right column contains shaded rectangles labeled $z_1$, $z_2$, $z_3$, and $z_m$, also arranged vertically from top to bottom with equal spacing. Each label in the left column is connected by straight black lines to multiple rectangles in the right column, forming a network of intersecting lines. Dotted ellipses are placed vertically between $x_4$ and $x_n$ and between $z_3$ and $z_m$, indicating continuation. The labels $x_1$, $x_2$, $x_3$, $x_4$, and $x_n$ are positioned to the left of their respective lines, while the labels $z_1$, $z_2$, $z_3$, and $z_m$ are centered within their rectangles.}
        {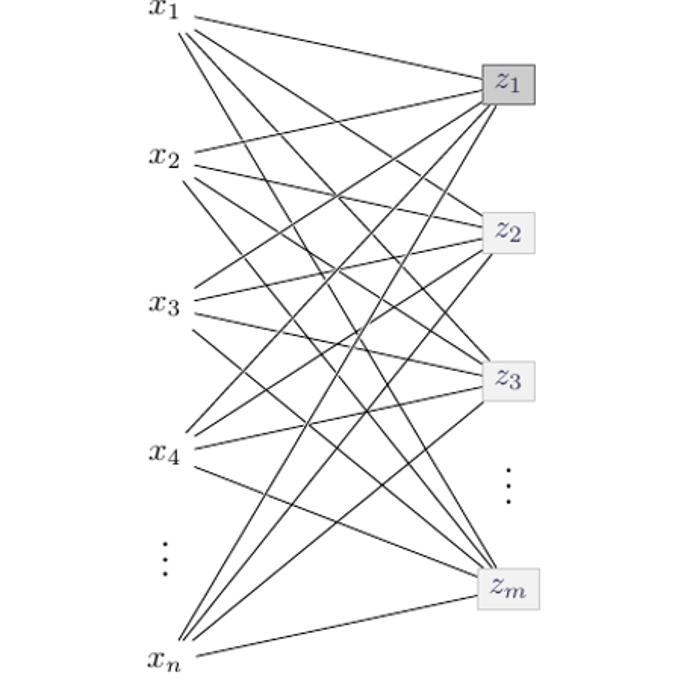}
        {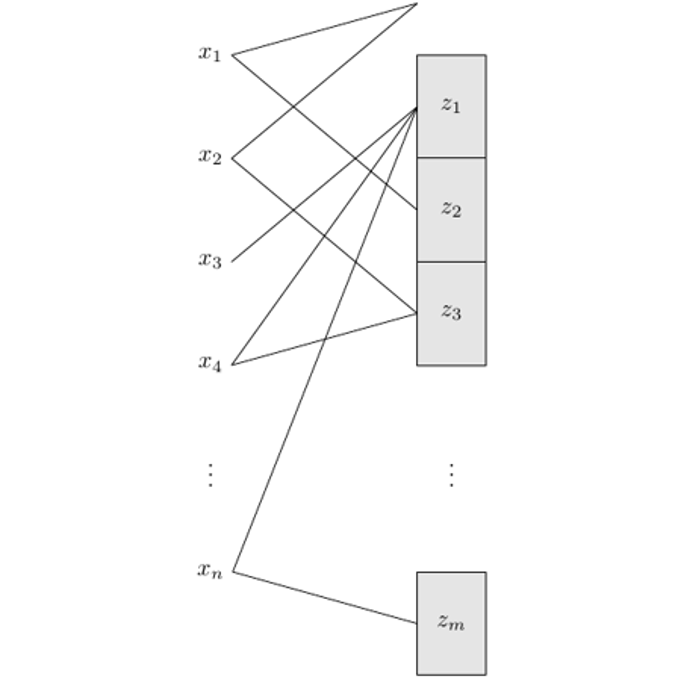}
        {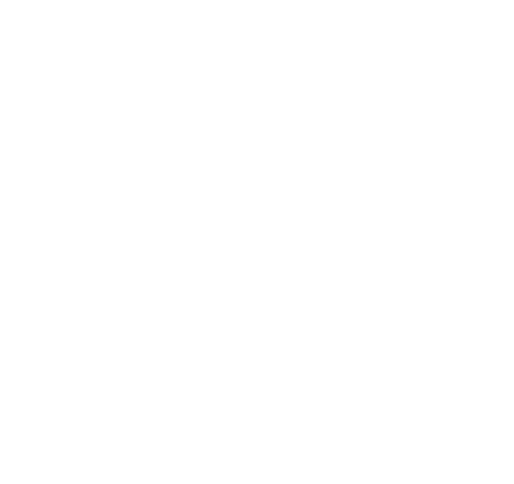}
        {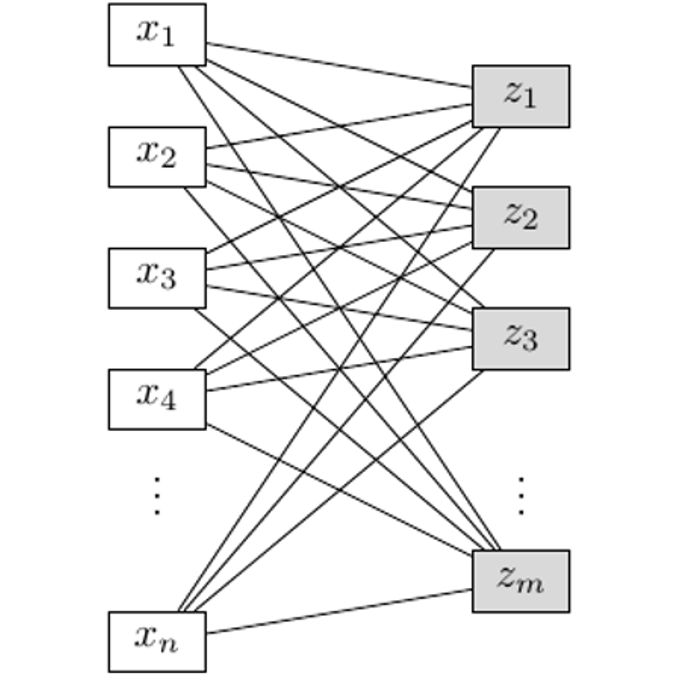}
    \midrule
    \exrowapptwo{The bar chart displays accuracy percentages on the y-axis ranging from 80\% to 100\% with increments of 10\%, labeled "Accuracy (\%)" on the left. The x-axis is labeled "Number of talkers" and includes five categories: 0, 1, 2, 3, and 4. Each category contains three vertical bars. The first bar is black, representing "MPVAD-SC," the second bar is blue, representing "MPVAD-MC," and the third bar is red, representing "MPVAD-F." Above each bar, there is a numerical label indicating the exact accuracy percentage. For category 0, the black bar is labeled 80, the blue bar 82, and the red bar 85. For category 1, the black bar is labeled 81, the blue bar 83, and the red bar 86. For category 2, the black bar is labeled 82, the blue bar 84, and the red bar 87. For category 3, the black bar is labeled 83, the blue bar 85, and the red bar 88. For category 4, the black bar is labeled 84, the blue bar 86, and the red bar 89. A legend is positioned at the top right corner of the chart, indicating the color and label for each bar type. The chart background includes horizontal dashed lines at each 10\% increment on the y-axis.}
        {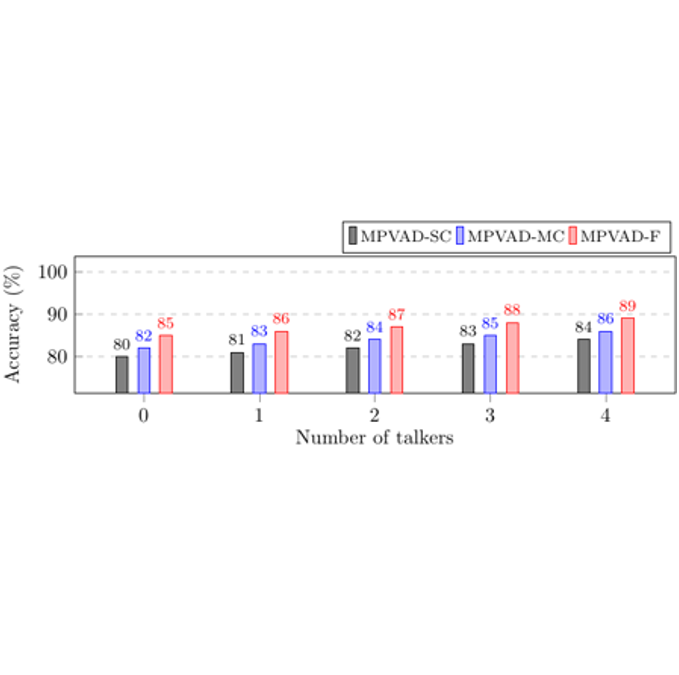}
        {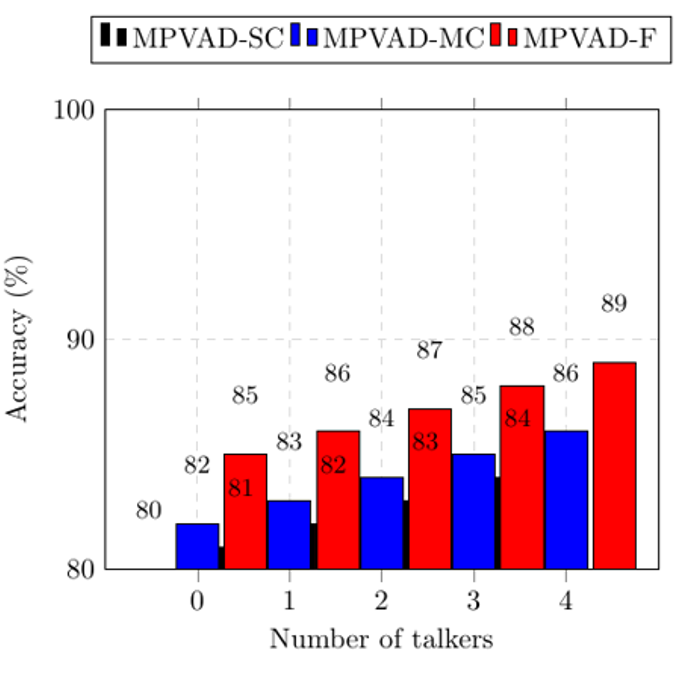}
        {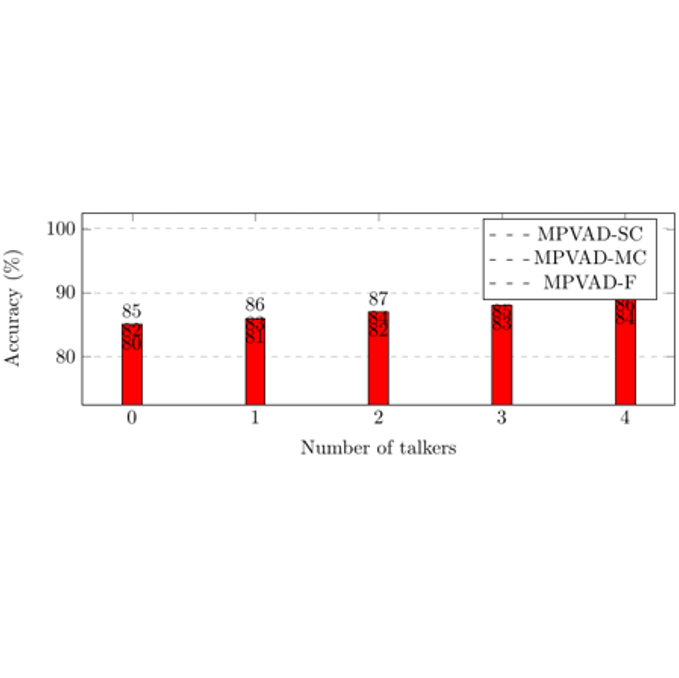}
        {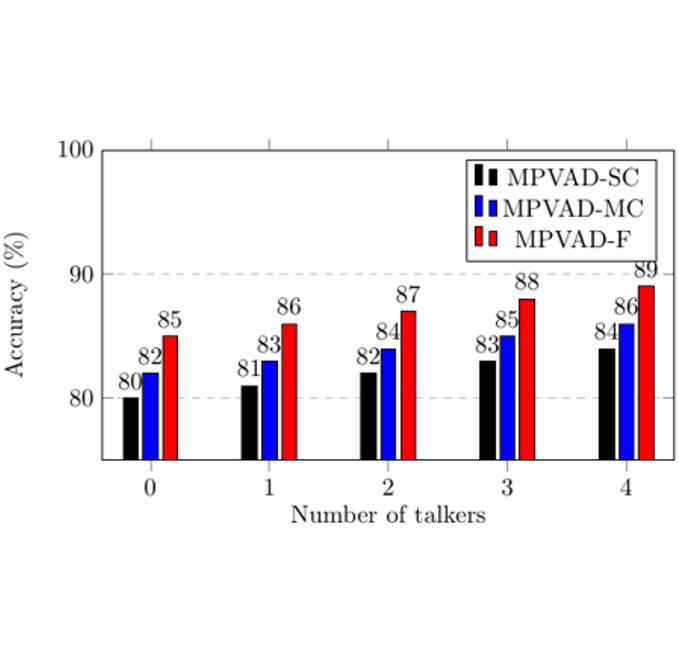}
    \midrule
    \exrowappthree{A diagram consists of several labeled arrows and nodes arranged in a structured format. At the top left, node $\Gamma_i$ is connected by a rightward arrow labeled $\vdash P$ to node $\Xi_i$. From $\Xi_i$, a rightward arrow labeled $\vdash Q$ leads to node $\Psi_i$. Below $\Gamma_i$, node $\exists_i m$ is connected by a downward arrow to node $\Gamma$. From $\Gamma$, a rightward arrow labeled $\exists_i l$ leads to a central node marked with a circle containing a plus sign. This central node is connected by a rightward arrow labeled $\exists_j l$ to node $\exists_j n$. From $\exists_j n$, a rightward arrow labeled $\Delta$ leads to node $\Psi_j$. Below $\Gamma$, node $\exists_j m$ is connected by a downward arrow to node $\Gamma_j$. From $\Gamma_j$, a rightward arrow labeled $\vdash P[j/i]$ leads to node $\Xi_j$. From $\Xi_j$, a rightward arrow labeled $\vdash Q[j/i]$ leads to node $\Psi_j$. A dotted arrow labeled $\exists_i$ connects $\Xi_i$ to the central node, and another dotted arrow labeled $\exists_j$ connects the central node to $\Xi_j$. A vertical arrow labeled $\exists_i n$ connects $\Psi_i$ to $\exists_j n$, and a vertical arrow labeled $\exists_j n$ connects $\exists_j n$ to $\Psi_j$. A horizontal dotted arrow labeled $\Delta_i = \ell z$ connects $\exists_j n$ to $\Psi_j$.}
        {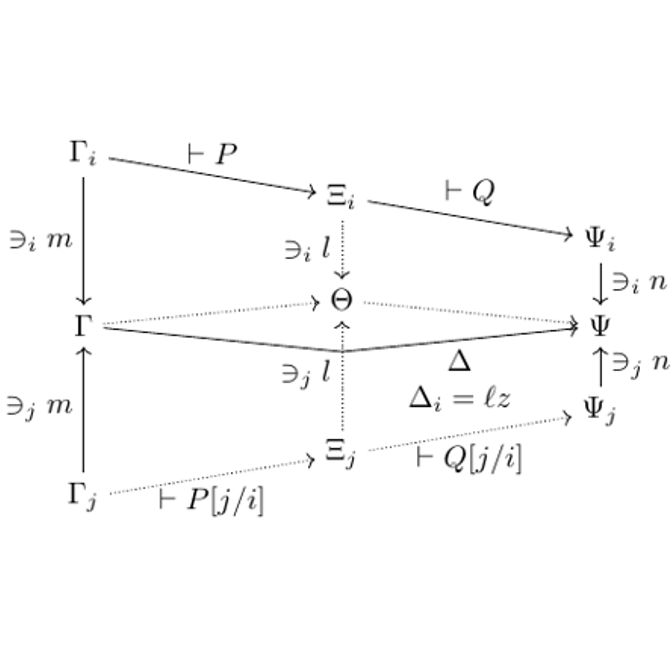}
        {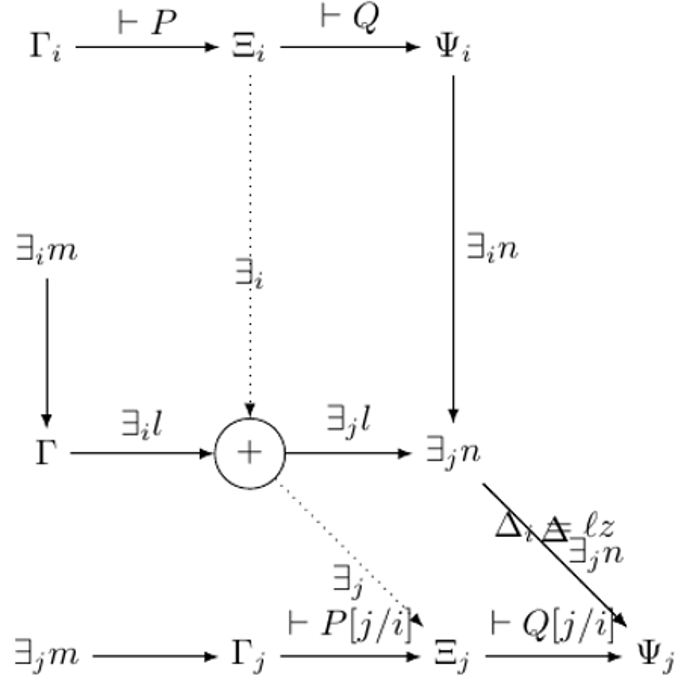}
        {structure/figures/not_compiled.png}
        {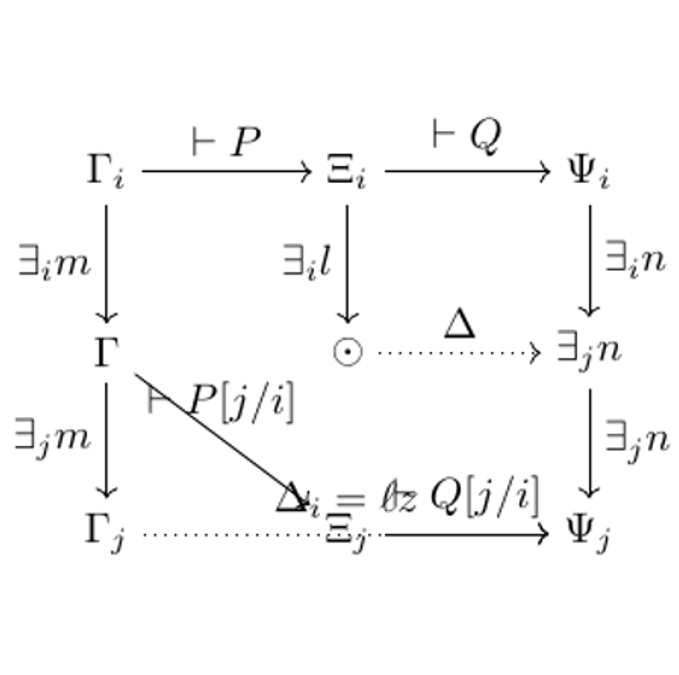}
    \midrule
    \exrowappfour{A control system diagram features a horizontal line starting from the left with a label $r(t)$, leading to a summation circle. The summation circle has a minus sign on the left and is labeled $e^o(t)$ on the right. From the summation circle, a horizontal line extends rightward into a dashed blue rectangle labeled $C(\alpha)$ at the bottom right. Inside the rectangle, there are three vertically aligned blocks labeled $C(\theta_1)$, $C(\theta_k)$, and $C(\theta_N)$ from top to bottom. Each block has a horizontal line extending rightward to a corresponding triangular amplifier labeled $\alpha_1$, $[\alpha]_k$, and $\alpha_N$. The outputs of these amplifiers converge at a summation circle on the right side of the rectangle. From this circle, a horizontal line labeled $u(t)$ extends rightward to a block labeled $G$. A horizontal line continues from $G$ to the right, labeled $y^o(t)$. A feedback line loops from $y^o(t)$ back to the summation circle, completing the system.}
        {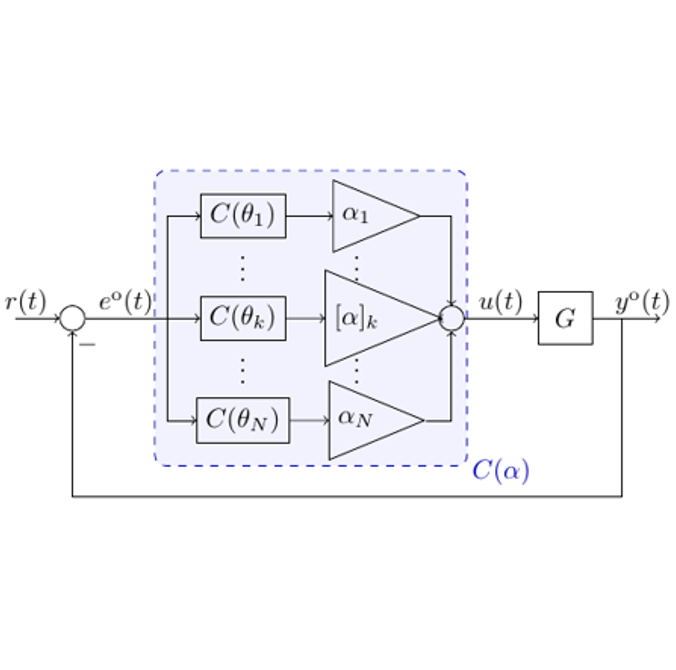}
        {structure/figures/not_compiled.png}
        {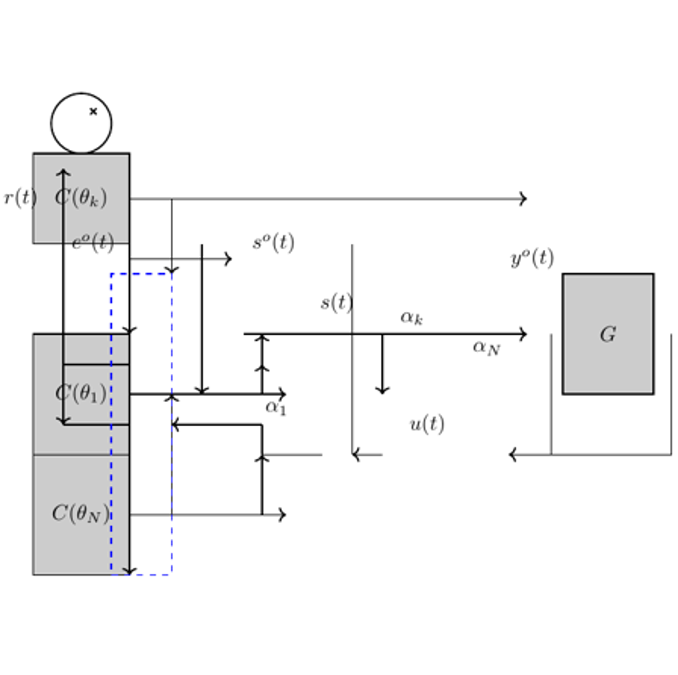}
        {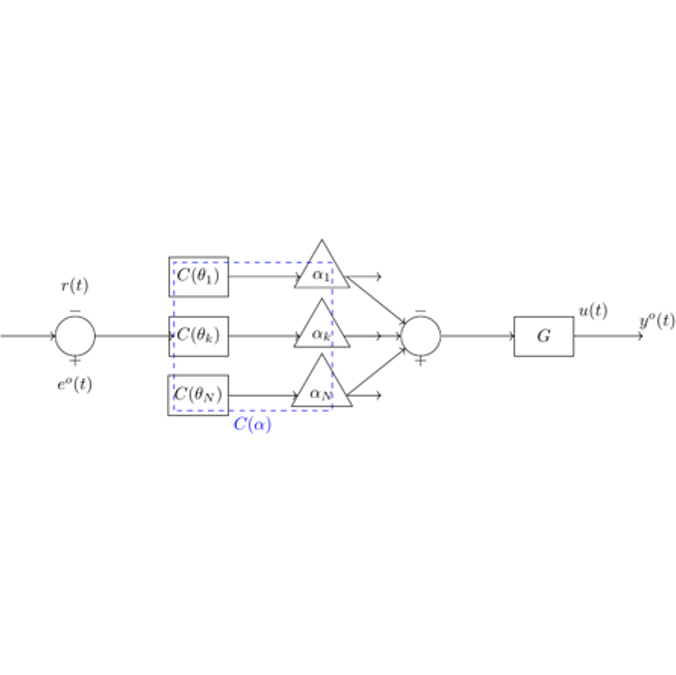}
    \midrule
    \exrowappfive{A rectangular diagram is enclosed by a dashed border with rounded corners. Inside, there are two main vertical paths. The left path begins with a downward arrow labeled "Cond$(\tilde{N}, \tilde{T})$" leading to a rectangle labeled "Up/Down$(\tilde{N}, \tilde{T})$". Below, another downward arrow connects to a rectangle labeled "Conv$_K(N, T)$", followed by another downward arrow leading to a rectangle labeled "LeakyReLU(0.2)". The right path starts with a downward arrow labeled "Input$(N, T)$" leading to a rectangle labeled "ChannelNorm". Below, a downward arrow connects to a circle with a dot inside, representing a multiplication operation. The left path has a rightward arrow from "Conv$_K(N, T)$" connecting to a rectangle labeled "Conv$_K$". This rectangle has a rightward arrow leading to the multiplication circle on the right path. Below the multiplication circle, a downward arrow leads to a circle with a plus inside, representing an addition operation. The left path continues with a rightward arrow from "LeakyReLU(0.2)" connecting to another rectangle labeled "Conv$_K$". This rectangle has a rightward arrow leading to the addition circle on the right path. Below the addition circle, a downward arrow leads to a label "Output$(N, T)$". The entire diagram is divided into two sections by a vertical dashed line, with the left section containing the "Cond$(\tilde{N}, \tilde{T})$" path and the right section containing the "Input$(N, T)$" path.}
        {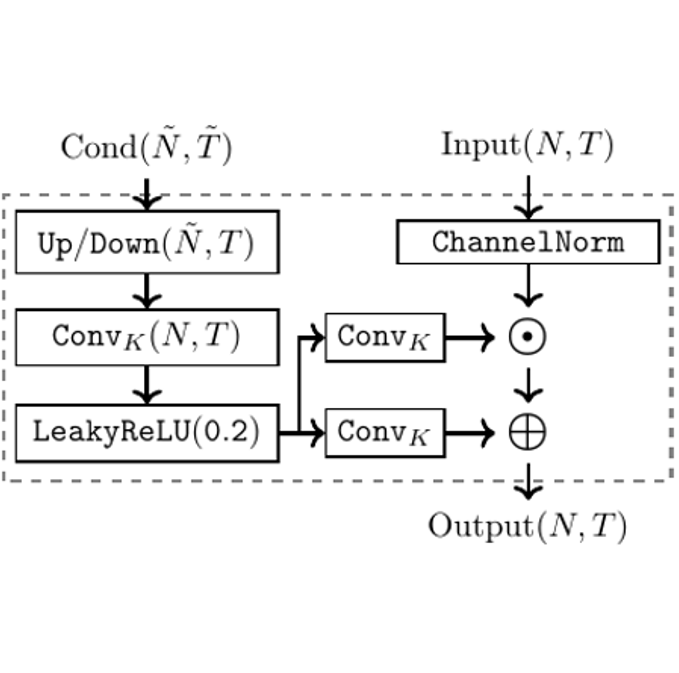}
        {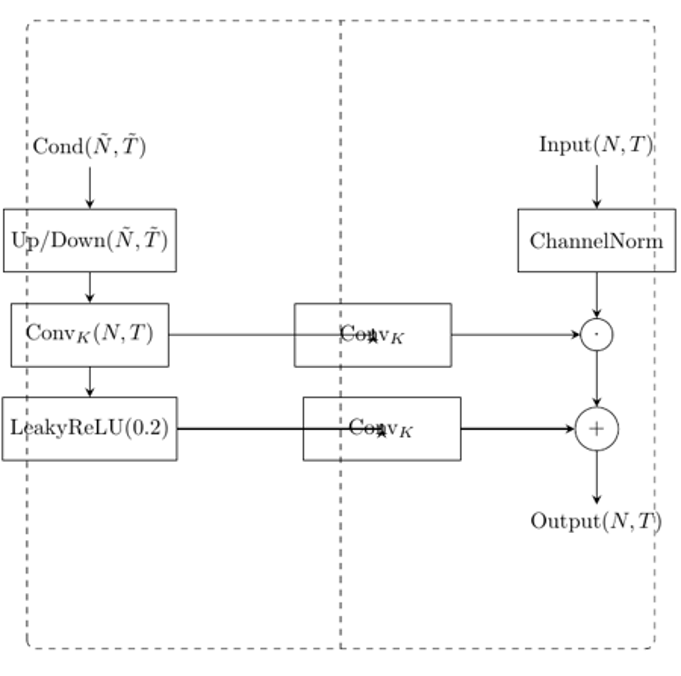}
        {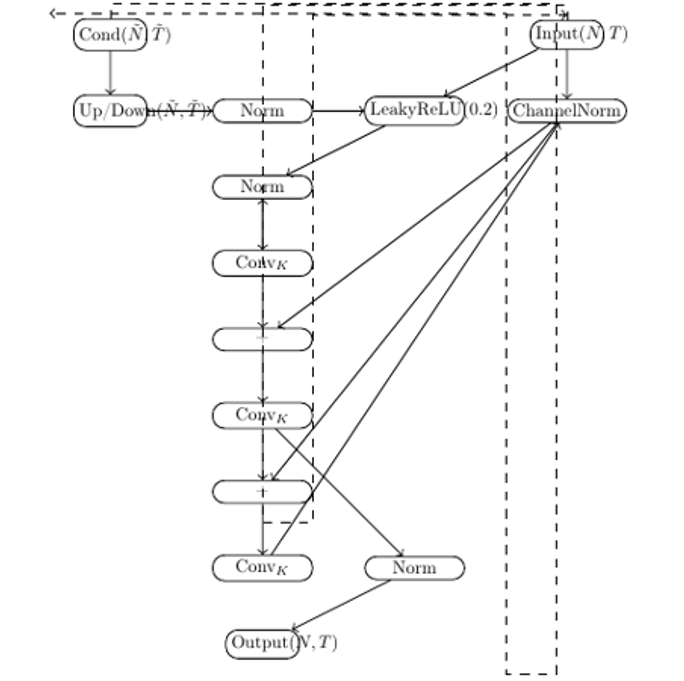}
        {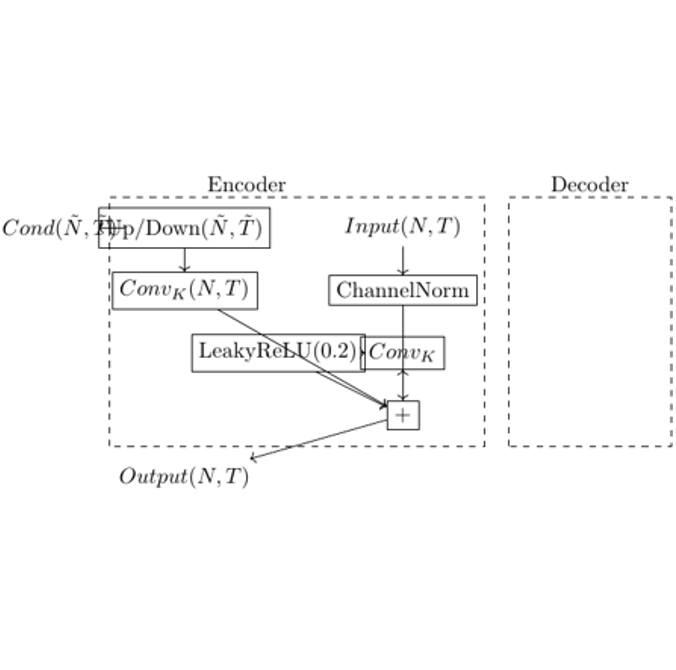}
    \midrule
    \exrowappsix{A black irregular polygon labeled $(P)_K$ is centered in the image. Seven black arrows labeled $(U_K)_{\sigma_1}$ through $(U_K)_{\sigma_7}$ point outward from each vertex of the polygon, with labels positioned near the arrowheads. To the right of the polygon, a set of equations is displayed in black text. The equations are vertically aligned and read as follows: $\mathcal{F}_K = \{\sigma_i\}_{i=1}^7$, $\mathcal{U}^{\text{ext}}_K = \{(U_K)_{\sigma_i}\}_{i=1}^7$, $\mathcal{T}_K = \{\kappa_i\}_{i=1}^7$, $\mathcal{F}^{\text{ext}}_K = \{\sigma_i\}_{i=1}^7$, $\mathcal{F}^{\text{int}}_K = \{\sigma_i\}_{i=8}^{14}$, and $\mathcal{F}_h = \mathcal{F}^{\text{ext}}_K \cup \mathcal{F}^{\text{int}}_K$. The text is right-aligned and positioned to the right of the polygon.}
        {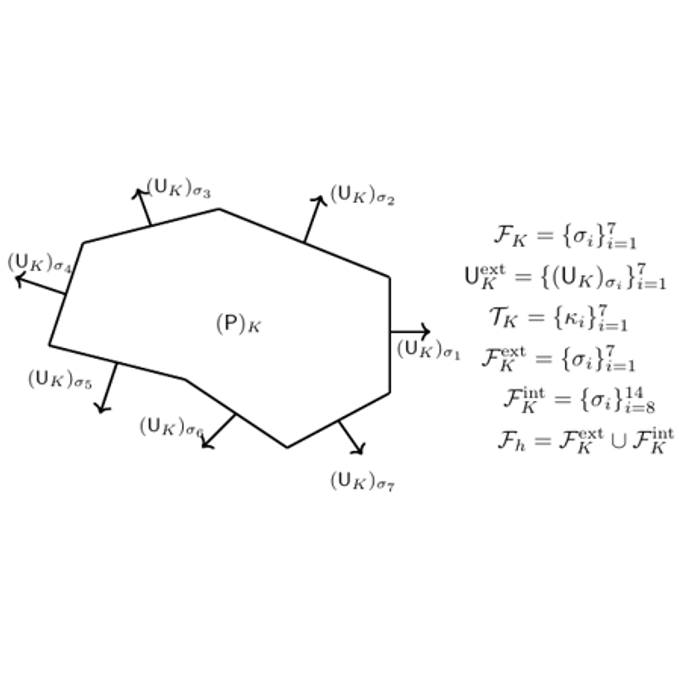}
        {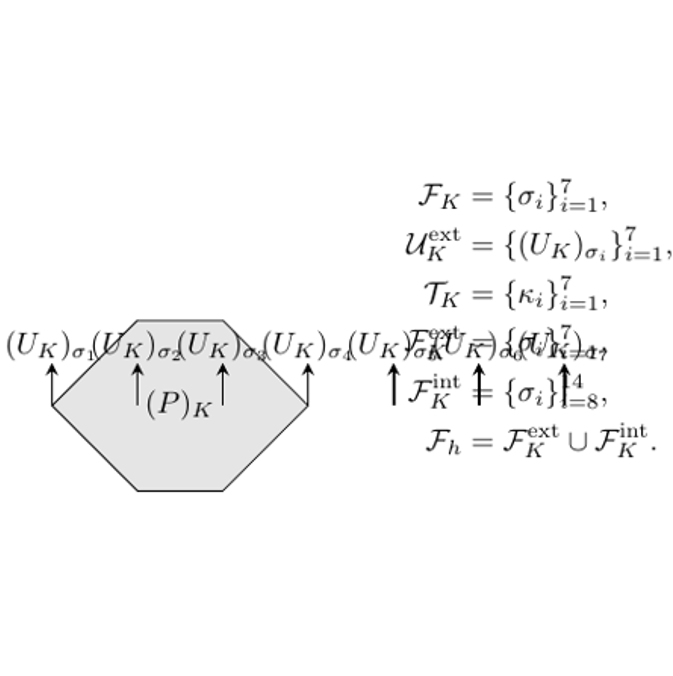}
        {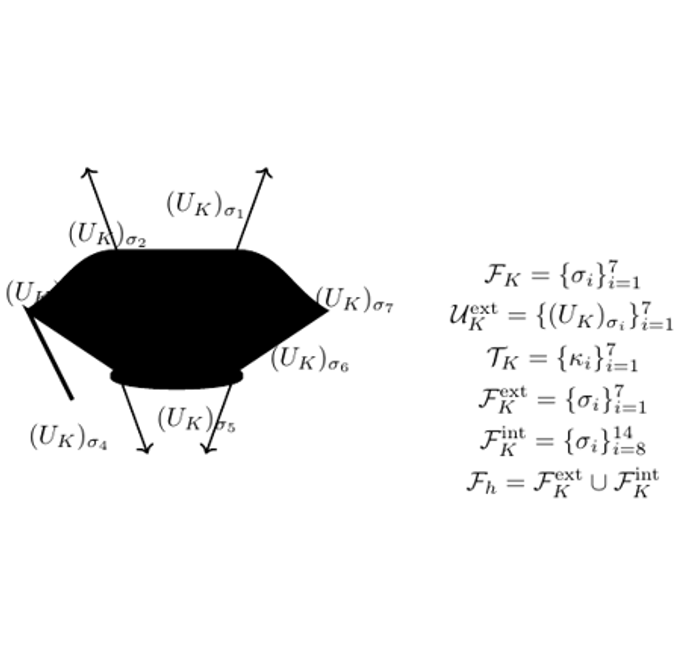}
        {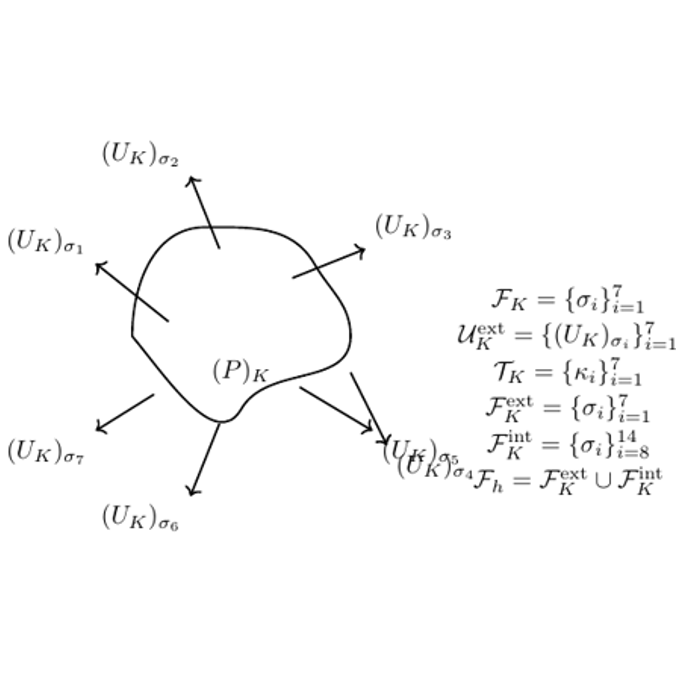}
    \midrule
    \exrowappseven{State diagram with two circles labeled $q_0$ and $q_2$. Circle $q_0$ is on the left, connected to circle $q_2$ on the right by a horizontal arrow labeled "true | int saved = 0; int x'_0, \ldots, int x'_n; | B" with the label "init" above the arrow. Circle $q_2$ has a loop arrow on its right side labeled "cond | assert(\pi); op; | A" with the label "loop_head" above the loop. Below the diagram, a blue rectangular box contains two lines of text. The first line reads "op \equiv \text{if}(\text{nondet}() \wedge \text{saved} = 0)\{ x'_0 = x_0; \ldots; x'_n = x_n; \text{saved} = 1;\}" and the second line reads "\pi \equiv (\text{saved} = 1) \implies (x'_0 \neq x_0 \lor x'_1 \neq x_1 \lor \cdots \lor x'_n \neq x_n)".}
        {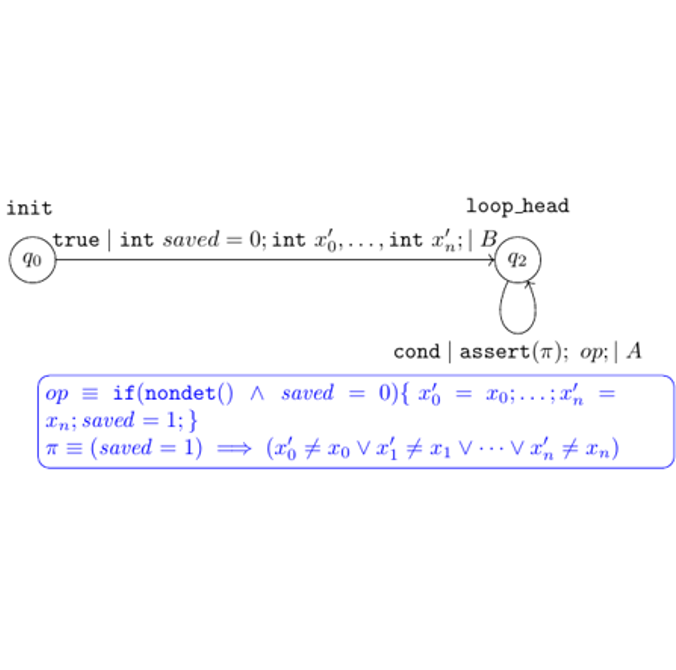}
        {structure/figures/not_compiled.png}
        {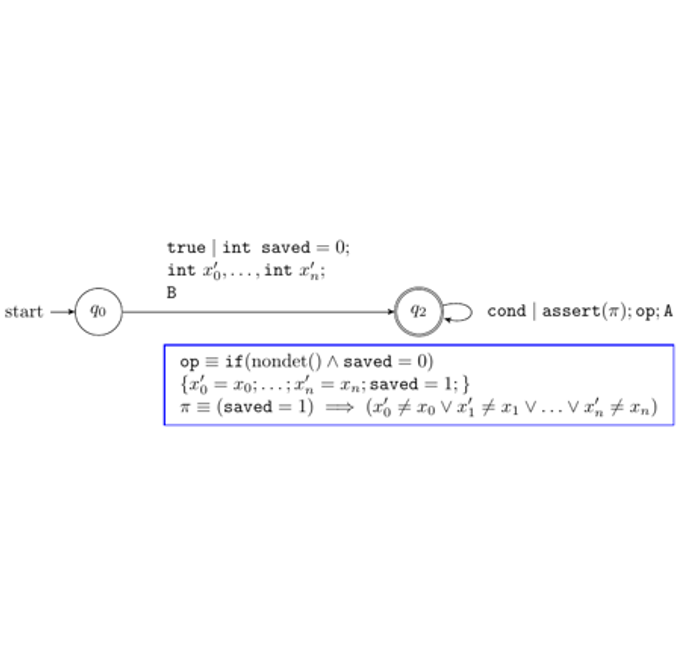}
        {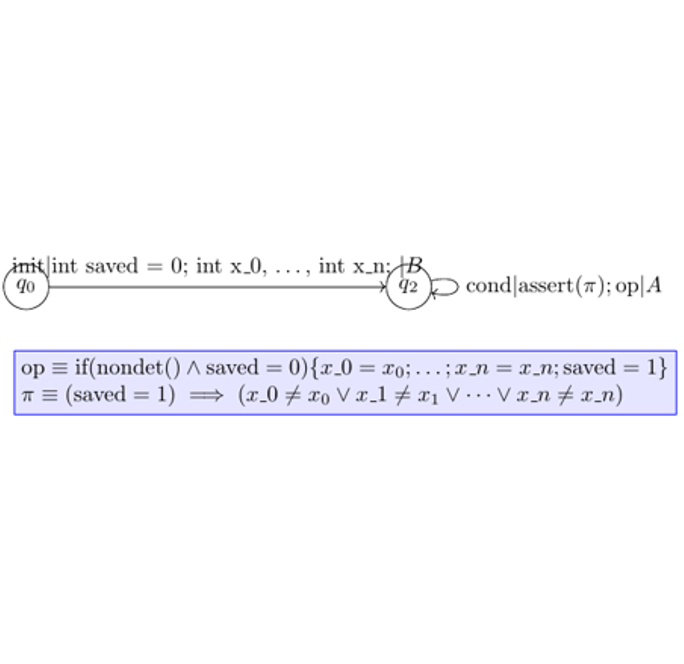}
    \bottomrule 
\end{tabularx}
\end{table}

\setlength{\aboverulesep}{0pt}
\setlength{\belowrulesep}{0pt}
\setlength{\tabcolsep}{4pt}
\renewcommand{\arraystretch}{1.03}
\setlength{\abovetopsep}{0pt}

\newcommand{\exrowapptwoone}[5]{%
  \desc{#1} & \imgcell{#2} & \imgcellverygood{#3} & \imgcellgood{#4} & \imgcellgood{#5} \\
}
\newcommand{\exrowapptwotwo}[5]{%
  \desc{#1} & \imgcell{#2} & \imgcellverygood{#3} & \imgcellgood{#4} & \imgcellverygood{#5} \\
}
\newcommand{\exrowapptwothree}[5]{%
  \desc{#1} & \imgcell{#2} & \imgcellgood{#3} & \imgcellokay{#4} & \imgcellgood{#5} \\
}
\newcommand{\exrowapptwofour}[5]{%
  \desc{#1} & \imgcell{#2} & \imgcellverygood{#3} & \imgcellokay{#4} & \imgcellgood{#5} \\
}
\newcommand{\exrowapptwofive}[5]{%
  \desc{#1} & \imgcell{#2} & \imgcellbad{#3} & \imgcell{#4} & \imgcellokay{#5} \\
}
\newcommand{\exrowapptwosix}[5]{%
  \desc{#1} & \imgcell{#2} & \imgcell{#3} & \imgcell{#4} & \imgcellokay{#5} \\
}
\newcommand{\exrowapptwoseven}[5]{%
  \desc{#1} & \imgcell{#2} & \imgcell{#3} & \imgcellbad{#4} & \imgcellokay{#5} \\
}

\begin{table}[t]
\centering
\caption{Exemplary scientific TikZ figures produced by one baseline LLM (GPT-5) and two of our finetuned LLMs (TikZilla-8B, and TikZilla-8B-RL) using the prompts from the first column which have been VLM augmented based on the Ground Truth figures in the second column. \legendbox{green}-boxed figures have been rated as very good, \legendbox{yellow} as good, \legendbox{orange} as bad, and \legendbox{red} as very bad by human annotators. Empty cells indicate non-compilable TikZ code.}
\label{tab:examples_2}
\begin{tabularx}{\textwidth}{YCCCC}
    \toprule
    \headercell{Prompt} & \headercell{Ground Truth} & \headercell{GPT-5} & \headercell{TikZilla-8B} & \headercell{TikZilla-8B-RL} \\
    \midrule

    \exrowapptwoone{A red rectangle on the left labeled with $\\mu_\\alpha$ at the top, $T$ in the middle, and $\\epsilon_{k\\alpha}(t)$ at the bottom. A blue rectangle on the right labeled with $\\mu_\\beta$ at the top, $T$ in the middle, and $\\epsilon_{k\\beta}$ at the bottom. Between the rectangles, a red circle labeled $\\epsilon_A$ is on the left, and a blue circle labeled $\\epsilon_B$ is on the right. A black arrow labeled $\\Gamma_\\alpha$ points from the red rectangle to the red circle, and another black arrow labeled $\\Gamma_\\beta$ points from the blue circle to the blue rectangle. A dashed black line labeled $U$ connects the red circle to the blue circle, with arrows pointing in both directions.}
        {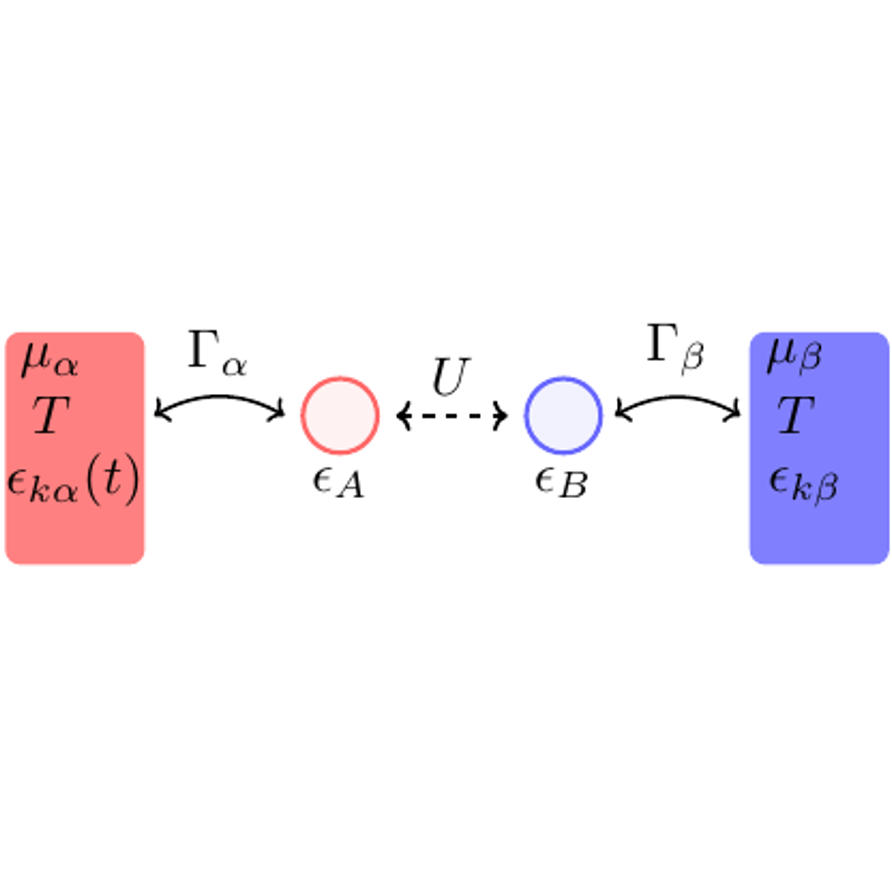}
        {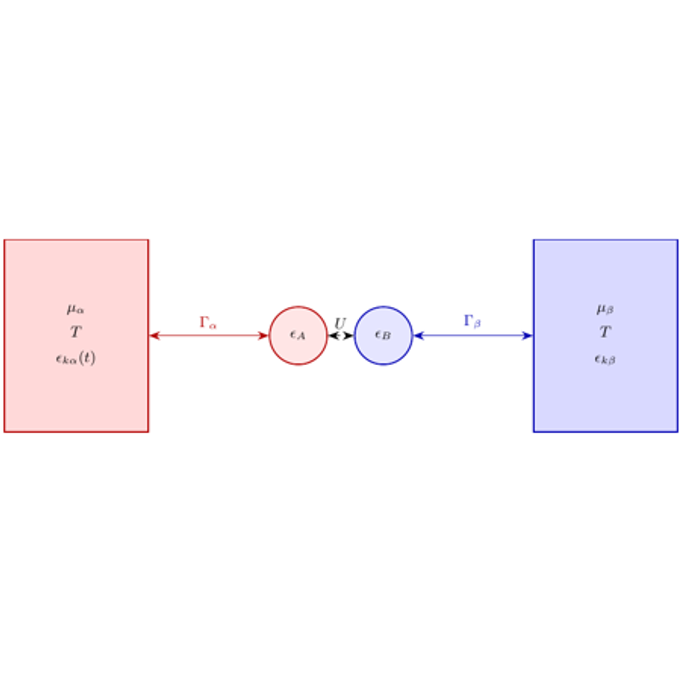}
        {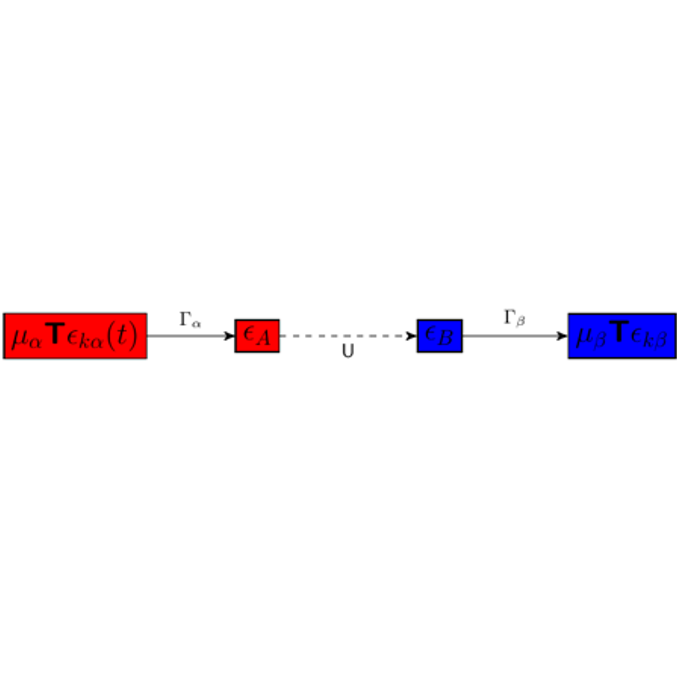}
        {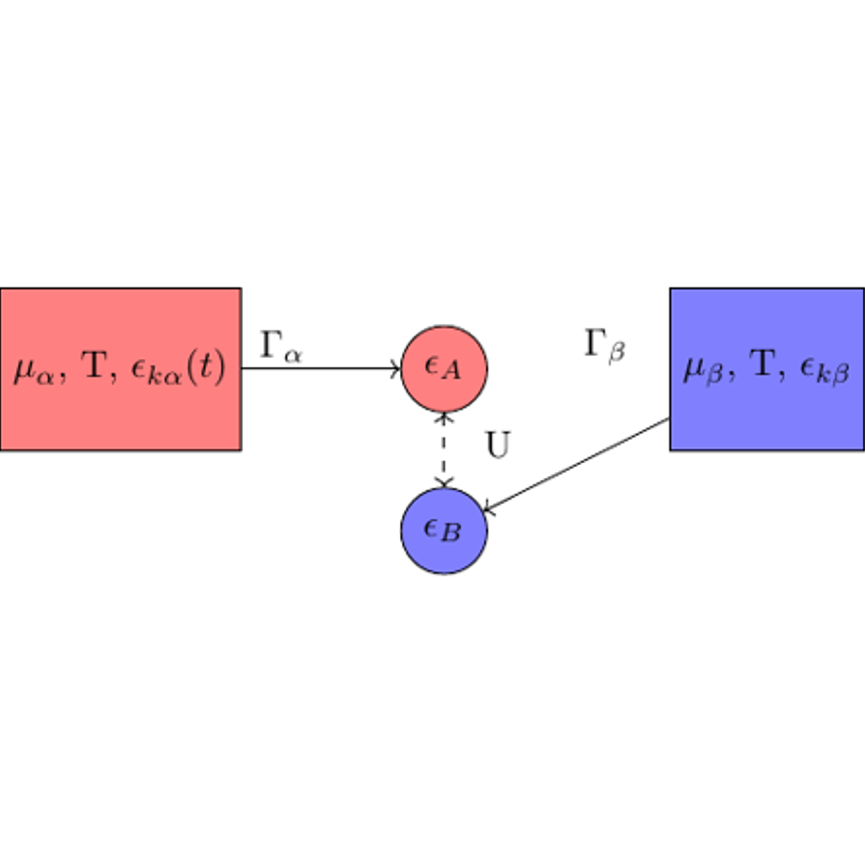}
    \midrule
    \exrowapptwofive{A flowchart with a series of connected shapes. At the top left, an oval labeled \"Input\" connects with a arrow to a rectangle labeled \"Initialization.\" This rectangle connects with a downward arrow to another rectangle labeled \"Grid variation,\" which is inside a larger gray rectangle. The gray rectangle is labeled \"Computation\" on the left side. Below \"Grid variation,\" a downward arrow leads to a rectangle labeled \"Generation of Feasible Operation Region,\" followed by another downward arrow leading to a rectangle labeled \"Generation of Feasible Planning Region.\" A downward arrow from this rectangle points to a diamond labeled \"Additional Grid?\" with two arrows branching from it. The leftward arrow labeled \"False\" leads to an oval labeled \"Output.\" The rightward arrow labeled \"True\" loops back to the top of the gray rectangle, connecting to the rectangle labeled \"Grid variation.\".}
        {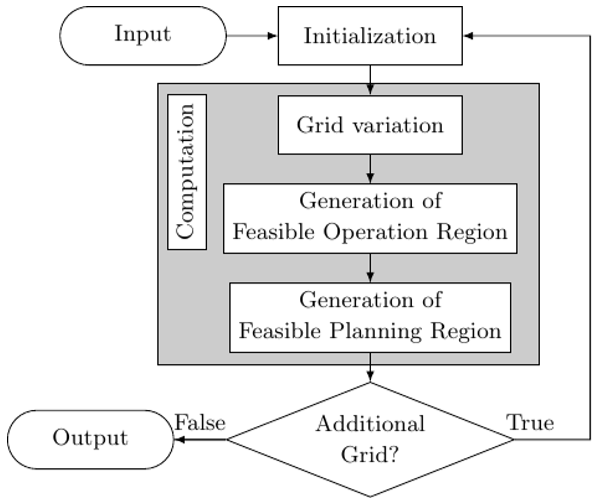}
        {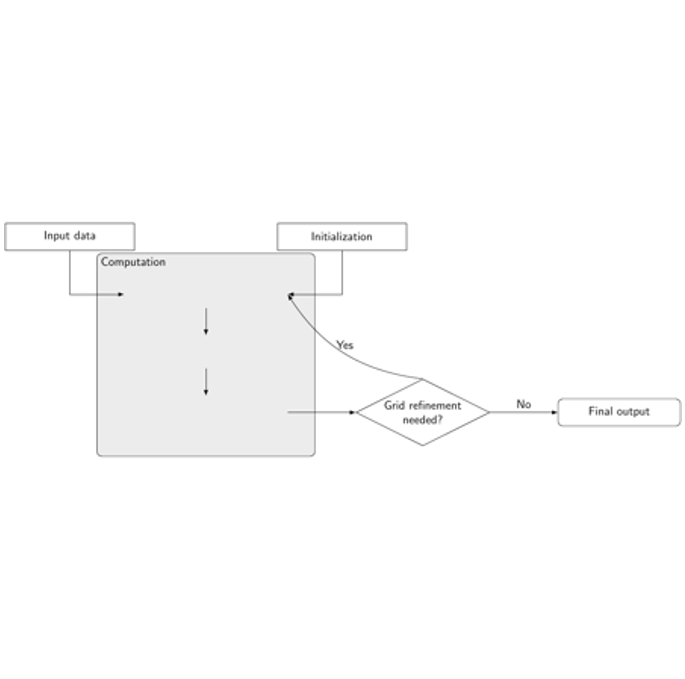}
        {structure/figures/not_compiled.png}
        {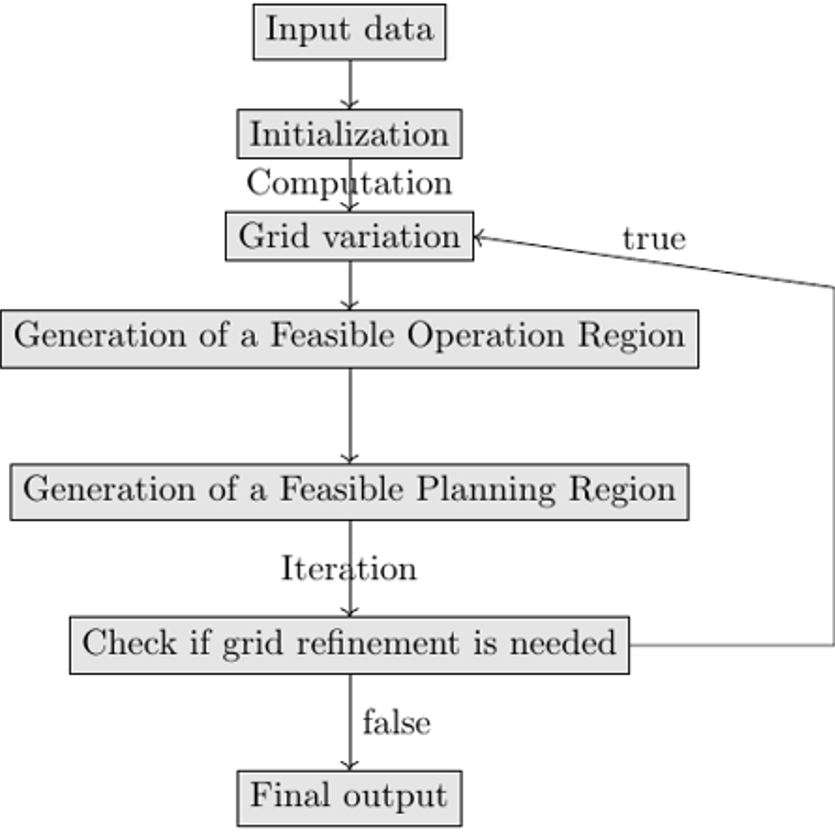}
    \midrule
    \exrowapptwotwo{A horizontal black arrow extends from left to right, labeled $u_i$ at the tip. Above the arrow, three adjacent colored rectangles are aligned horizontally. The first rectangle on the left is yellow, labeled $\\emptyset$ in black at its center. The second rectangle is magenta, labeled DA + RD in black at its center. The third rectangle is cyan, labeled DA + $\\sim$RD in black at its center. Below the arrow, three vertical black tick marks intersect the arrow. The first tick mark is labeled $u^d$ directly below the yellow rectangle, the second tick mark is labeled $u^r$ directly below the magenta rectangle, and the third tick mark is at the base of the arrowhead.}
        {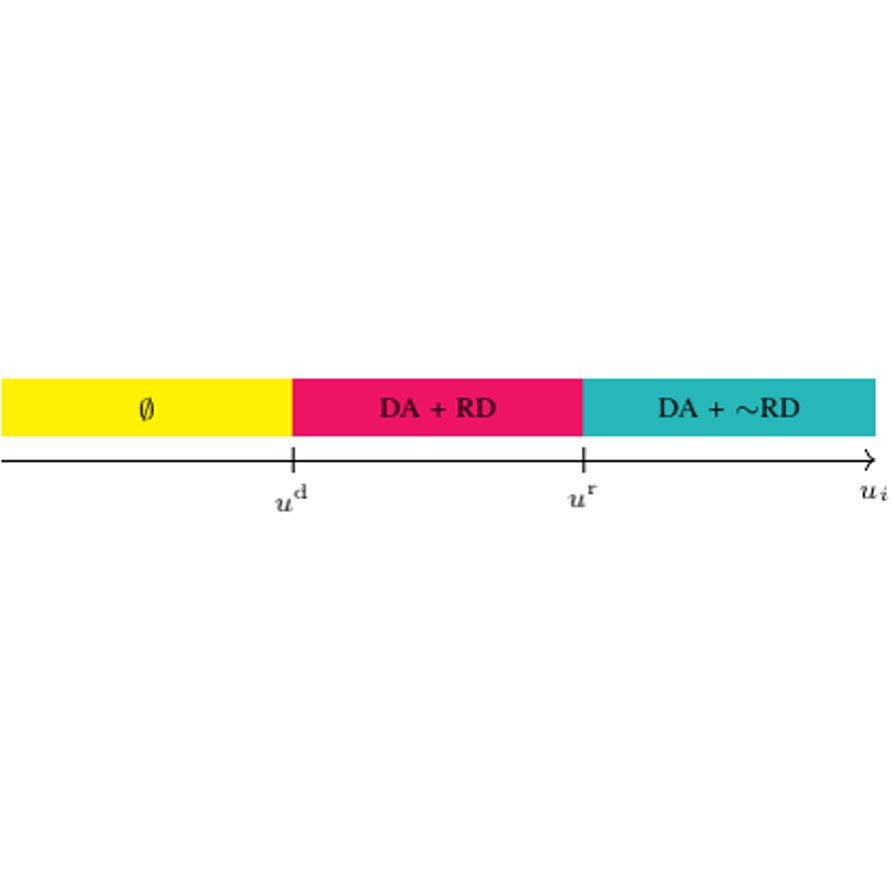}
        {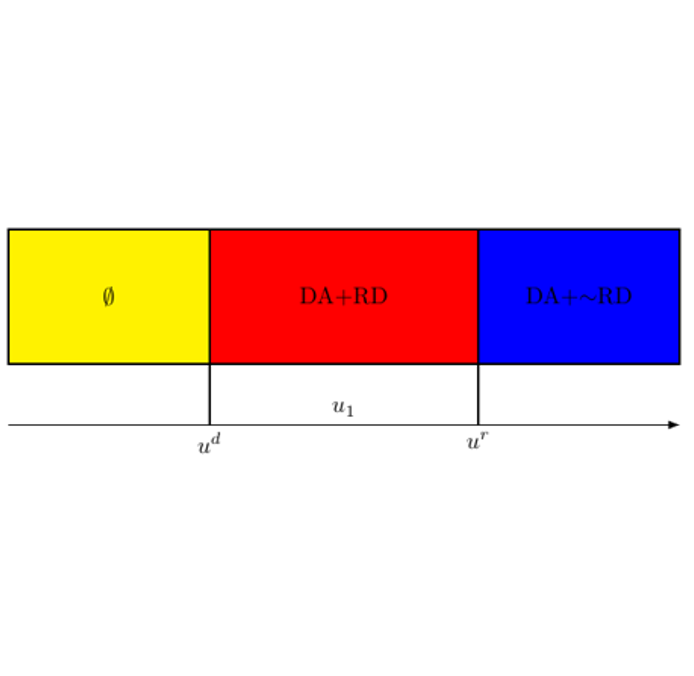}
        {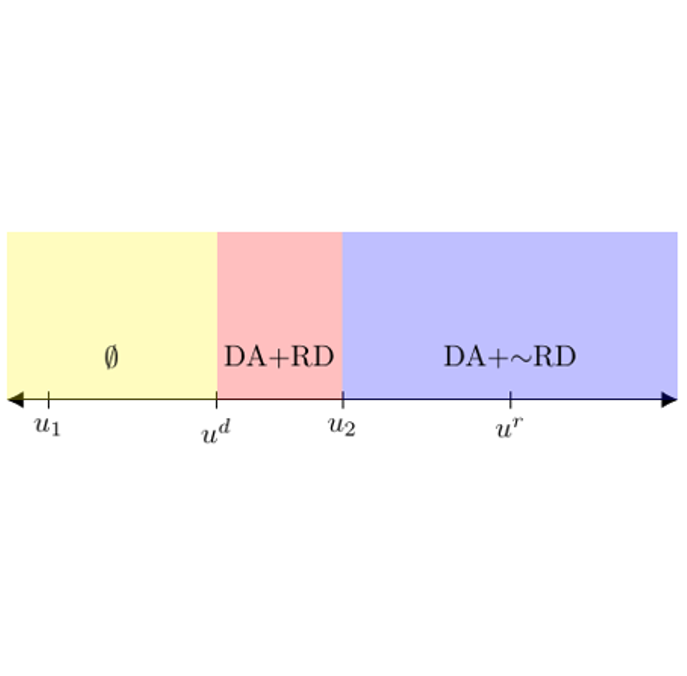}
        {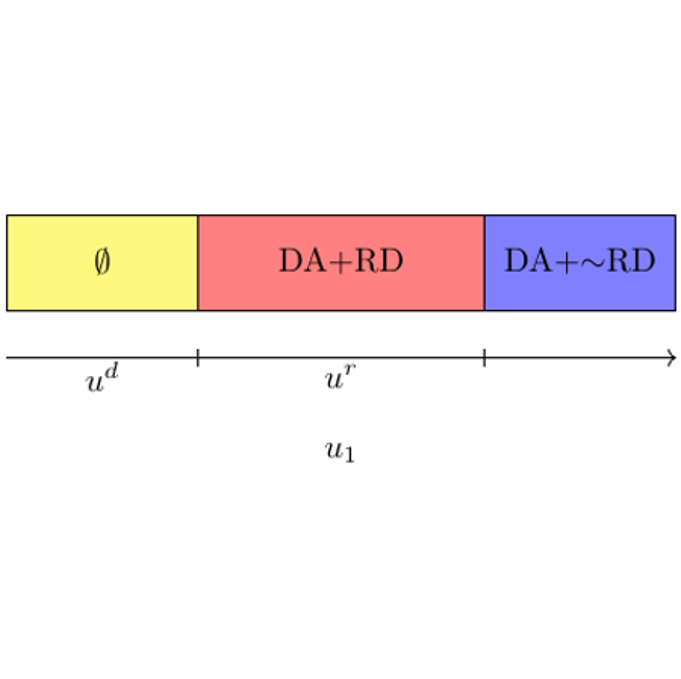}
    \midrule
    \exrowapptwothree{A Cartesian coordinate system with a horizontal red zigzag line along the x-axis and a vertical black arrow along the y-axis. The origin is marked with a black dot labeled $0$. A gray shaded circle with a dashed outline is centered at the origin, intersecting the x-axis. A red dot labeled $t$ is placed on the x-axis to the right of the origin, inside the circle. A black line extends from the origin to the red dot, forming an angle with the x-axis. This line is labeled $M^2$ near the red dot. The label $s$ is positioned in the top right corner of the image, outside the coordinate system.}
        {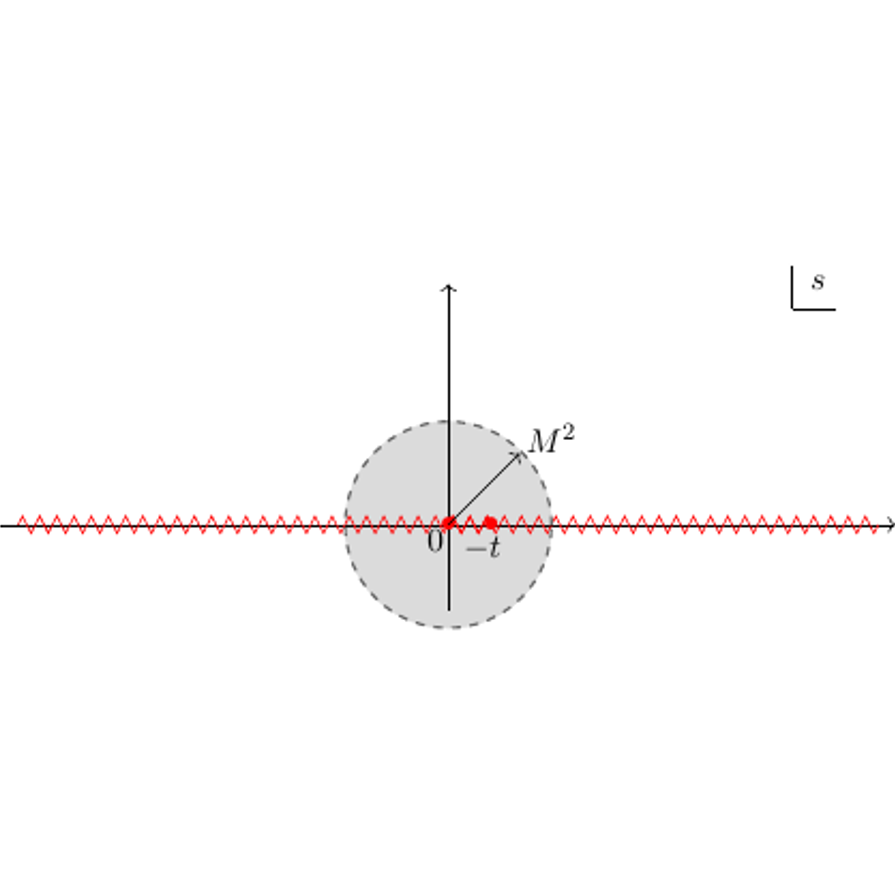}
        {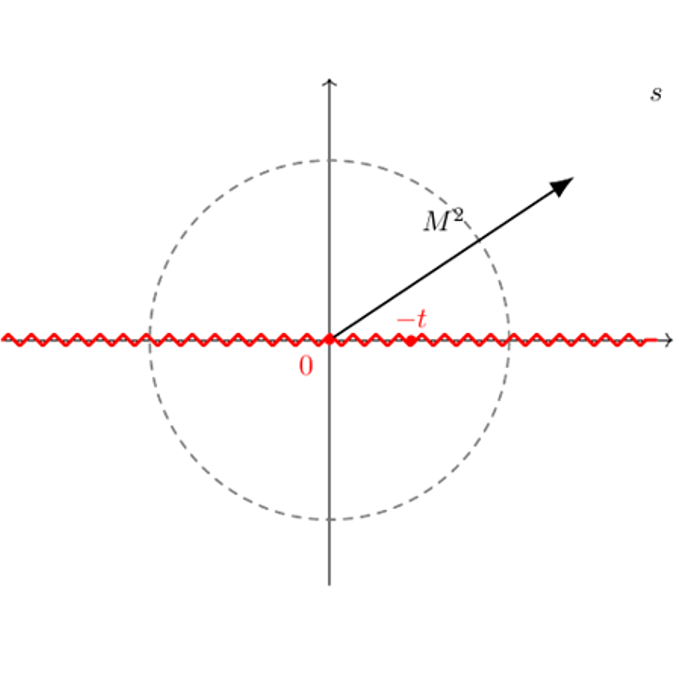}
        {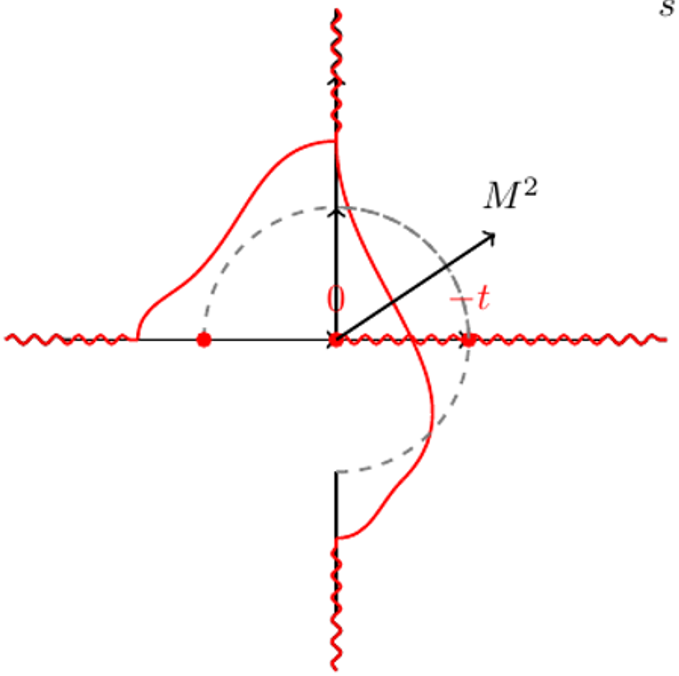}
        {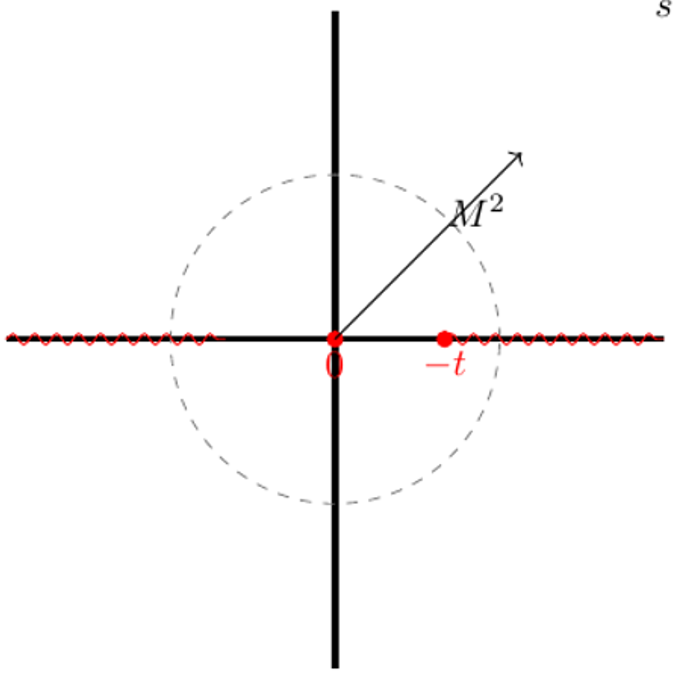}
    \midrule
    \exrowapptwosix{A line chart with the x-axis labeled \"Network size\" ranging from 0 to 140 in increments of 20 and the y-axis labeled \"Power savings \%\" ranging from 0 to 50 in increments of 10. The chart contains six lines: a solid red line with circular markers labeled \"Line (A)\" and a solid black line labeled \"Ring (A)\" both starting at the origin and curving upwards, a solid blue line with triangular markers labeled \"Star (A)\" starting at the origin and remaining mostly horizontal around 10\%, a dashed red line labeled \"Line (H)\" and a dotted black line labeled \"Ring (H)\" both following a similar upward curve to their (A) counterparts, and a dashed blue line labeled \"Star (H)\" remaining mostly horizontal around 10\%. Vertical dashed lines are drawn at x=14 labeled \"NSFNET\" and x=24 labeled \"USNET\". The legend is placed inside a white box with a black border at the bottom right corner of the chart.}
        {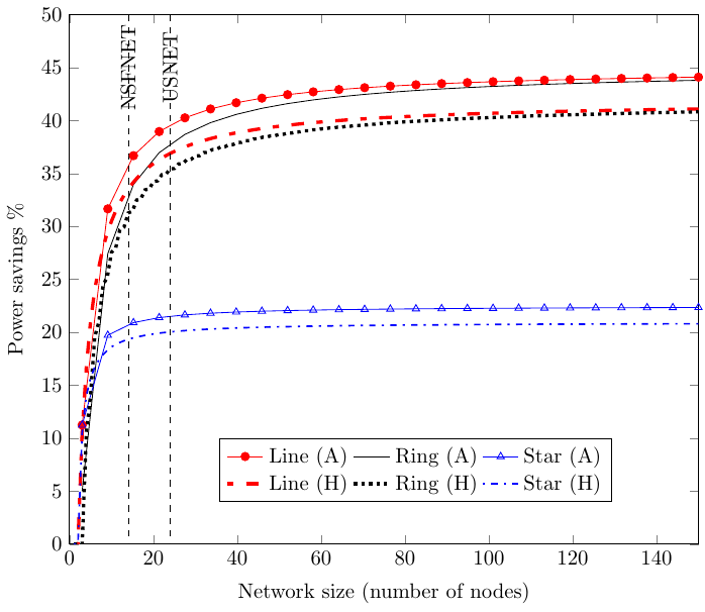}
        {structure/figures/not_compiled.png}
        {structure/figures/not_compiled.png}
        {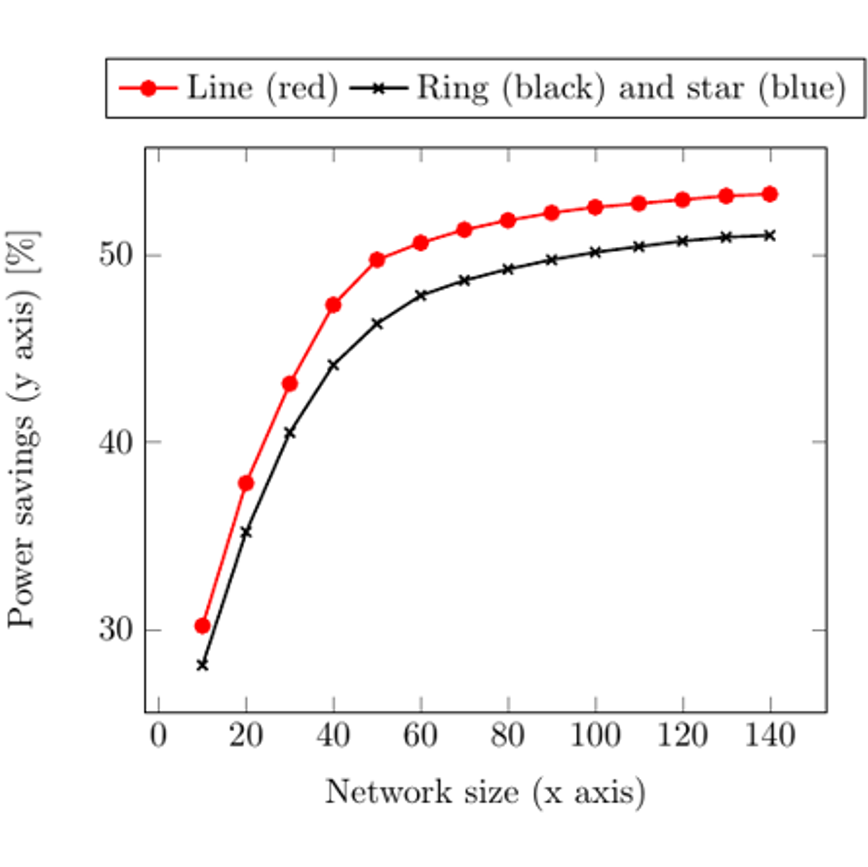}
    \midrule
    \exrowapptwofour{An automaton with four states labeled as 0, 1, 2, and 3. The initial state is 0, as indicated by the incoming arrow labeled 'start'. From state 0, the automaton transition to state 1 on input a, to state 2 on input b or f and to state 3 on input g. State 1 has a self-loop on input c, state 2 loops back to itself on inputs c or d and state 3 has a self loop on input d. States 1, 2 and 3 are indicated by double circles around it.}
        {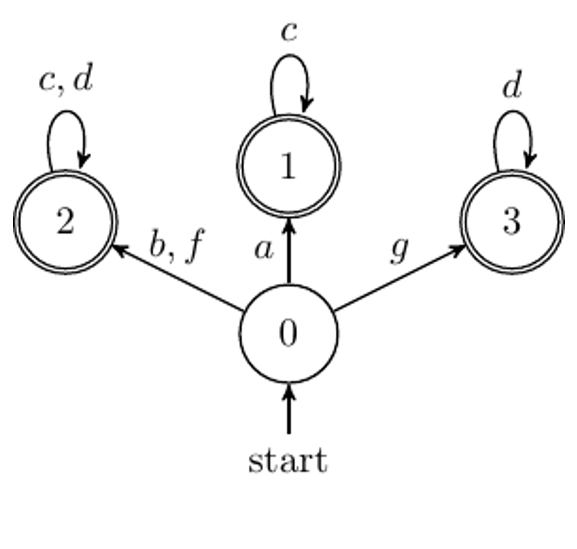}
        {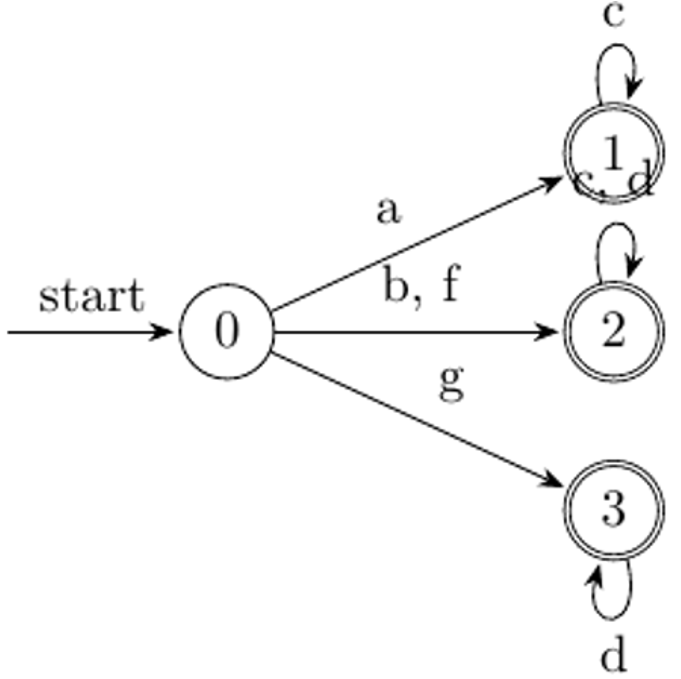}
        {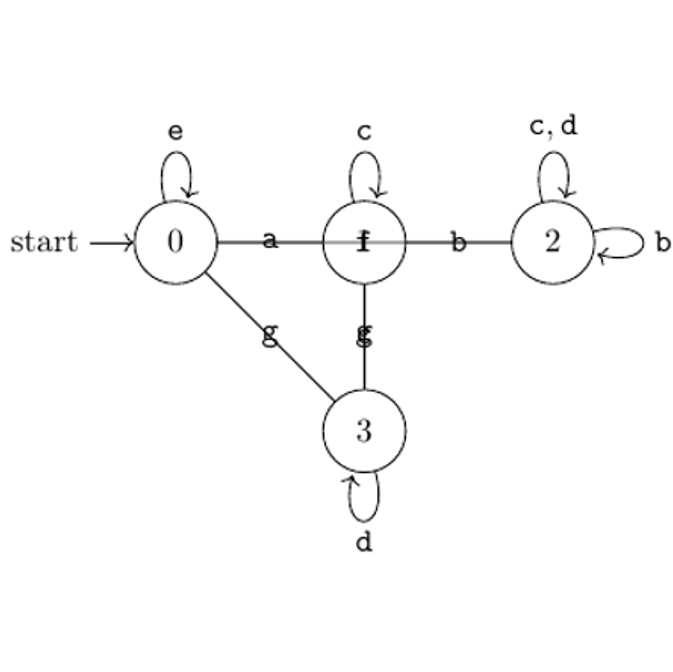}
        {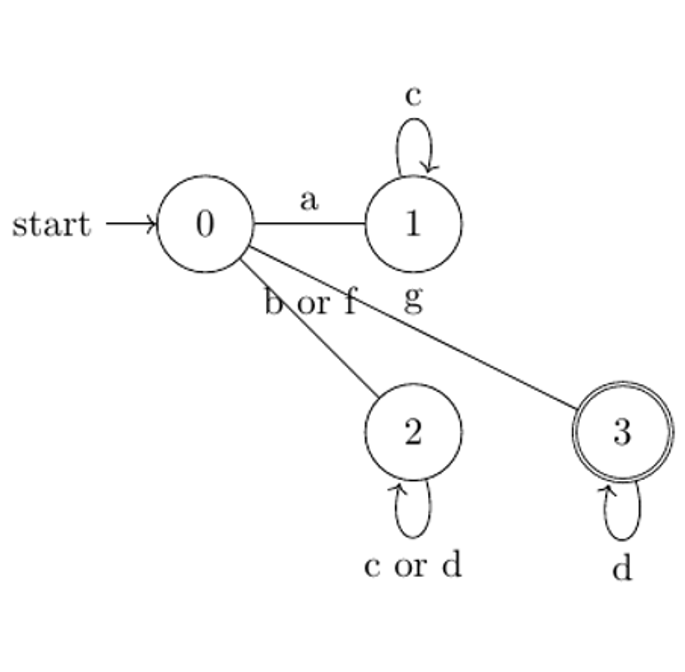}
    \midrule
    \exrowapptwoseven{A zigzag pattern composed of alternating red and blue lines connects a series of black dots and red squares vertically. The pattern starts at the top with a black dot connected to a red square by a blue line, followed by a red line connecting the red square to the next black dot. This alternating pattern continues downwards, with each black dot connected to a red square by a blue line and each red square connected to the next black dot by a red line. The sequence ends at the bottom with a black dot labeled $w$. Two dashed horizontal lines are placed above the topmost black dot and below the bottommost black dot, with another dashed line in the middle. Curly braces on the right side of the pattern span the sections between the dashed lines, with the top brace labeled $(m_{2j+1} - m_{2j})$ and the bottom brace labeled $(m_{2j+2} - m_{2j+1})$.}
        {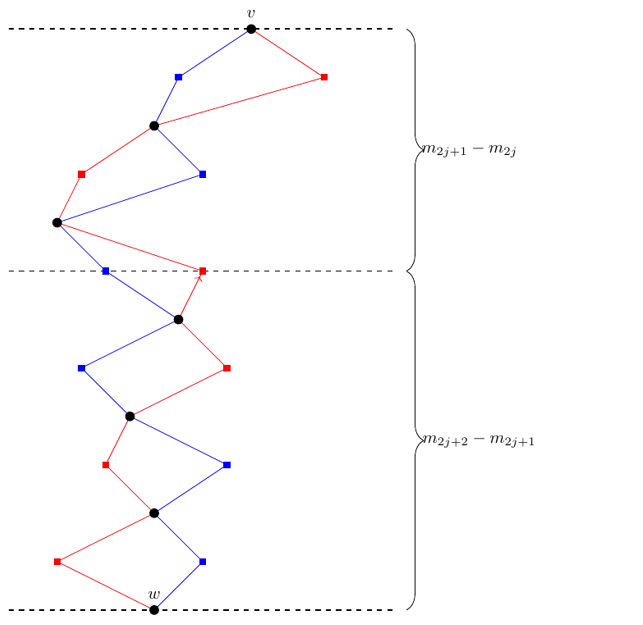}
        {structure/figures/not_compiled.png}
        {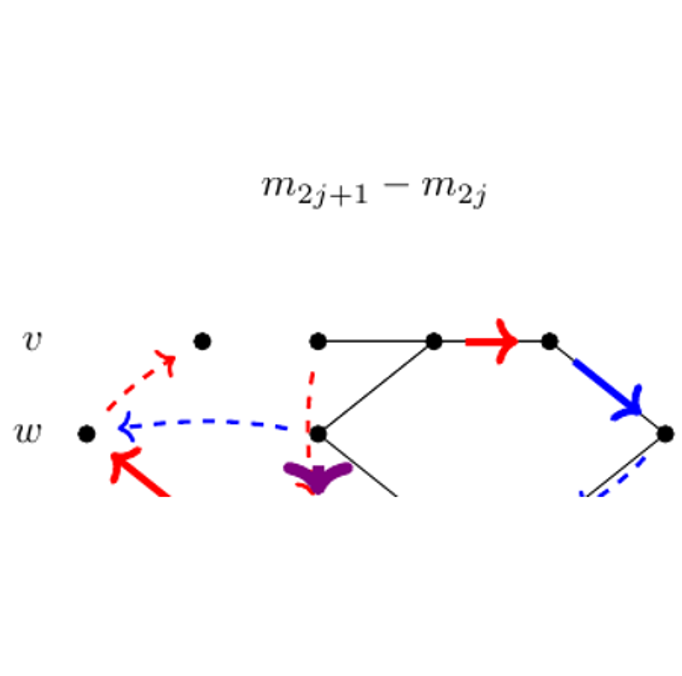}
        {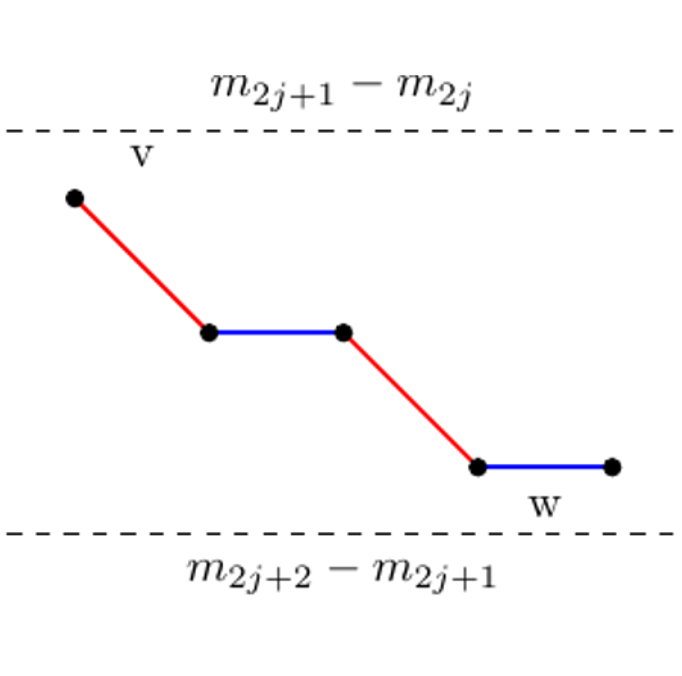}
    \bottomrule
\end{tabularx}
\end{table}

\setlength{\aboverulesep}{0pt}
\setlength{\belowrulesep}{0pt}
\setlength{\tabcolsep}{4pt}
\renewcommand{\arraystretch}{1.03}
\setlength{\abovetopsep}{0pt}

\newcommand{\exrowappthreeone}[5]{%
  \desc{#1} & \imgcell{#2} & \imgcellokay{#3} & \imgcellbad{#4} & \imgcellgood{#5} \\
}
\newcommand{\exrowappthreetwo}[5]{%
  \desc{#1} & \imgcell{#2} & \imgcell{#3} & \imgcell{#4} & \imgcell{#5} \\
}
\newcommand{\exrowappthreethree}[5]{%
  \desc{#1} & \imgcell{#2} & \imgcell{#3} & \imgcellokay{#4} & \imgcellverygood{#5} \\
}
\newcommand{\exrowappthreefour}[5]{%
  \desc{#1} & \imgcell{#2} & \imgcellbad{#3} & \imgcellbad{#4} & \imgcellbad{#5} \\
}
\newcommand{\exrowappthreefive}[5]{%
  \desc{#1} & \imgcell{#2} & \imgcellbad{#3} & \imgcellokay{#4} & \imgcellgood{#5} \\
}
\newcommand{\exrowappthreesix}[5]{%
  \desc{#1} & \imgcell{#2} & \imgcellokay{#3} & \imgcellokay{#4} & \imgcellgood{#5} \\
}
\newcommand{\exrowappthreeseven}[5]{%
  \desc{#1} & \imgcell{#2} & \imgcellbad{#3} & \imgcell{#4} & \imgcellgood{#5} \\
}

\begin{table}[t]
\centering
\caption{Exemplary scientific TikZ figures produced by one baseline LLM (GPT-4o) and two of our finetuned LLMs (TikZilla-8B and TikZilla-8B-RL) using the prompts from the first column which have been VLM augmented based on the Ground Truth figures in the second column. \legendbox{green}-boxed figures have been rated as very good, \legendbox{yellow} as good, \legendbox{orange} as bad, and \legendbox{red} as very bad by human annotators. Empty cells indicate non-compilable TikZ code.}
\label{tab:examples_3}
\begin{tabularx}{\textwidth}{YCCCC}
    \toprule
    \headercell{Prompt} & \headercell{Ground Truth} & \headercell{GPT-4o} & \headercell{TikZilla-8B} & \headercell{TikZilla-8B-RL} \\
    \midrule

    \exrowappthreeone{The image is a line chart with the x-axis labeled "simulation time $\tau$" ranging from 0 to 400, marked at intervals of 100. The y-axis is labeled $|b_{E_0}|$ and ranges from 0 to 8, with a scale factor of $10^{-2}$ indicated at the top left. The chart contains two lines and a horizontal reference line. The first line is blue with circular markers, representing $\langle E_0 \rangle, p = 0.0001$, and it fluctuates between 0 and approximately 2. The second line is red with triangular markers, representing $\langle E_0 \rangle, p = 0.001$, and it also fluctuates between 0 and approximately 2, with more pronounced peaks. A green horizontal line is drawn at $y = 8$, representing the value $8 \cdot 10^{-2}$. The legend is located inside the chart at the top right, containing three entries: a blue line with circular markers labeled $\langle E_0 \rangle, p = 0.0001$, a red line with triangular markers labeled $\langle E_0 \rangle, p = 0.001$, and a green line labeled $8 \cdot 10^{-2}$.}
        {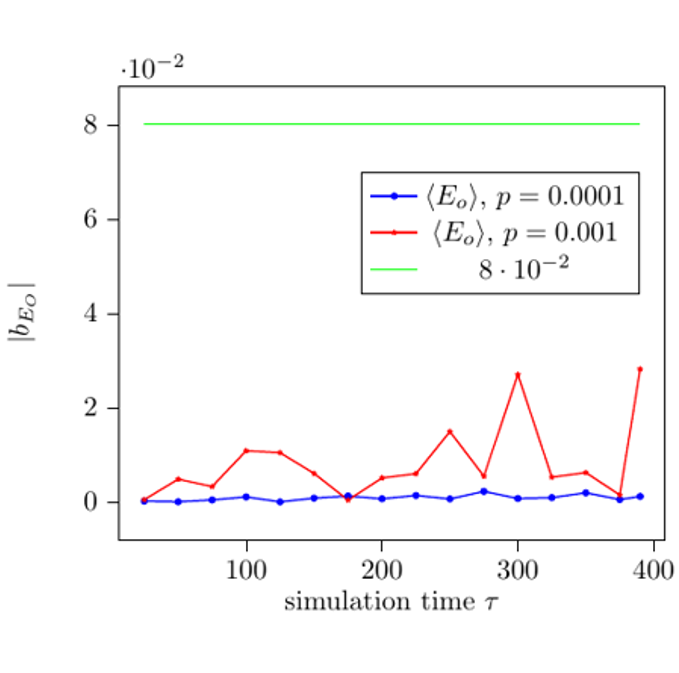}
        {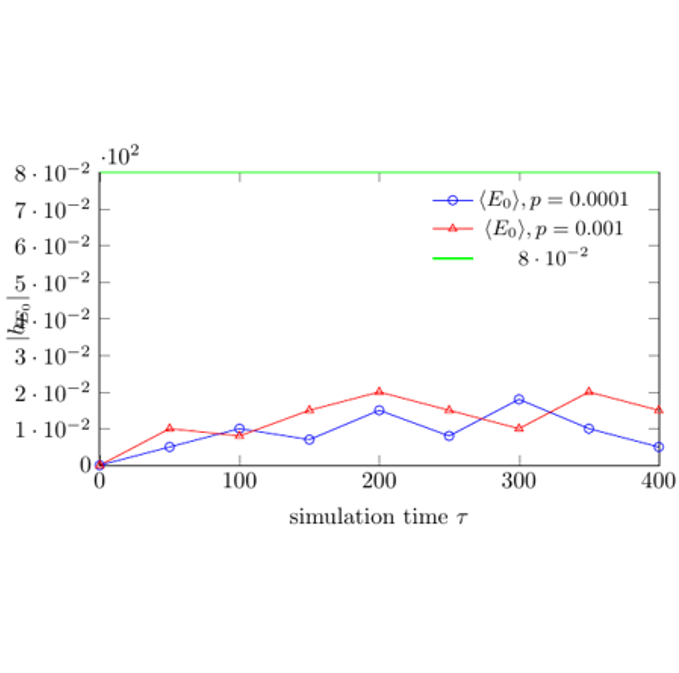}
        {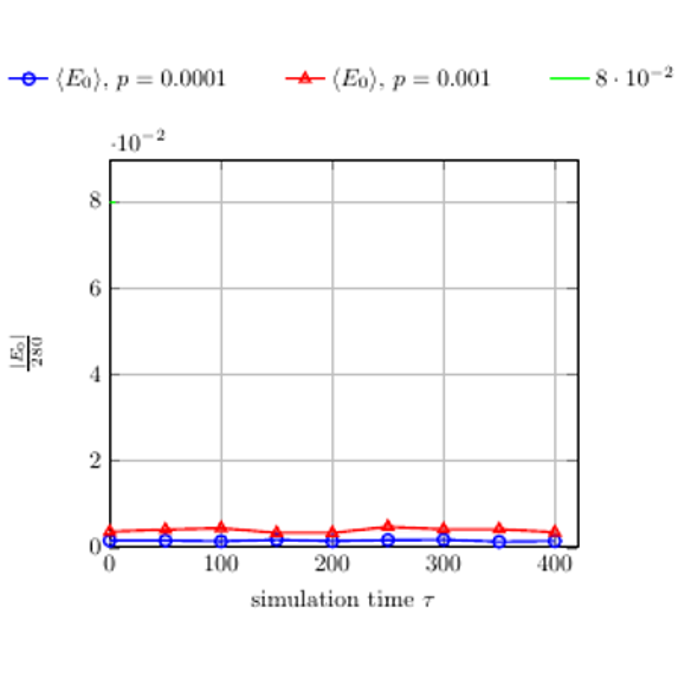}
        {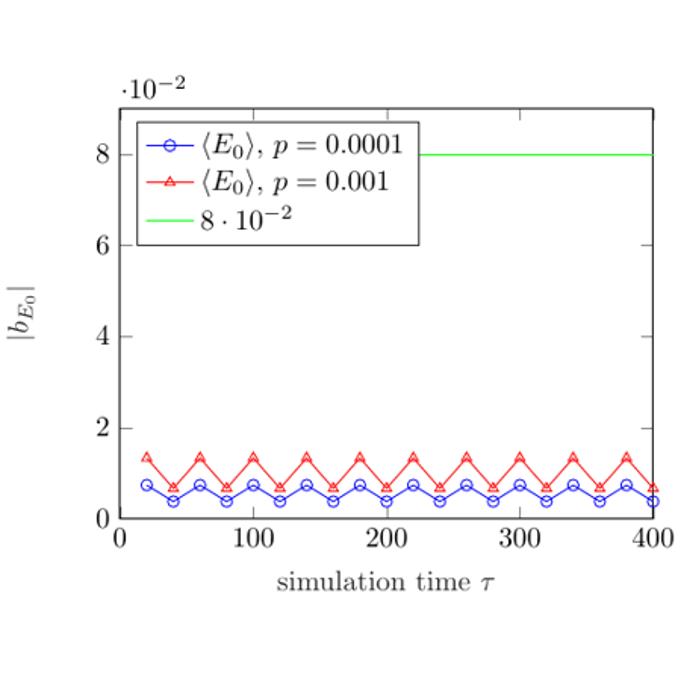}
    \midrule
    \exrowappthreetwo{A block diagram for a verification workflow. It takes two inputs, labeled 'Spec' and 'Safe', which enter the system from the top-left. Inside the main box, the process begins with a purple block labeled 'Invariant Generator', which receives the Spec input and produces$I_{new}$. This output is stored in a cylinder labeled 'Invs'. From Invs, a set of invariants I $\subseteq$ Invs is passed to the next component, the 'CTI Eliminator' shown as a blue rectangle. Directly below is another blue rectangle labeled 'CTI Generator', which also receives the Spec input and outputs 'CTIs' for the CTI Eliminator. Both blue rectangles are inside a big gray rectangle. On the right side of the diagram is a white rectangle labeled $\text{Ind} \triangleq \bigwedge \text{Safe} \wedge A_1 \wedge \cdots \wedge A_k \wedge A_{k+1}$. It receives two inputs: Safe and $A_{k+1}$, the latter coming from the CTI Eliminator. An arrow points to the CTI Generator and another arrow exits this block to the right, labeled 'Output'.}
        {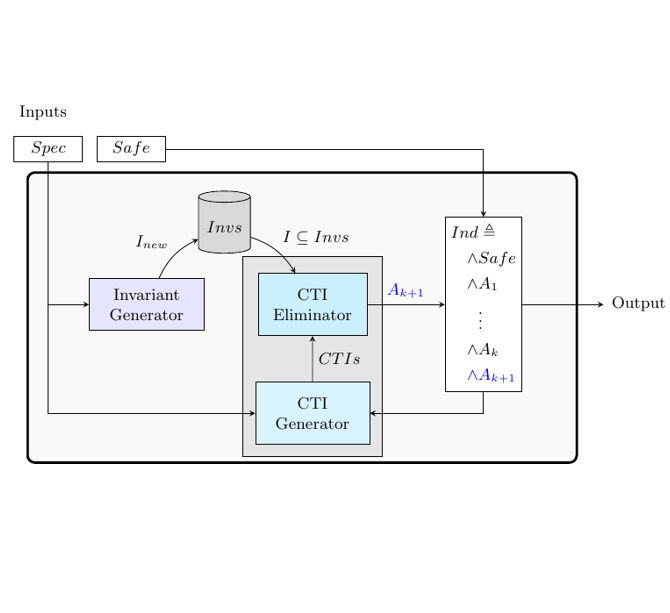}
        {structure/figures/not_compiled.png}
        {structure/figures/not_compiled.png}
        {structure/figures/not_compiled.png}
    \midrule
    \exrowappthreethree{A workflow for cross-validation using k-folds. It consists of four circular stages that are connected by arrows and the entire process is repeated k times. The first circle is labeled '1/k training set' and annotated beneath as 'k-folding (k=10)'. An arrow leads to the second circle, which is labeled 'up to 10 training instances' and annotated beneath as 'Training '(local search)'. The process continues to a third circle labeled '(k-1)/k training set' and annotated 'Validation (subsetting)' beneath. From there, a final arrow lead to a circle labeled '100\% test set' and annotated 'Test' beneath. A horizontal bracket across the top first three circles notes that its 'repeated k times using k folds'.}
        {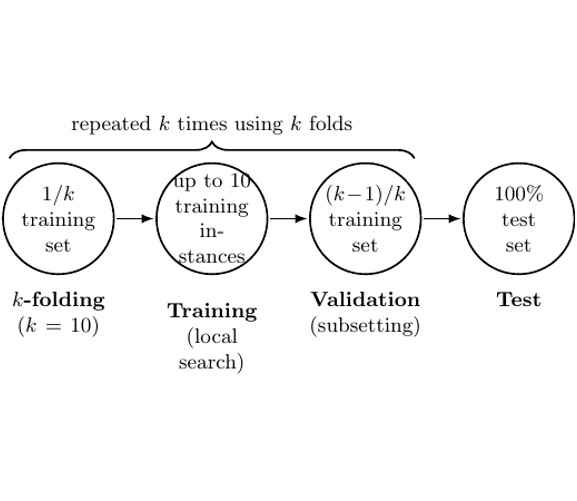}
        {structure/figures/not_compiled.png}
        {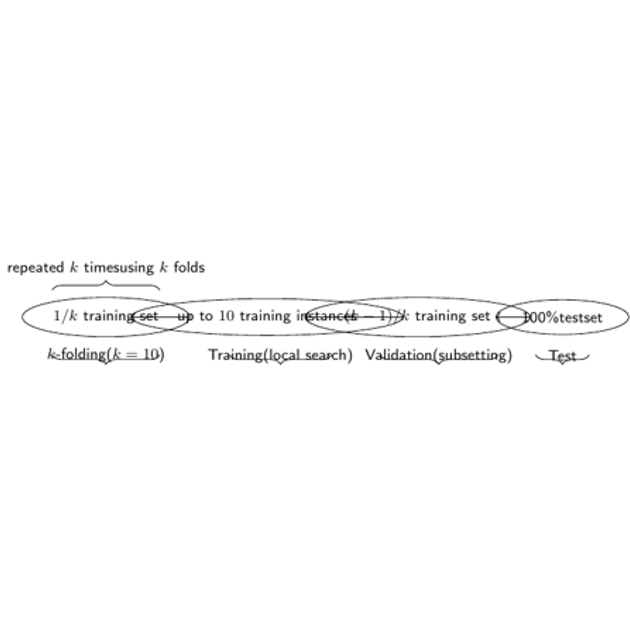}
        {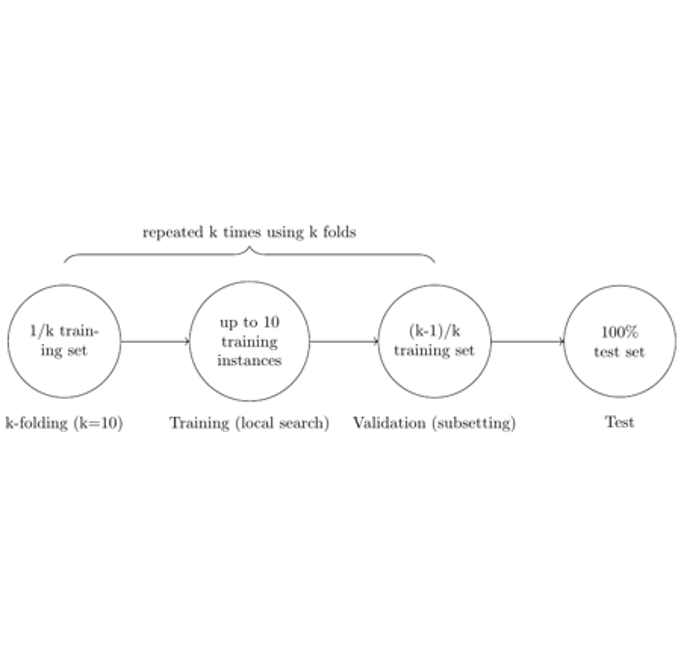}
    \midrule
    \exrowappthreefour{A flowchart divided into three vertical sections, each outlined with a blue dashed border. These sections are labeled: 'Unlimited DG Loop' on the left, 'Projection and Reconstruction' in the center and 'FD Loop' on the right. Each section contains a sequence of boxes connected by arrows that indicate the computational flow. In the Unlimited DG Loop, the flow begins at the top with a rounded rectangle labeled 'Send ghost cells and fluxes'. A downward arrow leads to a rectangle labeled 'Compute $u_i^{,n+1}$. Another downward arrow connects to a block labeled '$TCI(u_i^{,n+1})$'. This splits into two branches: an arrow labeled 'Passed' in green continues downward to a rectangle that is labeled $u_i^{n+1} = u_i^{*,n+1}$, while an arrow labeled 'Failed' in red exits to the right, connecting to the middle section. In the Projection and Reconstruction section, the incoming red arrow leads to a rectangle labeled $\mathcal{P}(u_i^n),\ \mathcal{P}(F_i^{i,n}),\ \mathcal{P}(S_i^n)$. A arrow to the right connects it to the FD Loop section with a rectangle labeled 'Compute'. This rectangle connects to another rectangle labeled 'TCI', which then splits into two branches labeled 'Passed' in green and 'Failed' in red.}
        {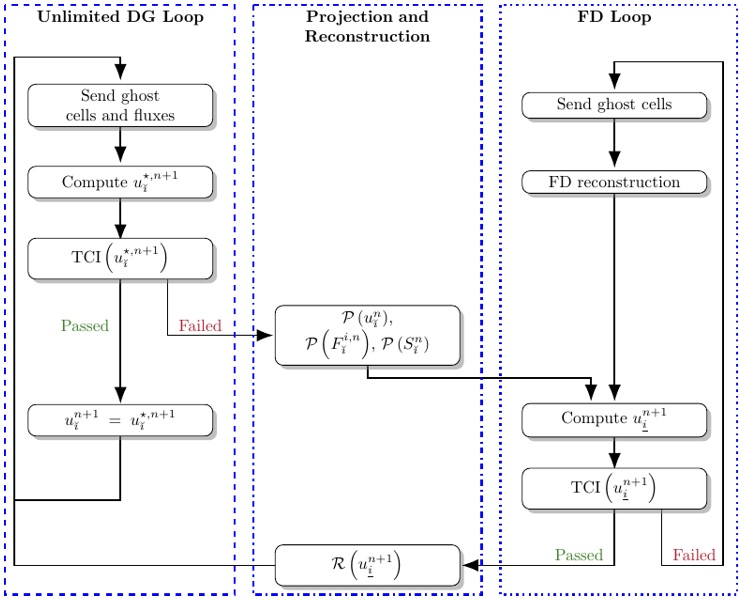}
        {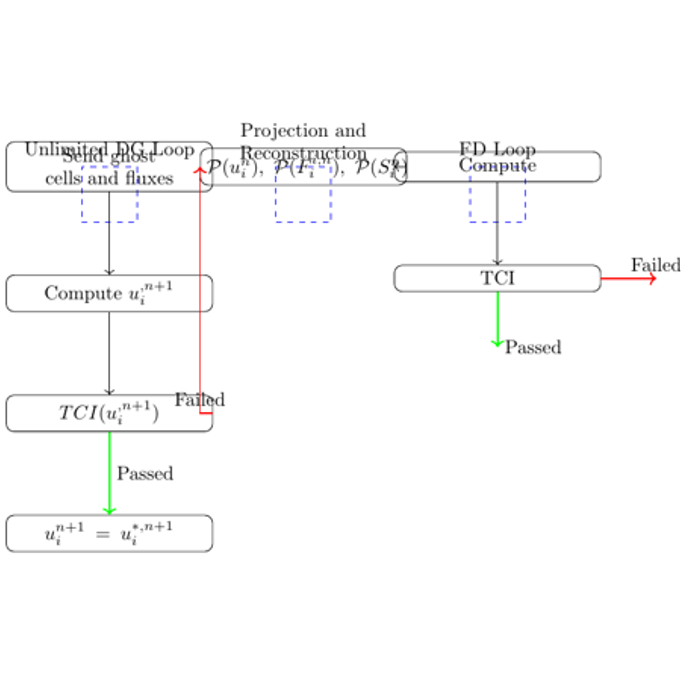}
        {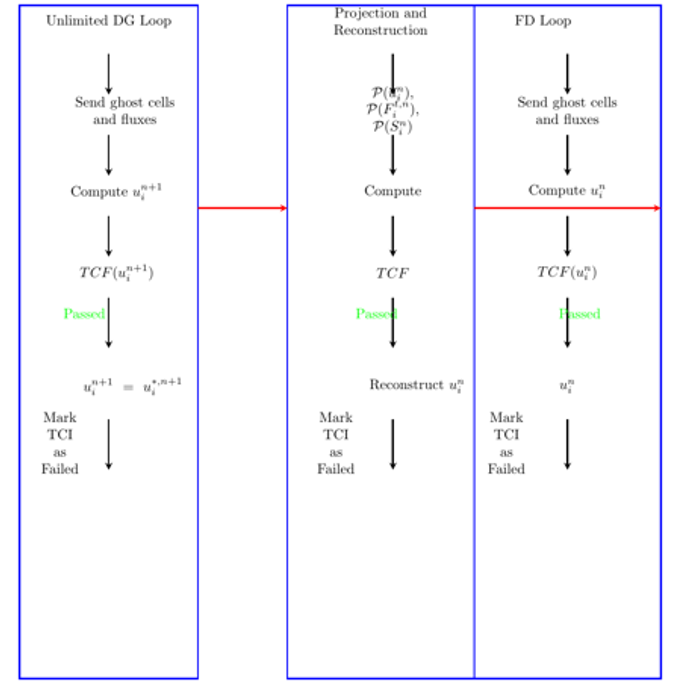}
        {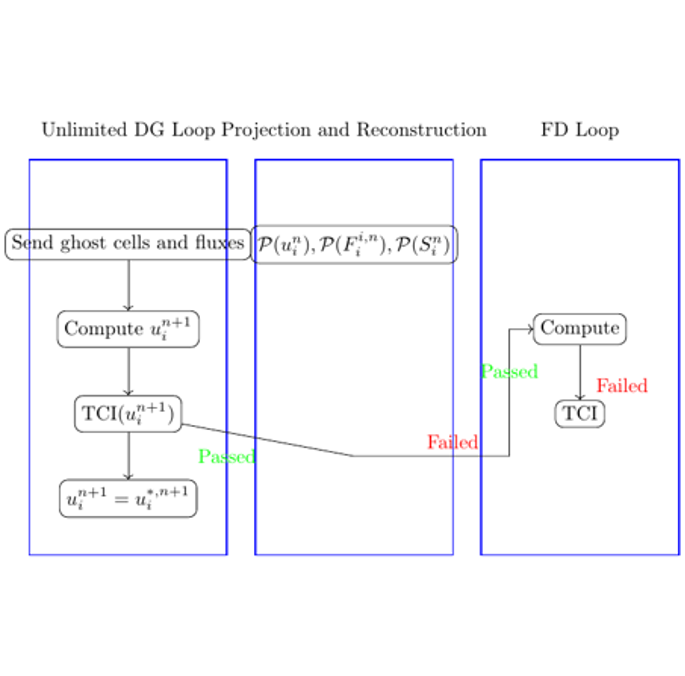}
    \midrule
    \exrowappthreefive{A circular pie chart is divided into eight colored segments with distinct labels and percentages. Starting from the top and moving clockwise, the first segment is labeled "Breast" with a percentage of 33.5\% and is colored blue. The second segment is labeled "Other" with a percentage of 12.1\% and is colored purple. The third segment is labeled "Lymph Nodes" with a percentage of 4.7\% and is colored gray. The fourth segment is labeled "Brain" with a percentage of 5.3\% and is colored light blue. The fifth segment is labeled "Kidney" with a percentage of 4.7\% and is colored green. The sixth segment is labeled "Liver" with a percentage of 6.6\% and is colored red. The seventh segment is labeled "Prostate" with a percentage of 9.4\% and is colored orange. The eighth segment is labeled "Lung" with a percentage of 6.5\% and is colored yellow. The ninth segment is labeled "Colorectal" with a percentage of 17.2\% and is colored cyan. Each label is placed outside the corresponding segment, with lines connecting the labels to the segments.}
        {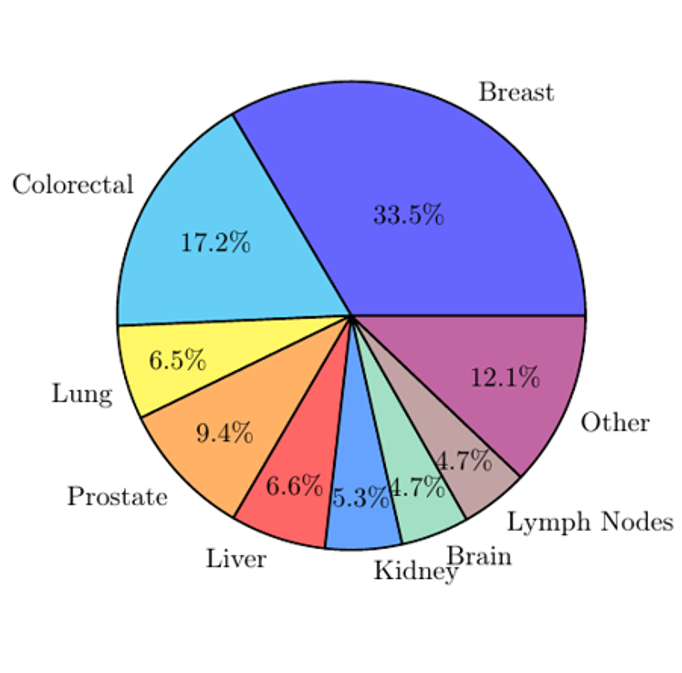}
        {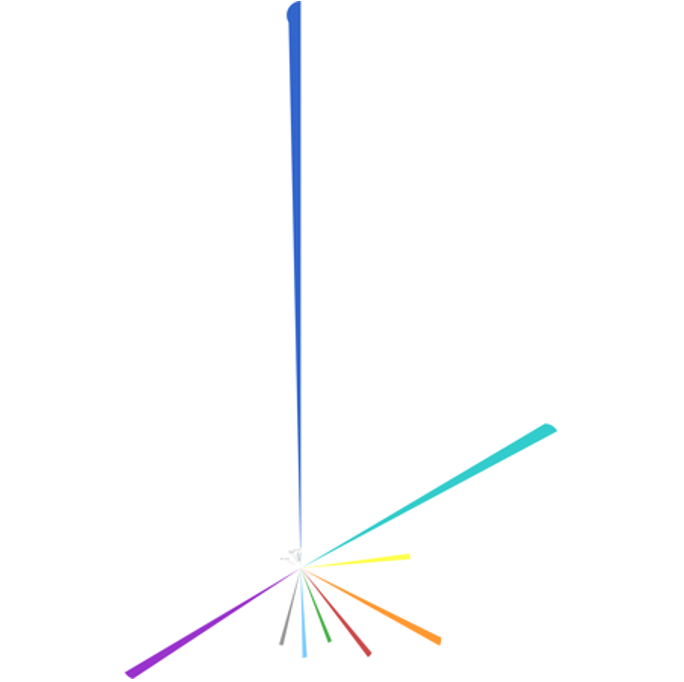}
        {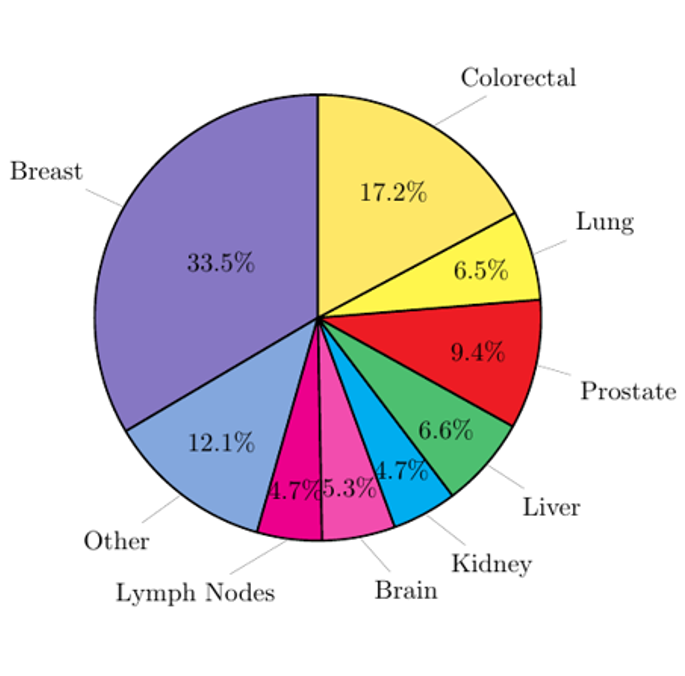}
        {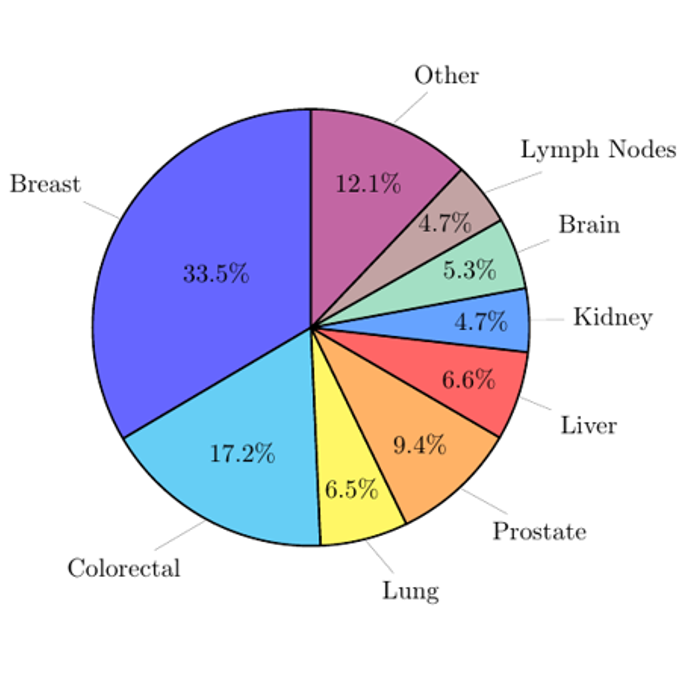}
    \midrule
    \exrowappthreesix{Two black rectangles are positioned horizontally in the center of the image. The left rectangle contains the label $\phi n$ in white, while the right rectangle contains the label Nor in white. Above the left rectangle, there is a curved arrow pointing downwards, labeled $\{ID, SR^{-1}\}$. Above the right rectangle, there is a similar curved arrow pointing downwards, labeled $\{ID, X, RZ^{-1}, CU, Rev, Lshift, Rshift\}$. A curved arrow connects the left rectangle to the right rectangle, labeled $\{QFT\ n\}$, and another curved arrow connects the right rectangle back to the left rectangle, labeled $\{QFT^{-1}\ n\}$.}
        {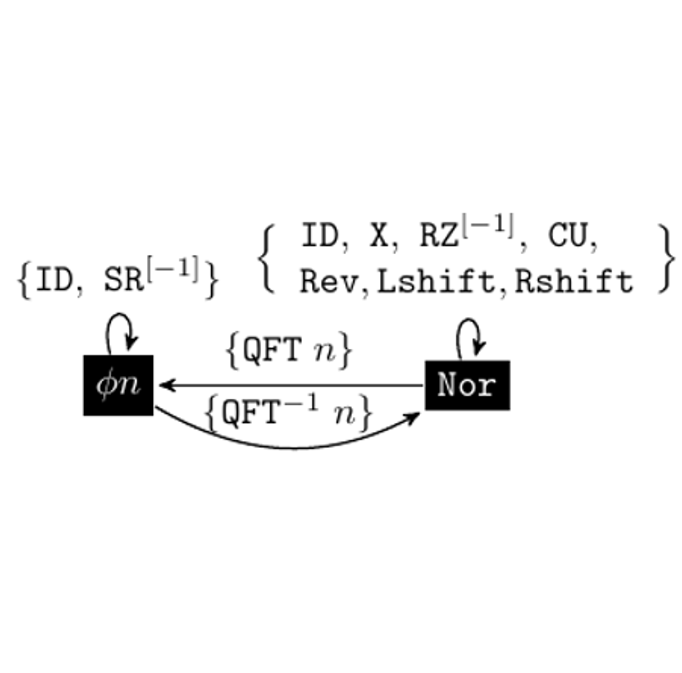}
        {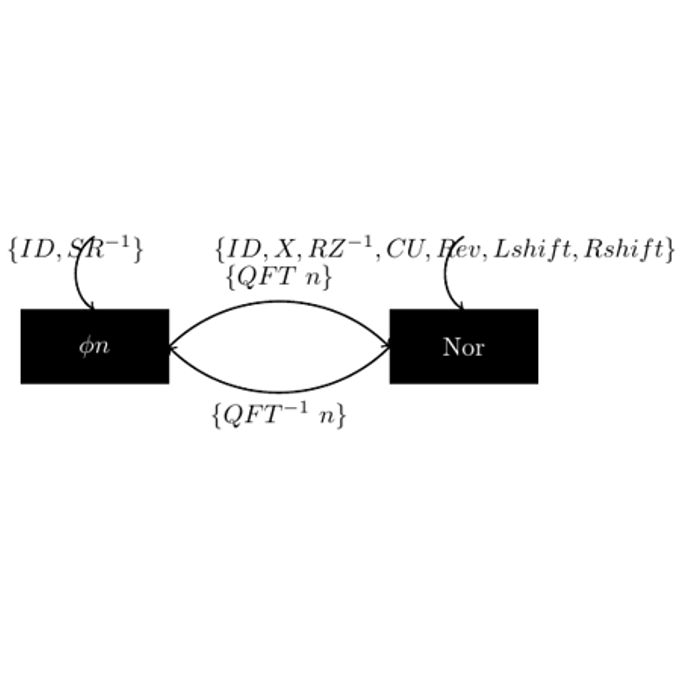}
        {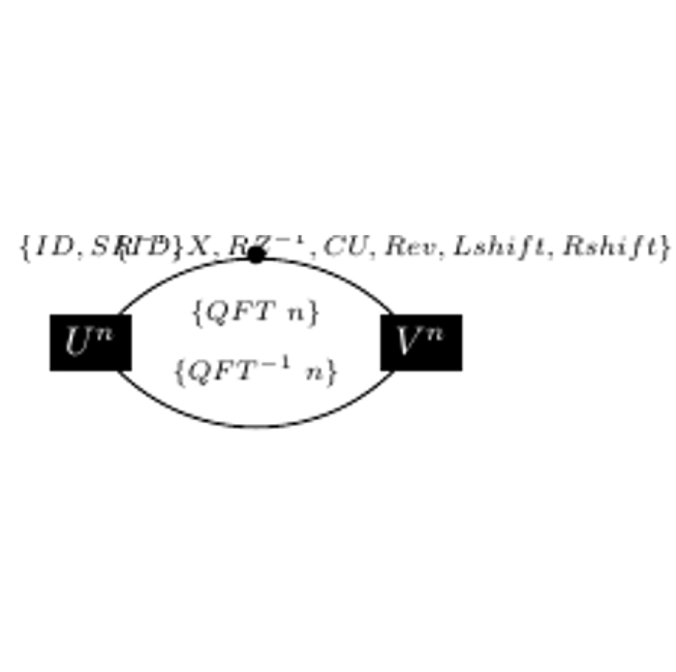}
        {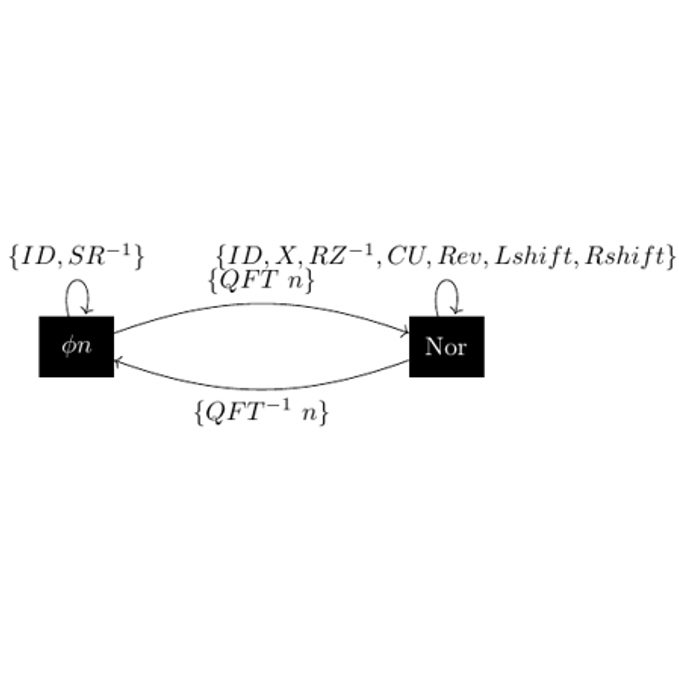}
    \midrule
    \exrowappthreeseven{A large black curved shape occupies the top left, resembling a section of a circle, with a blue parallelogram labeled "tangent space" inside it. The parallelogram is oriented diagonally, with a red dashed arrow labeled "v" pointing from the bottom left to the top right, ending at a point labeled "x". Above the parallelogram, the red text "{D(z_i)}" is positioned. Two black arrows labeled "D" and "E" point downward from the curved shape to a smaller coordinate system at the bottom right. This coordinate system has two black axes, with the vertical axis labeled "\mathbb{R}^k" and the horizontal axis extending to the right. A red dashed horizontal line labeled "{z_i}" extends from a black dot labeled "z" on the horizontal axis. The entire diagram is annotated with "\mathcal{D} \in \mathbb{R}^d" in black text near the top right of the curved shape.}
        {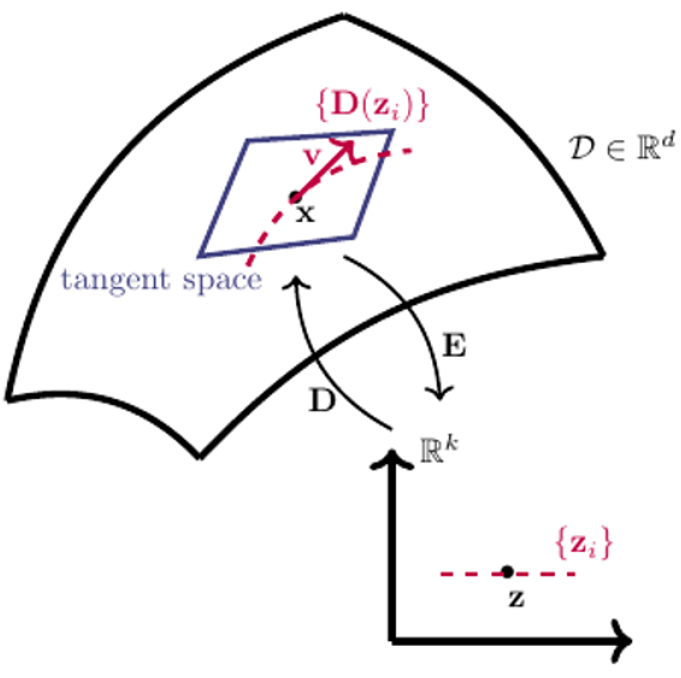}
        {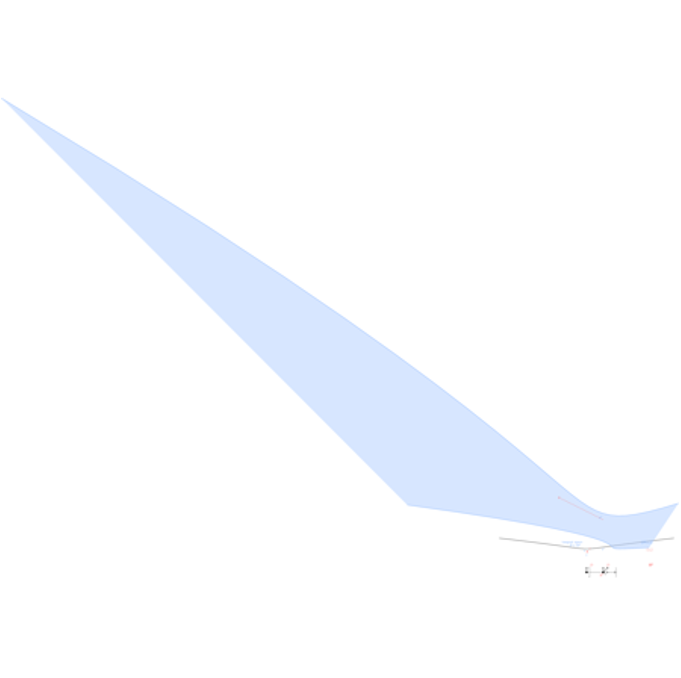}
        {structure/figures/not_compiled.png}
        {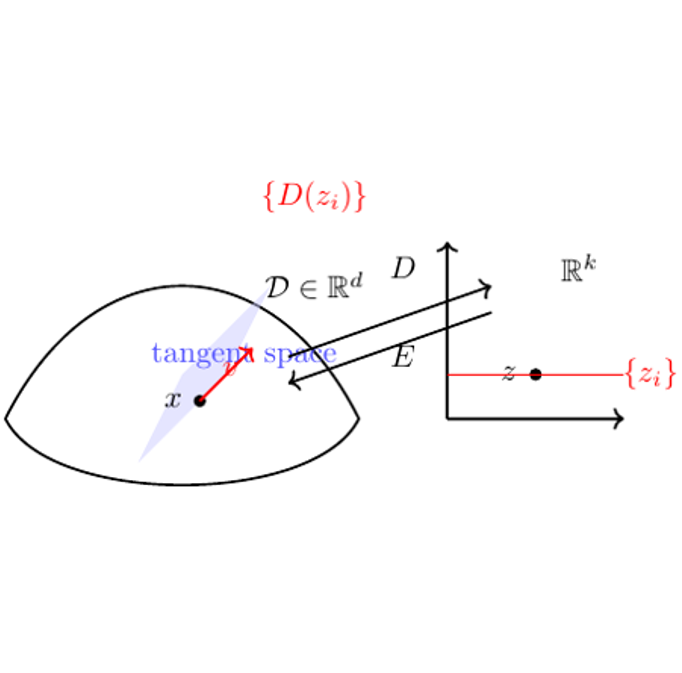}
    \bottomrule
\end{tabularx}
\end{table}

\setlength{\aboverulesep}{0pt}
\setlength{\belowrulesep}{0pt}
\setlength{\tabcolsep}{4pt}
\renewcommand{\arraystretch}{1.03}
\setlength{\abovetopsep}{0pt}

\newcommand{\exrowappfourone}[5]{%
  \desc{#1} & \imgcell{#2} & \imgcell{#3} & \imgcellgood{#4} & \imgcellokay{#5} \\
}
\newcommand{\exrowappfourtwo}[5]{%
  \desc{#1} & \imgcell{#2} & \imgcellverygood{#3} & \imgcellokay{#4} & \imgcellverygood{#5} \\
}
\newcommand{\exrowappfourthree}[5]{%
  \desc{#1} & \imgcell{#2} & \imgcellbad{#3} & \imgcellbad{#4} & \imgcellokay{#5} \\
}
\newcommand{\exrowappfourfour}[5]{%
  \desc{#1} & \imgcell{#2} & \imgcellokay{#3} & \imgcellbad{#4} & \imgcellgood{#5} \\
}
\newcommand{\exrowappfourfive}[5]{%
  \desc{#1} & \imgcell{#2} & \imgcell{#3} & \imgcellverygood{#4} & \imgcellgood{#5} \\
}
\newcommand{\exrowappfoursix}[5]{%
  \desc{#1} & \imgcell{#2} & \imgcellokay{#3} & \imgcellbad{#4} & \imgcellgood{#5} \\
}
\newcommand{\exrowappfourseven}[5]{%
  \desc{#1} & \imgcell{#2} & \imgcell{#3} & \imgcell{#4} & \imgcellgood{#5} \\
}

\begin{table}[t]
\centering
\caption{Exemplary scientific TikZ figures produced by one baseline LLM (GPT-4o) and two of our finetuned LLMs (TikZilla-8B, and TikZilla-8B-RL) using the prompts from the first column which have been VLM augmented based on the Ground Truth figures in the second column. \legendbox{green}-boxed figures have been rated as very good, \legendbox{yellow} as good, \legendbox{orange} as bad, and \legendbox{red} as very bad by human annotators. Empty cells indicate non-compilable TikZ code.}
\label{tab:examples_4}
\begin{tabularx}{\textwidth}{YCCCC}
    \toprule
    \headercell{Prompt} & \headercell{Ground Truth} & \headercell{GPT-4o} & \headercell{TikZilla-8B} & \headercell{TikZilla-8B-RL} \\
    \midrule

    \exrowappfourone{A block diagram features two light blue rectangular blocks labeled $D_0(s)$ and $N_0(s)$, positioned vertically with $D_0(s)$ on top and $N_0(s)$ below. The block $D_0(s)$ is labeled "device dynamics" beneath it, while $N_0(s)$ is labeled "network dynamics" below. To the left of $D_0(s)$, a vertical bracket labeled $[\Delta p_d \, \Delta q_d]$ points to a black dot, which connects horizontally to $D_0(s)$ with a line labeled $[-]$. Above this line, another vertical bracket labeled $[\Delta p \, \Delta q]$ points downward. To the right of $D_0(s)$, a horizontal arrow labeled $[\Delta \omega \, \Delta |v|]$ points rightward. Below $D_0(s)$, a horizontal line connects to a black dot to the left of $N_0(s)$. From this dot, a vertical bracket labeled $[\Delta p_e \, \Delta q_e]$ points leftward. The block $N_0(s)$ has a horizontal arrow extending from its right side, labeled $[\Delta \omega_d \, \Delta |v|_d]$, pointing rightward.}
        {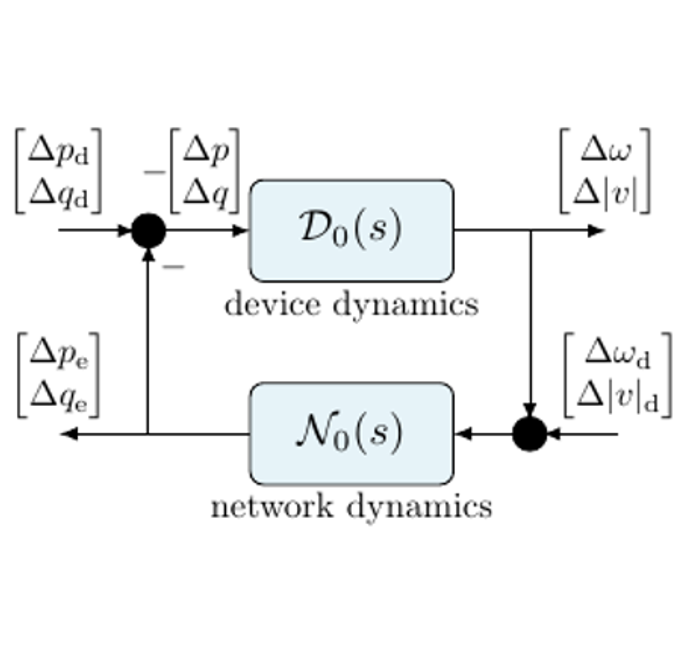}
        {structure/figures/not_compiled.png}
        {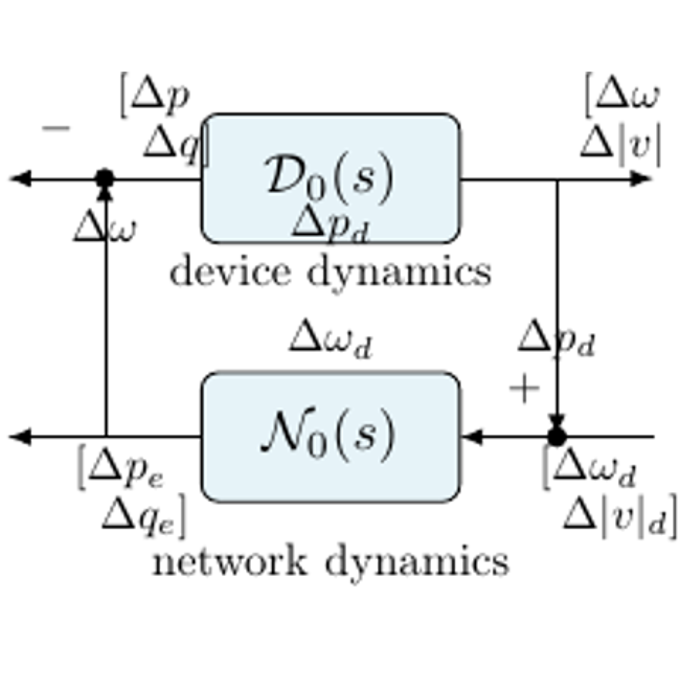}
        {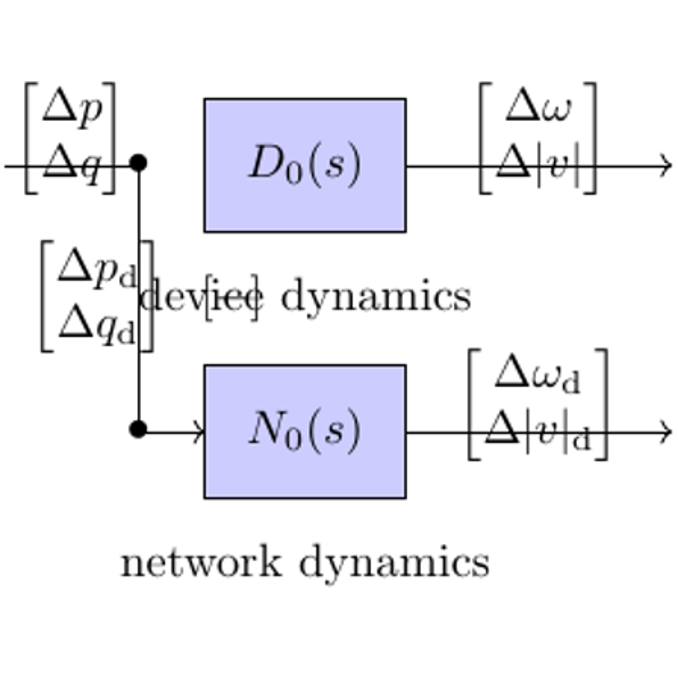}
    \midrule
    \exrowappfourtwo{The bar chart contains two groups of vertical bars labeled "GSM8K" and "MATH" on the x-axis, with the y-axis labeled "Performance Change" ranging from -10 to 10 in increments of 2. Each group contains three bars. In the "GSM8K" group, the first bar is blue, extending from 0 to 10, labeled "10.0" at the top. The second bar is green, extending from 0 to -1.8, labeled "-1.8" at the bottom. The third bar is red, extending from 0 to -6.0, labeled "-6.0" at the bottom. In the "MATH" group, the first bar is blue, extending from 0 to 9.6, labeled "9.6" at the top. The second bar is green, extending from 0 to 2.4, labeled "2.4" at the top. The third bar is red, extending from 0 to -1.2, labeled "-1.2" at the bottom. A dashed horizontal line crosses the y-axis at 0. Below the chart, a legend identifies the colors: blue for "Llama2-7B," green for "Llama3.1-8B," and red for "Llama-R1-8B."}
        {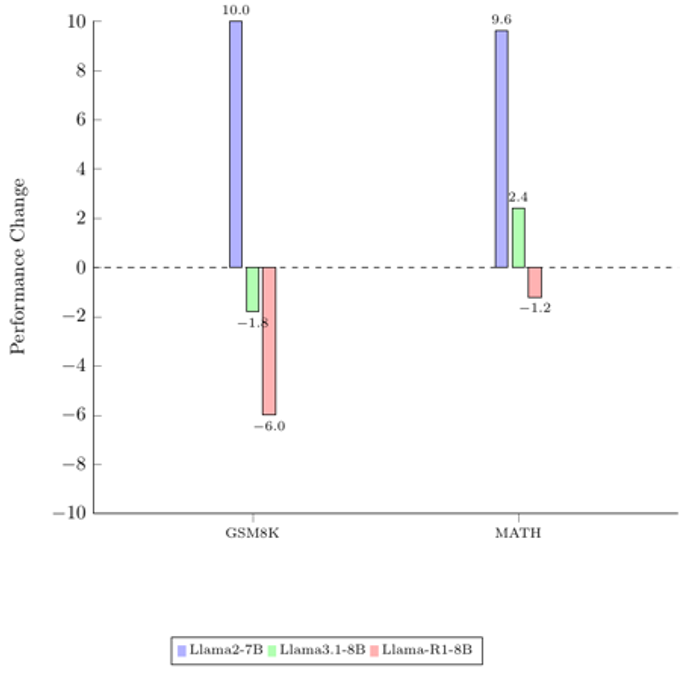}
        {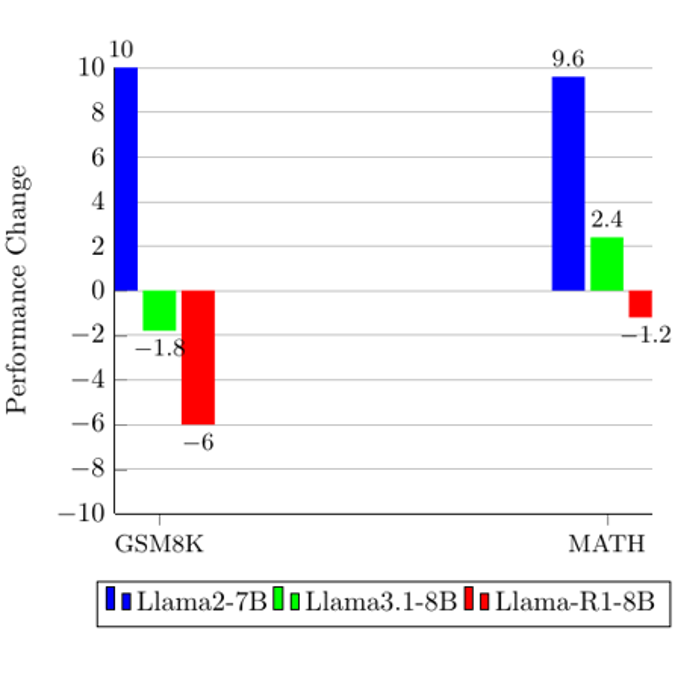}
        {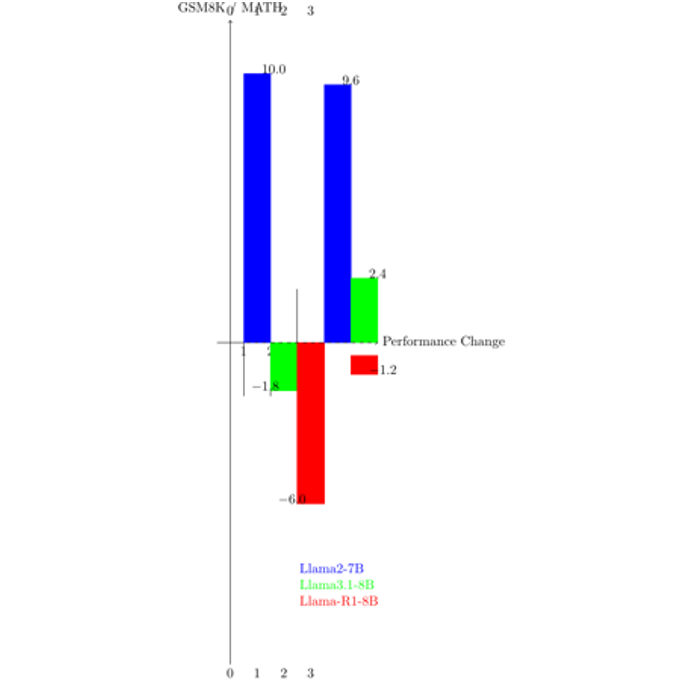}
        {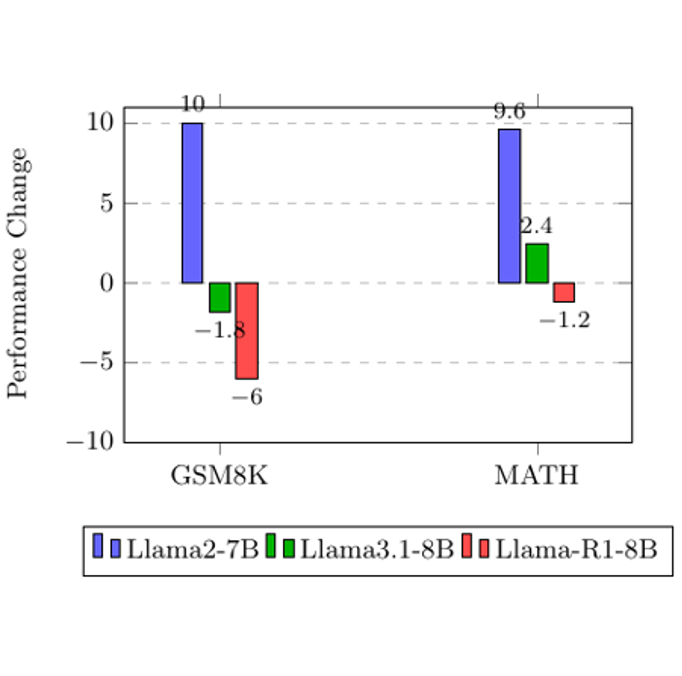}
    \midrule
    \exrowappfourthree{In the upper section, there is a central point from which four black lines radiate outward, dividing the space into four quadrants. Each quadrant is labeled in red with italicized text: "Author A" in the top left, "Author C" in the top right, "Author B" in the bottom left, and "Author D" in the bottom right. In the "Author A" quadrant, there are five red squares. In the "Author C" quadrant, there are five red diamonds. In the "Author B" quadrant, there are five red pentagons. In the "Author D" quadrant, there are five red triangles. A blue square is located near the intersection of the lines, with a black curved arrow pointing from the blue square to the "Author B" quadrant, labeled "prg" in black. Below this, a horizontal bar chart is present. The x-axis is labeled with "A", "B", "C", and "D" corresponding to the authors, and the y-axis is labeled "c" on the left side. The chart title "Confidences for prg" is centered above the bars. The bar for "A" is blue and tall, the bar for "B" is red and equally tall, while the bars for "C" and "D" are red and significantly shorter.}
        {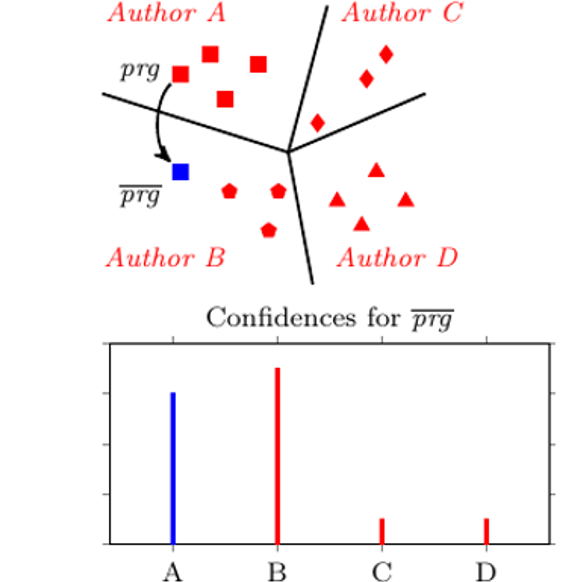}
        {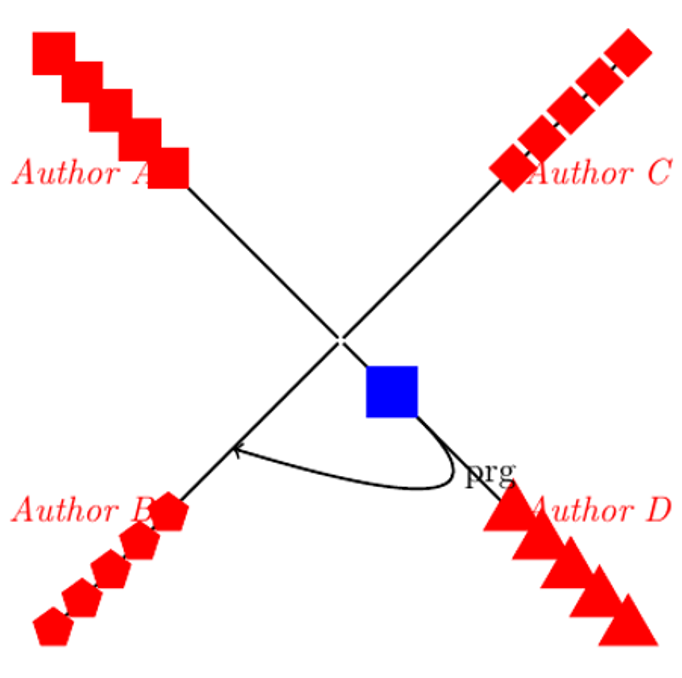}
        {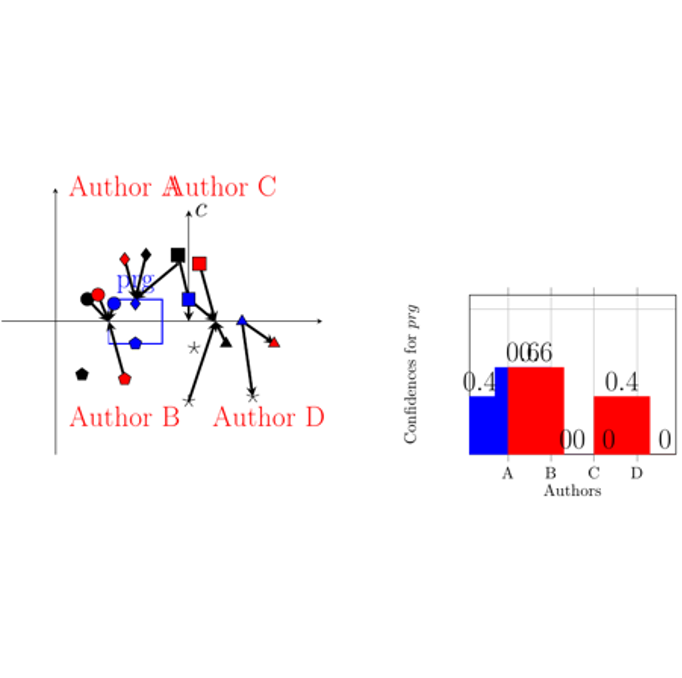}
        {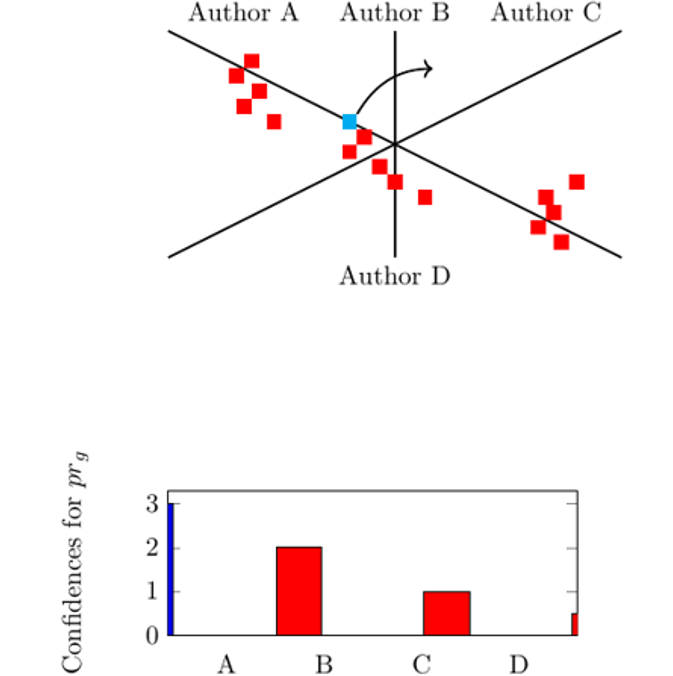}
    \midrule
    \exrowappfourfour{A red dashed square labeled $p_1$ at the top left corner and $p_2$ at the bottom left corner is positioned above a blue dashed square labeled $p_3$ at the top left corner, $p_4$ at the top right corner, and $p_5$ at the bottom right corner. The red square is filled with a light red color, and the blue square is filled with a light blue color. The red square overlaps the blue square at the top left corner of the blue square. A smaller solid purple square labeled $B_{te}$ is centered at the overlapping region. The label $p_e^+(x)$ is placed above the red square with a double-headed arrow indicating the width of the red square. The label $p_e^-(x)$ is placed below the blue square with a double-headed arrow indicating the width of the blue square. The points $p_1$ and $p_3$ are marked with red circles, while points $p_2$, $p_4$, and $p_5$ are marked with blue circles.}
        {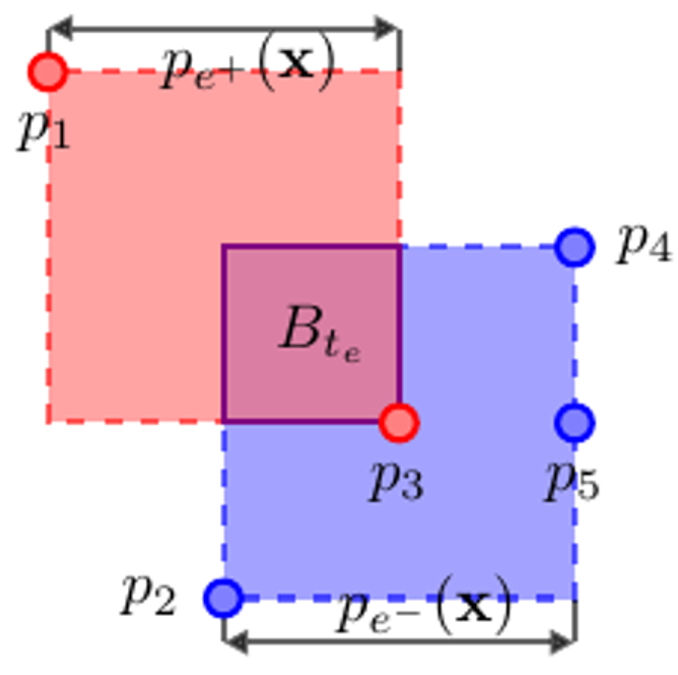}
        {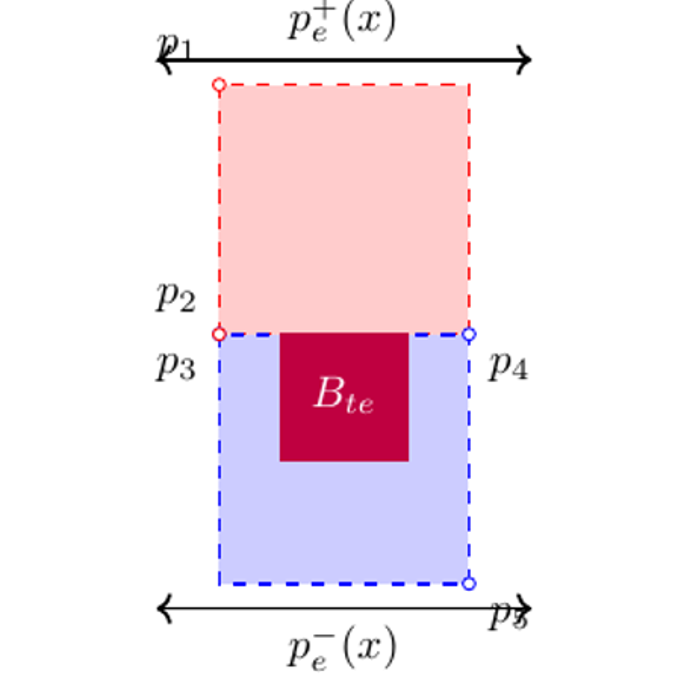}
        {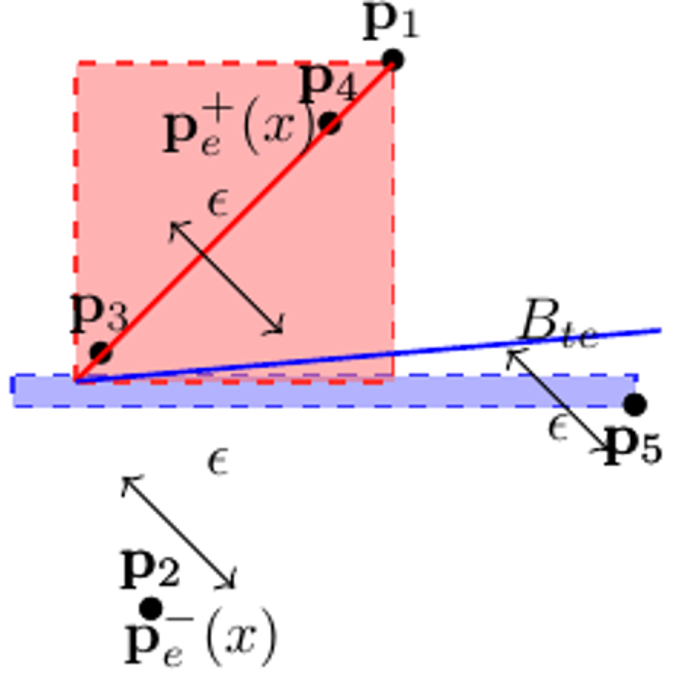}
        {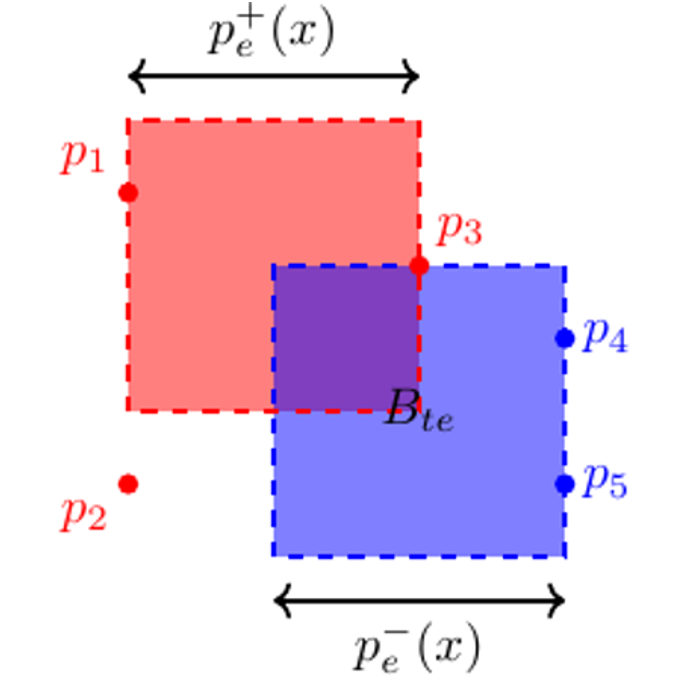}
    \midrule
    \exrowappfourfive{A sequence of green circles labeled $X_0$, $X_1$, $X_2$, and $X_{T-1}$ is arranged horizontally from left to right. Each circle is connected to the next by a rightward-pointing arrow labeled $A$. Below each circle, there is a corresponding blue square labeled $O_0$, $O_1$, $O_2$, and $O_{T-1}$, respectively. Each circle is connected to its corresponding square by a vertical black arrow labeled $B$. A dashed red horizontal line runs across the image, intersecting the vertical arrows. The sequence continues with ellipses between $X_2$ and $X_{T-1}$, and after $X_{T-1}$, indicating continuation.}
        {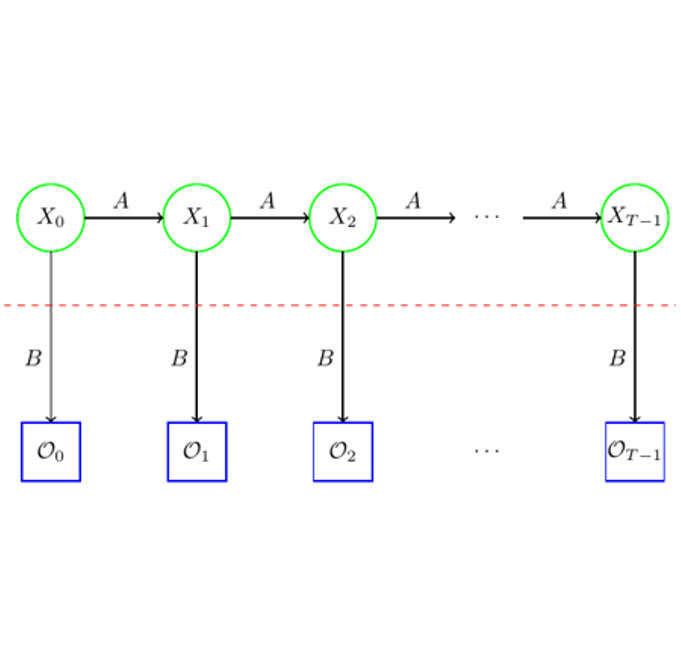}
        {structure/figures/not_compiled.png}
        {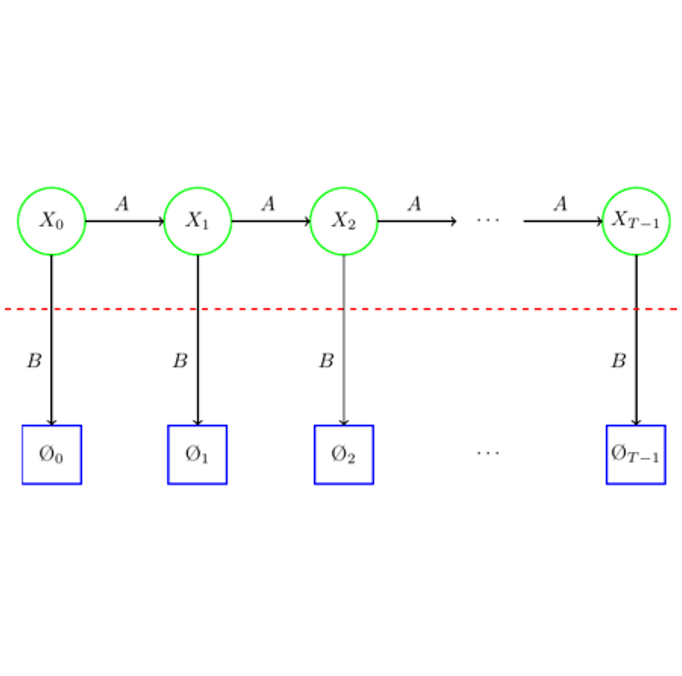}
        {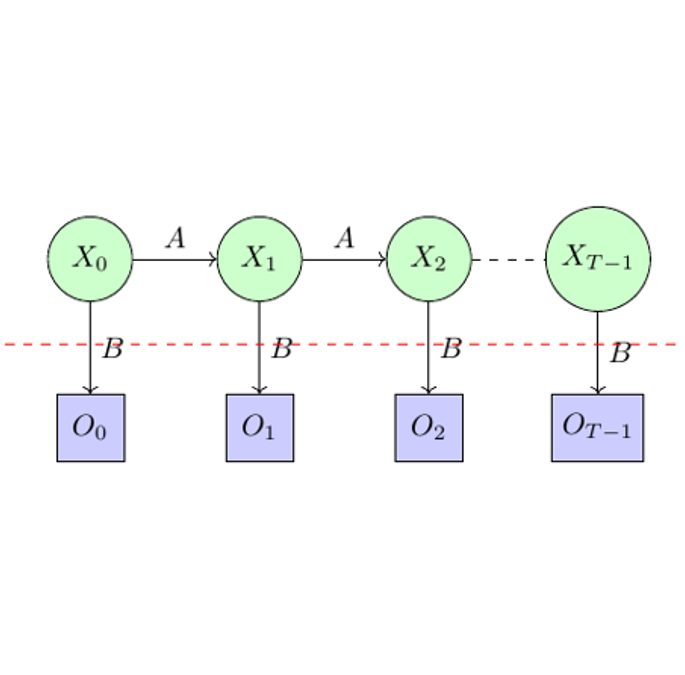}
    \midrule
    \exrowappfoursix{On the left, a blue curved line connects two black dots labeled $q_0$ at the left end and $p$ at the right end. Above the curve, a black dot labeled $q_t$ is positioned slightly to the right of center. A black arrow extends from $q_t$ pointing rightward, labeled $-\nabla_{W_2} \mathcal{F}(q_t)$. Below the curve, a red dashed line connects $q_0$ and $p$, labeled $W_2(q_0, p)$ in red. The label "Wasserstein space:" is positioned above the curve. On the right, a blue circle contains two black dots, with a black dot labeled $x_t \sim q_t$ positioned outside and to the left of the circle. A black arrow points from $x_t \sim q_t$ to the circle, labeled $v_t = -\nabla_x \frac{\delta \mathcal{F}}{\delta q_t}(x) \big|_{x=x_t}$. The label "Euclidean space:" is positioned above the circle. Another blue circle containing two black dots is positioned to the right, labeled $p$.}
        {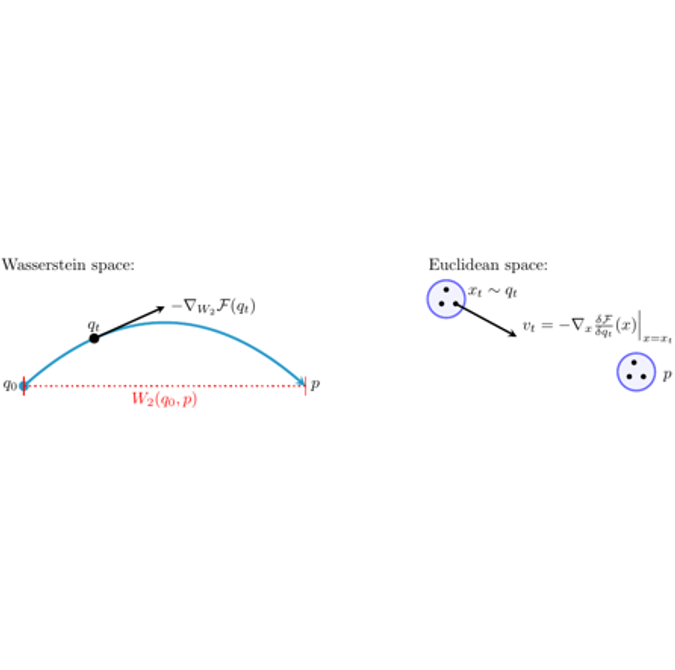}
        {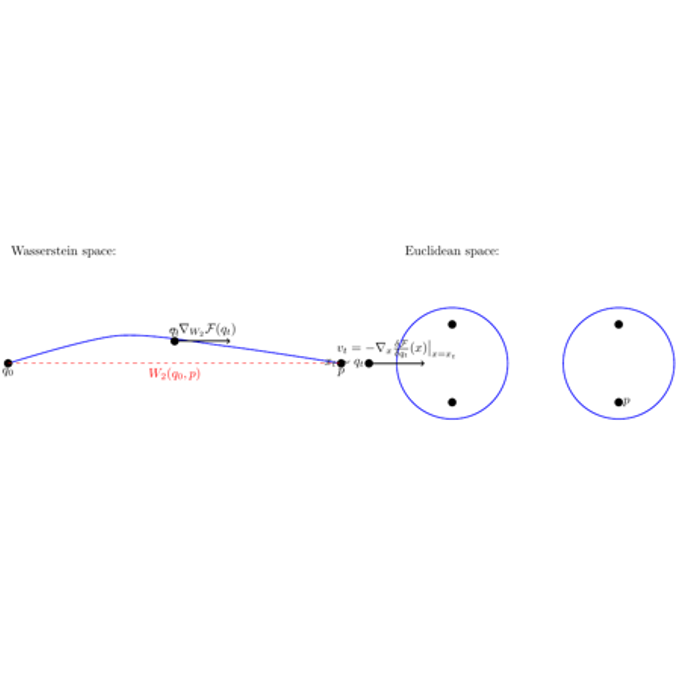}
        {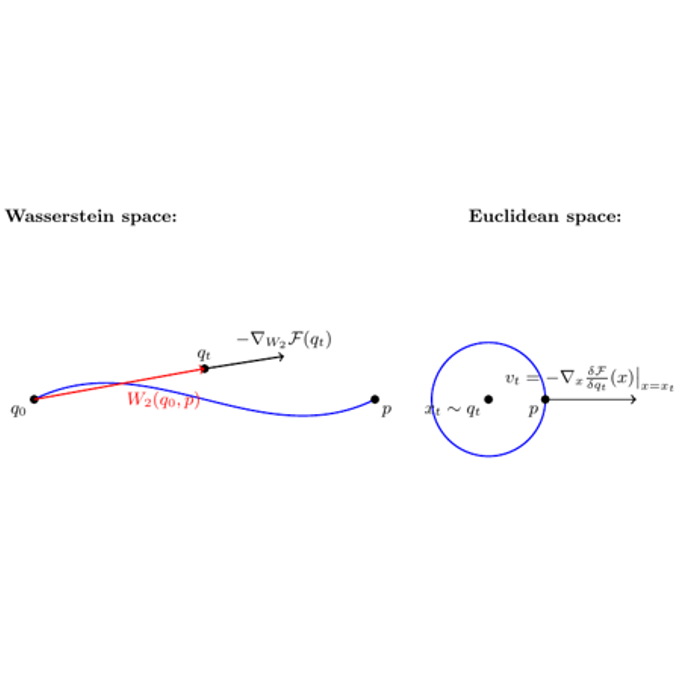}
        {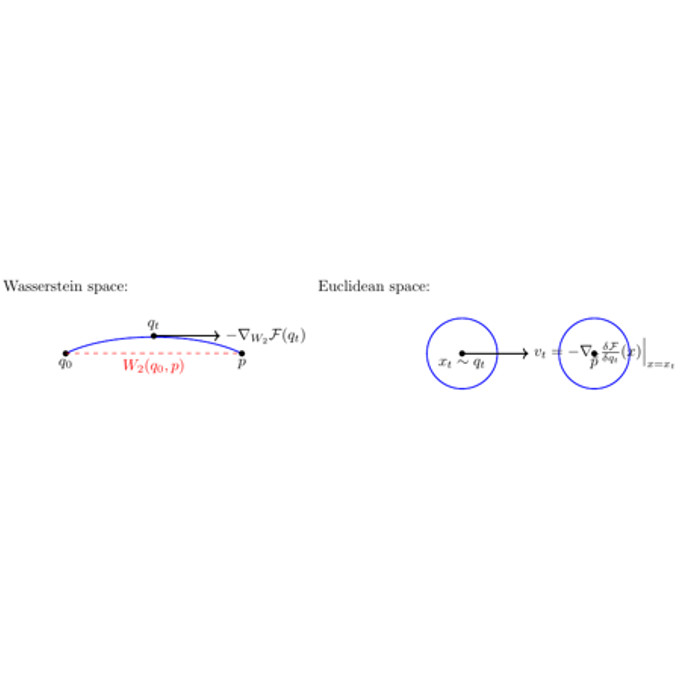}
    \midrule
    \exrowappfourseven{A box plot comparing the performance of six different models based on their 'Test NMAE (\%)', expressed on the vertical axis, which is labeled from 1 to 3. The horizontal axis lists the model names, which are, from left to right: 'Reg-Unet', 'Reg', 'Reg-VGG', 'Residual' and 'IncDice'. Each box represents the interquartile range of NMAE values for a model, with the horizontal line inside the box indicating the median value. Whiskers extend from each box to show the range of non-outlier data and individual diamond markers indicate outliers. The Reg-Unet model has the highest NMAE values with a median close to 2.5\% and a range from just above 2.0\% to over 3.0\%. Reg shows a significantly lower median around 1.1\%, with a very small spread and an outlier below 1.0\% and one around 1.2\%. Reg-VGG shows a wider interquartile range from about 1.0\% to 1.4\% and a median close to 1.2\%. The Residual model has a small spread, similar to Reg, with a slightly higher median and includes an outlier above 1.3\%. Finally, IncDice shows a slightly wider spread, with a median near 1.4\% and one upper outlier approaching 1.6\%.}
        {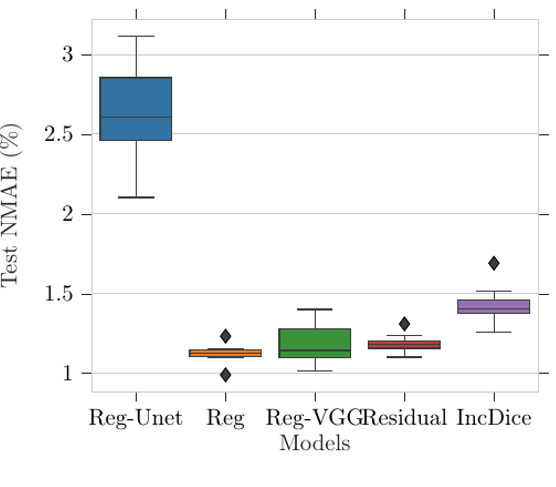}
        {structure/figures/not_compiled.png}
        {structure/figures/not_compiled.png}
        {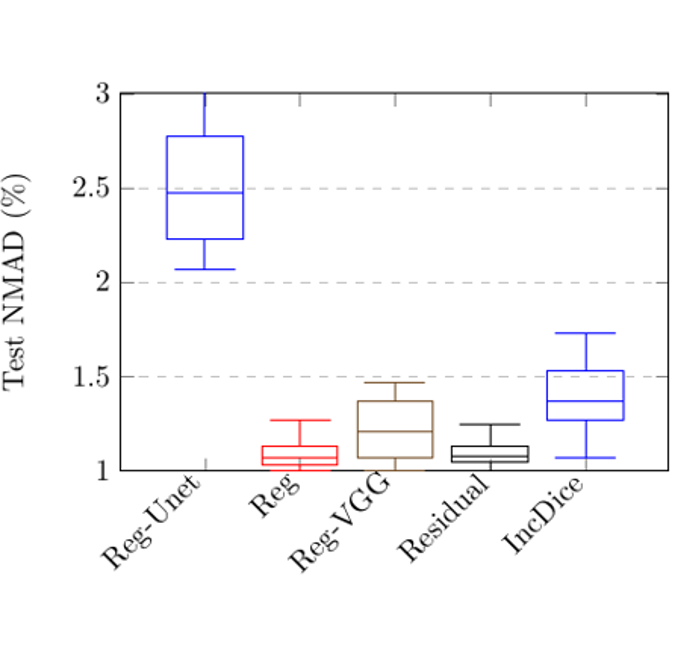}
    \bottomrule
\end{tabularx}
\end{table}

\end{document}